%% file: main.tex
\documentclass[accepted]{uai2021} 

\usepackage[american]{babel}

\usepackage{natbib} 
\usepackage{mathtools} 
\usepackage{siunitx} 
\usepackage{booktabs} 
\usepackage{colortbl}

\usepackage{url}
\usepackage[title]{appendix}
\usepackage{graphicx}
\usepackage{subfig}
\usepackage{algorithm}
\usepackage{algpseudocode}
\usepackage{soul}
\usepackage{xcolor}
\usepackage{amssymb}
\usepackage{wrapfig}
\usepackage{amsthm}
\usepackage{amsmath}
\usepackage[capitalise]{cleveref}
\usepackage{setspace}

\colorlet{tablegray}{gray!30}
\definecolor{mydarkblue}{rgb}{0,0.08,0.45}

\hypersetup{ %
  colorlinks=true,
  linkcolor=mydarkblue,
  citecolor=mydarkblue,
  filecolor=mydarkblue,
  urlcolor=mydarkblue,
}

\usepackage{pgfplots}
\pgfplotsset{compat=newest}
\usepgfplotslibrary{groupplots}
\usepgfplotslibrary{dateplot}

\usepackage{tikz}
\usepackage{tikzit}
\usetikzlibrary{external}
\usetikzlibrary{positioning}
\tikzset{>=latex}

\graphicspath{{figs/}}

\makeatletter
\def\th@plain{%
  \thm@notefont{}
  \itshape 
}
\def\th@definition{%
  \thm@notefont{}
  \normalfont 
}
\makeatother

\input{math_commands.tex}

\usepackage{thmtools}
\usepackage{thm-restate}

\newtheorem{proposition}{Proposition}

\theoremstyle{definition}

\title{Learnable Uncertainty under Laplace Approximations}

\author[1]{Agustinus Kristiadi\thanks{Correspondence to: agustinus.kristiadi@uni-tuebingen.de}} 
\author[1]{Matthias Hein}
\author[1,2]{Philipp Hennig}
\affil[1]{University of T\"{u}bingen, T\"{u}bingen, Germany}
\affil[2]{Max Planck Institute for Intelligent Systems, T\"{u}bingen, Germany}

\begin{document}
\maketitle

\begin{abstract}
  Laplace approximations are classic, computationally lightweight means for constructing Bayesian neural networks (BNNs). As in other approximate BNNs, one cannot necessarily expect the induced predictive uncertainty to be calibrated. Here we develop a formalism to explicitly ``train'' the uncertainty in a decoupled way to the prediction itself. To this end, we introduce \emph{uncertainty units} for Laplace-approximated networks: Hidden units associated with a particular weight structure that can be added to any pre-trained, point-estimated network. Due to their weights, these units are inactive---they do not affect the predictions. But their presence changes the geometry (in particular the Hessian) of the loss landscape, thereby affecting the network's uncertainty estimates under a Laplace approximation. We show that such units can be trained via an uncertainty-aware objective, improving standard Laplace approximations' performance in various uncertainty quantification tasks.
\end{abstract}

\section{Introduction}
\label{sec:intro}
\input{contents/01_intro.tex}

\section{Background}
\label{sec:background}
\input{contents/02_background.tex}

\section{Learnable Uncertainty Units under Laplace Approximations}
\label{sec:method}
\input{contents/03_method.tex}

\section{Related work}
\label{sec:related}
\input{contents/04_related.tex}

\section{Experiments}
\label{sec:experiments}
\input{contents/05_experiments.tex}

\section{Conclusion}
\label{sec:conclusion}
\input{contents/06_conclusion.tex}


\begin{acknowledgements}
    The authors gratefully acknowledge financial support by the European Research Council through ERC StG Action 757275 / PANAMA; the DFG Cluster of Excellence ``Machine Learning - New Perspectives for Science'', EXC 2064/1, project number 390727645; the German Federal Ministry of Education and Research (BMBF) through the T\"{u}bingen AI Center (FKZ: 01IS18039A); and funds from the Ministry of Science, Research and Arts of the State of Baden-W\"{u}rttemberg. The authors are also grateful to all the anonymous reviewers for their critical and helpful feedback. AK is grateful to the International Max Planck Research School for Intelligent Systems (IMPRS-IS) for support. AK also thanks all members of the Methods of Machine Learning group for helpful feedback.
\end{acknowledgements}


\bibliography{main}

\clearpage

\begin{appendices}
  \crefalias{section}{appendix}

  \section{Proofs}
  \label{appendix:proofs}
  \input{contents/91_proofs.tex}

  \section{Implementation}
  \label{appendix:implementation}
  \input{contents/92_further_details.tex}


  \section{Additional Results}
  \label{appendix:additional_exps}
  \input{contents/94_further_exps.tex}

\end{appendices}

\end{document}

%% file: math_commands.tex
\usepackage{amsmath,amsfonts,bm}

\def\1{\bm{1}}

\DeclareMathAlphabet{\mathsfit}{\encodingdefault}{\sfdefault}{m}{sl}
\SetMathAlphabet{\mathsfit}{bold}{\encodingdefault}{\sfdefault}{bx}{n}

\newcommand{\R}{\mathbb{R}}

\newcommand{\N}{\mathcal{N}}
\newcommand{\diag}[1]{\mathrm{diag}(#1)}

\renewcommand{\R}{\mathbb{R}}

\newcommand{\D}{\mathcal{D}}

\newcommand{\abs}[1]{\vert #1 \vert}
\newcommand{\inv}{{-1}}

\renewcommand{\L}{\mathcal{L}}

\newcommand{\map}{\textnormal{MAP}}

%% file: contents/01_intro.tex

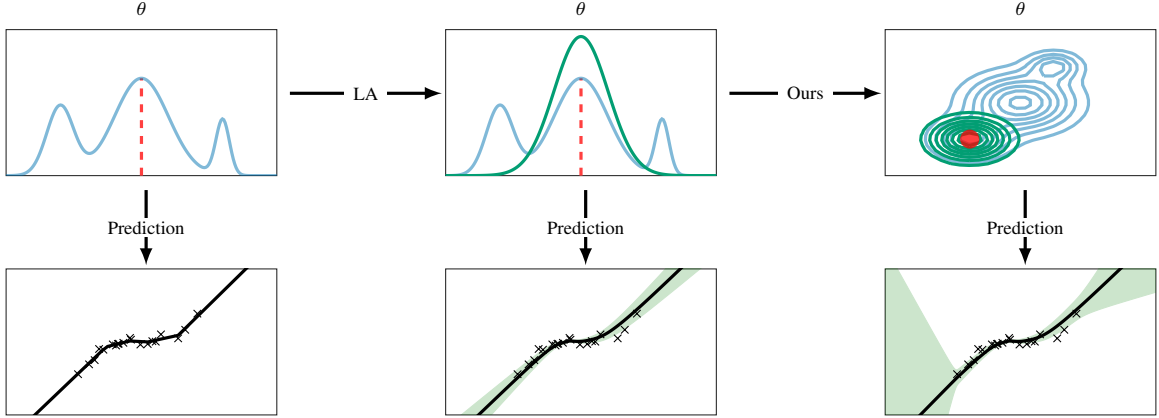
\begin{figure*}[t]
  \centering

  \input{figs/fig1.tex}

  \caption{
    A schematic of our method. \textbf{Top row:} blue and green curves represent the true and the Laplace-approximated posteriors over the parameter space, respectively---the point estimates are in red. \textbf{Bottom row:} predictions induced by the respective Laplace approximation---lines and shades are predictive means and 95\% confidence intervals, respectively. Our method adds further degrees of freedom to the parameter space---as induced by additional hidden units with a particular weight structure---and finds a point in the augmented space that induces the same predictions but with better-calibrated uncertainty estimates (esp. w.r.t. outliers), under a Laplace approximation.
  }

  \label{fig:one}
\end{figure*}

The point estimates of neural networks (NNs)---constructed as \emph{maximum a posteriori} (MAP) estimates via regularized empirical risk minimization---empirically achieve high predictive performance. However, they tend to underestimate the uncertainty of their predictions and thus be overconfident \citep{nguyen2015deep,guo17calibration}, which could be disastrous in safety-critical applications such as autonomous driving. Bayesian inference offers a principled path to overcome this issue. The goal is to turn ``vanilla''  NNs into Bayesian neural networks (BNNs), i.e. equipping a NN with the posterior over its weights, inferred by Bayes' theorem and subsequently taken into account when making predictions \citep{mackay1992practical,neal1995bayesian}.

Since the cost of exact posterior inference in a BNN is often prohibitive, approximate Bayesian methods are commonly employed instead. Laplace approximations (LAs) are classic methods for such a purpose \citep{mackay1992practical}. Intuitively, the key idea is to obtain an approximate posterior by ``surrounding'' a MAP estimate of a network with a Gaussian, based on the loss landscape's geometry around it. More formally, they form a Gaussian approximation to the exact posterior, whose mean equals the network's MAP estimate and whose covariance equals the negative inverse Hessian (or approximations thereof) of the loss function, evaluated at the MAP estimate. LAs can thus be applied to any pre-trained, point-estimated network in a cost-efficient, \emph{post-hoc} manner, especially thanks to recent advances in software toolkits for second-order optimization \citep{yao2019pyhessian,dangel2020backpack}. This is in contrast to alternative approximate Bayesian methods such as variational Bayes \citep{hinton1993keeping,graves_practical_2011,blundell_weight_2015} and Markov Chain Monte Carlo \citep{neal1993bayesian,welling2011bayesian} which require either costly network re-training or posterior sampling.

A standard practice in contemporary LAs is to tune a single hyperparameter---the prior precision---to calibrate their predictive uncertainty \citep{ritter_scalable_2018}. However, this scalar parametrization allows only for a very limited form of uncertainty calibration. Below, we propose a more flexible framework to tune the uncertainty of Laplace-approximated BNNs without changing their point estimates. The idea is to introduce additional hidden units, associated with partly zero weights, to the hidden layers of any MAP-trained network. Because of their weight structure, they are partly inactive and do not affect the prediction of the underlying network. However, they can still contribute to the Hessian of the loss with respect to the parameters, and hence induce additional structure to the posterior covariance under a Laplace approximation---these units are thus \emph{uncertainty units} under Laplace approximations. Furthermore, the non-zero weights associated with these units can then be trained via an uncertainty-aware objective \citep[etc.]{lee2018training,hendrycks2018deep}, such that they improve the predictive uncertainty quantification performance of the Laplace-approximated BNN. \Cref{fig:one} provides intuition.

In summary, we
\begin{enumerate}[label=(\roman*)]
    \item introduce uncertainty units: hidden units with a particular structure in their associated weights that can be applied to any MAP-trained network,
    \item show that these units maintain the output of the network, while non-trivially affecting the loss landscape's curvature (the Hessian), thus also affecting predictive uncertainty under Laplace approximations, and
    \item present a training method for the non-zero weights associated with these units via an uncertainty-aware objective so that they improve the uncertainty calibration of Laplace approximations.
\end{enumerate}

%% file: figs/fig1.tex

\begin{tikzpicture}[align=center]

\tikzstyle{every node}=[font=\scriptsize]

\node[inner sep=1pt] (map) at (0, 0) {
    \input{figs/fig1_ori_map}
};
\node[inner sep=1pt, right=2cm of map] (laplace) {
    \input{figs/fig1_laplace_map}
};
\node[inner sep=1pt, right=2cm of laplace] (lula) {
    \input{figs/fig1_lula_map}
};

\node[inner sep=1pt, below=1cm of map] (pred_map) {
    \input{figs/toy_reg_map}
};
\node[inner sep=1pt, below=1cm of laplace] (pred_laplace) {
    \input{figs/toy_reg_laplace}
};
\node[inner sep=1pt, below=1cm of lula] (pred_lula) {
    \input{figs/toy_reg_lula}
};

\draw[->, very thick] (map.east) -- (laplace.west)
    node[midway, fill=white] {LA};

\draw[->, very thick] (laplace.east) -- (lula.west)
    node[midway, fill=white] {Ours};

\draw[->, very thick, inner sep=1pt] (map.south) -- (pred_map.north)
    node[midway, fill=white] {Prediction};

\draw[->, very thick, inner sep=1pt] (laplace.south) -- (pred_laplace.north)
    node[midway, fill=white] {Prediction};

\draw[->, very thick, inner sep=1pt] (lula.south) -- (pred_lula.north)
    node[midway, fill=white] {Prediction};

\end{tikzpicture}

%% file: figs/fig1_ori_map.tex
\begin{tikzpicture}

\definecolor{color0}{rgb}{0.00392156862745098,0.450980392156863,0.698039215686274}
\definecolor{color1}{rgb}{0.00784313725490196,0.619607843137255,0.450980392156863}

\begin{axis}[
width=0.3\textwidth,
height=0.15\textheight,
axis line style={white!15!black},
tick align=outside,
x grid style={white!80!black},
xmin=-5, xmax=5,
xtick style={draw=none},
xtick={0},
xticklabels={$\theta$},
xticklabel pos=top,
ymajorticks=false,
ymin=0, ymax=0.418757193438432
]
\draw[draw=red!75, dashed, very thick] (axis cs:0,0) -- (axis cs:0,0.279171464536624);

\addplot [very thick, color0, opacity=0.5]
table {%
-5 6.79558165427567e-05
-4.94974874371859 0.000100858857916717
-4.89949748743719 0.000148243422709998
-4.84924623115578 0.000215764392514243
-4.79899497487437 0.000310958605662428
-4.74874371859296 0.000443735776315003
-4.69849246231156 0.000626945713816154
-4.64824120603015 0.000877014433258935
-4.59798994974874 0.00121463276439035
-4.54773869346734 0.00166546983333596
-4.49748743718593 0.00226087057433293
-4.44723618090452 0.00303848183432533
-4.39698492462312 0.00404273667229508
-4.34673366834171 0.00532511256693119
-4.2964824120603 0.00694406827500631
-4.24623115577889 0.00896455818868653
-4.19597989949749 0.0114570245341206
-4.14572864321608 0.0144957788568444
-4.09547738693467 0.0181567067741517
-4.04522613065327 0.0225142650141361
-3.99497487437186 0.0276377872761245
-3.94472361809045 0.0335871739981593
-3.89447236180905 0.0404081076522888
-3.84422110552764 0.0481270050041539
-3.79396984924623 0.0567459846844289
-3.74371859296482 0.066238185165634
-3.69346733668342 0.0765438071225232
-3.64321608040201 0.0875672678784598
-3.5929648241206 0.099175838249875
-3.5427135678392 0.111200079940034
-3.49246231155779 0.1234363141693
-3.44221105527638 0.135651232613802
-3.39195979899498 0.147588616971725
-3.34170854271357 0.158977974174059
-3.29145728643216 0.169544733789785
-3.24120603015075 0.179021507504865
-3.19095477386935 0.187159792666319
-3.14070351758794 0.193741426083069
-3.09045226130653 0.198589070545873
-3.04020100502513 0.201575050166857
-2.98994974874372 0.202627941383632
-2.93969849246231 0.201736468259949
-2.8894472361809 0.198950432126569
-2.8391959798995 0.194378610981014
-2.78894472361809 0.188183775143634
-2.73869346733668 0.180575163569509
-2.68844221105528 0.171798932497257
-2.63819095477387 0.162127210642419
-2.58793969849246 0.151846463541555
-2.53768844221106 0.141245880167285
-2.48743718592965 0.130606449594763
-2.43718592964824 0.120191301614952
-2.38693467336683 0.110237754256244
-2.33668341708543 0.10095135737806
-2.28643216080402 0.0925020600151453
-2.23618090452261 0.0850224744804104
-2.18592964824121 0.0786080747662244
-2.1356783919598 0.0733190597412416
-2.08542713567839 0.0691835384957018
-2.03517587939698 0.0662016575716556
-1.98492462311558 0.0643502858536434
-1.93467336683417 0.0635878978992982
-1.88442211055276 0.0638593438419163
-1.83417085427136 0.0651002561225628
-1.78391959798995 0.0672409125863458
-1.73366834170854 0.0702094450017666
-1.68341708542714 0.0739343461954618
-1.63316582914573 0.0783462836760967
-1.58291457286432 0.0833792704484952
-1.53266331658291 0.0889712738382593
-1.48241206030151 0.0950643610013803
-1.4321608040201 0.101604486790864
-1.38190954773869 0.108541027808602
-1.33165829145729 0.115826158071232
-1.28140703517588 0.123414149036096
-1.23115577889447 0.131260661816241
-1.18090452261307 0.13932208395261
-1.13065326633166 0.147554948379354
-1.08040201005025 0.15591545906561
-1.03015075376884 0.16435913671604
-0.979899497487437 0.172840589023395
-0.92964824120603 0.181313403217821
-0.879396984924623 0.18973015382522
-0.829145728643216 0.198042515322853
-0.778894472361809 0.206201467429987
-0.728643216080402 0.214157579773663
-0.678391959798994 0.221861362343071
-0.628140703517587 0.229263668262089
-0.57788944723618 0.236316135795474
-0.527638190954773 0.242971657040134
-0.477386934673366 0.249184861363507
-0.427135678391959 0.254912602296458
-0.376884422110552 0.260114437253762
-0.326633165829145 0.264753090143136
-0.276381909547738 0.268794887644798
-0.226130653266331 0.272210160711817
-0.175879396984924 0.274973603669949
-0.125628140703517 0.277064584193615
-0.0753768844221101 0.278467398405987
-0.0251256281407031 0.279171466394084
0.025125628140704 0.279171464536624
0.075376884422111 0.278467392200466
0.125628140703518 0.277064571553602
0.175879396984925 0.274973580448543
0.226130653266332 0.272210119526885
0.276381909547739 0.268794815860034
0.326633165829146 0.264752966548691
0.376884422110553 0.260114226731481
0.42713567839196 0.254912247379736
0.477386934673367 0.24918426906177
0.527638190954774 0.242970678530014
0.577889447236181 0.236314535505515
0.628140703517588 0.229261077397817
0.678391959798995 0.221857209899495
0.728643216080402 0.214150991432878
0.778894472361809 0.206191119305371
0.829145728643216 0.198026425156385
0.879396984924623 0.189705386864191
0.929648241206031 0.181275663538844
0.979899497487438 0.172783659574437
1.03015075376884 0.164274122988973
1.08040201005025 0.155789782463447
1.13065326633166 0.147371026624425
1.18090452261307 0.139055628218051
1.23115577889447 0.13087851491936
1.28140703517588 0.12287158762965
1.33165829145729 0.115063586257054
1.38190954773869 0.107480002172637
1.4321608040201 0.10014303581443
1.48241206030151 0.0930715973235079
1.53266331658291 0.0862813477457613
1.58291457286432 0.0797847784683799
1.63316582914573 0.0735913277496372
1.68341708542714 0.0677075366795053
1.73366834170854 0.062137255172913
1.78391959798995 0.0568819262754656
1.83417085427136 0.0519410118782927
1.88442211055276 0.0473126861089722
1.93467336683417 0.042995027239407
1.98492462311558 0.0389880937076741
2.03517587939699 0.0352974660056385
2.08542713567839 0.031940025887588
2.1356783919598 0.0289528153750819
2.18592964824121 0.0264055788182372
2.23618090452261 0.0244167919080825
2.28643216080402 0.0231714005868086
2.33668341708543 0.0229361048006881
2.38693467336683 0.0240652149071817
2.43718592964824 0.0269878502356168
2.48743718592965 0.0321670417338721
2.53768844221106 0.0400247664069747
2.58793969849246 0.0508350197125202
2.63819095477387 0.0645990140461274
2.68844221105528 0.0809294908159007
2.73869346733668 0.0989799850595056
2.78894472361809 0.117454287491008
2.8391959798995 0.134717967798718
2.8894472361809 0.149008765048064
2.93969849246231 0.158712390384148
2.98994974874372 0.162645109756901
3.04020100502513 0.1602746356486
3.09045226130653 0.151821986116238
3.14070351758794 0.138216986210477
3.19095477386935 0.120919405810523
3.24120603015075 0.101652943202475
3.29145728643216 0.0821186635153608
3.34170854271357 0.0637527665670481
3.39195979899498 0.0475734474939742
3.44221105527638 0.034132049410167
3.49246231155779 0.0235555143255341
3.5427135678392 0.0156486494678393
3.5929648241206 0.0100192129522735
3.64321608040201 0.00619448934116349
3.69346733668342 0.00370997360368805
3.74371859296483 0.00216361952711671
3.79396984924623 0.00123895411562144
3.84422110552764 0.000705628809406842
3.89447236180905 0.000407061837700499
3.94472361809045 0.000243186819551659
3.99497487437186 0.000153584262141068
4.04522613065327 0.000103639804873161
4.09547738693467 7.44537558891065e-05
4.14572864321608 5.61342400787938e-05
4.19597989949749 4.36738214429802e-05
4.2462311557789 3.45833220308346e-05
4.2964824120603 2.76162745363451e-05
4.34673366834171 2.21195270421219e-05
4.39698492462312 1.77192757757784e-05
4.44723618090452 1.4175814188257e-05
4.49748743718593 1.13183866923619e-05
4.54773869346734 9.01615204262845e-06
4.59798994974874 7.16473587852552e-06
4.64824120603015 5.67933504631681e-06
4.69849246231156 4.49059289637496e-06
4.74874371859297 3.54172783472277e-06
4.79899497487437 2.78631803255883e-06
4.84924623115578 2.18650126748012e-06
4.89949748743719 1.71148136919749e-06
4.94974874371859 1.3362815250375e-06
5 1.04070366233492e-06
};
\end{axis}

\end{tikzpicture}

%% file: figs/fig1_laplace_map.tex
\begin{tikzpicture}

\definecolor{color0}{rgb}{0.00392156862745098,0.450980392156863,0.698039215686274}
\definecolor{color1}{rgb}{0.00784313725490196,0.619607843137255,0.450980392156863}

\begin{axis}[
width=0.3\textwidth,
height=0.15\textheight,
axis line style={white!15!black},
tick align=outside,
x grid style={white!80!black},
xmin=-5, xmax=5,
xtick style={draw=none},
xtick={0},
xticklabels={$\theta$},
xticklabel pos=top,
ymajorticks=false,
ymin=0, ymax=0.418757193438432
]
\draw[draw=red!75, dashed, very thick] (axis cs:0,0) -- (axis cs:0,0.279171464536624);

\addplot [very thick, color0, opacity=0.5]
table {%
-5 6.79558165427567e-05
-4.94974874371859 0.000100858857916717
-4.89949748743719 0.000148243422709998
-4.84924623115578 0.000215764392514243
-4.79899497487437 0.000310958605662428
-4.74874371859296 0.000443735776315003
-4.69849246231156 0.000626945713816154
-4.64824120603015 0.000877014433258935
-4.59798994974874 0.00121463276439035
-4.54773869346734 0.00166546983333596
-4.49748743718593 0.00226087057433293
-4.44723618090452 0.00303848183432533
-4.39698492462312 0.00404273667229508
-4.34673366834171 0.00532511256693119
-4.2964824120603 0.00694406827500631
-4.24623115577889 0.00896455818868653
-4.19597989949749 0.0114570245341206
-4.14572864321608 0.0144957788568444
-4.09547738693467 0.0181567067741517
-4.04522613065327 0.0225142650141361
-3.99497487437186 0.0276377872761245
-3.94472361809045 0.0335871739981593
-3.89447236180905 0.0404081076522888
-3.84422110552764 0.0481270050041539
-3.79396984924623 0.0567459846844289
-3.74371859296482 0.066238185165634
-3.69346733668342 0.0765438071225232
-3.64321608040201 0.0875672678784598
-3.5929648241206 0.099175838249875
-3.5427135678392 0.111200079940034
-3.49246231155779 0.1234363141693
-3.44221105527638 0.135651232613802
-3.39195979899498 0.147588616971725
-3.34170854271357 0.158977974174059
-3.29145728643216 0.169544733789785
-3.24120603015075 0.179021507504865
-3.19095477386935 0.187159792666319
-3.14070351758794 0.193741426083069
-3.09045226130653 0.198589070545873
-3.04020100502513 0.201575050166857
-2.98994974874372 0.202627941383632
-2.93969849246231 0.201736468259949
-2.8894472361809 0.198950432126569
-2.8391959798995 0.194378610981014
-2.78894472361809 0.188183775143634
-2.73869346733668 0.180575163569509
-2.68844221105528 0.171798932497257
-2.63819095477387 0.162127210642419
-2.58793969849246 0.151846463541555
-2.53768844221106 0.141245880167285
-2.48743718592965 0.130606449594763
-2.43718592964824 0.120191301614952
-2.38693467336683 0.110237754256244
-2.33668341708543 0.10095135737806
-2.28643216080402 0.0925020600151453
-2.23618090452261 0.0850224744804104
-2.18592964824121 0.0786080747662244
-2.1356783919598 0.0733190597412416
-2.08542713567839 0.0691835384957018
-2.03517587939698 0.0662016575716556
-1.98492462311558 0.0643502858536434
-1.93467336683417 0.0635878978992982
-1.88442211055276 0.0638593438419163
-1.83417085427136 0.0651002561225628
-1.78391959798995 0.0672409125863458
-1.73366834170854 0.0702094450017666
-1.68341708542714 0.0739343461954618
-1.63316582914573 0.0783462836760967
-1.58291457286432 0.0833792704484952
-1.53266331658291 0.0889712738382593
-1.48241206030151 0.0950643610013803
-1.4321608040201 0.101604486790864
-1.38190954773869 0.108541027808602
-1.33165829145729 0.115826158071232
-1.28140703517588 0.123414149036096
-1.23115577889447 0.131260661816241
-1.18090452261307 0.13932208395261
-1.13065326633166 0.147554948379354
-1.08040201005025 0.15591545906561
-1.03015075376884 0.16435913671604
-0.979899497487437 0.172840589023395
-0.92964824120603 0.181313403217821
-0.879396984924623 0.18973015382522
-0.829145728643216 0.198042515322853
-0.778894472361809 0.206201467429987
-0.728643216080402 0.214157579773663
-0.678391959798994 0.221861362343071
-0.628140703517587 0.229263668262089
-0.57788944723618 0.236316135795474
-0.527638190954773 0.242971657040134
-0.477386934673366 0.249184861363507
-0.427135678391959 0.254912602296458
-0.376884422110552 0.260114437253762
-0.326633165829145 0.264753090143136
-0.276381909547738 0.268794887644798
-0.226130653266331 0.272210160711817
-0.175879396984924 0.274973603669949
-0.125628140703517 0.277064584193615
-0.0753768844221101 0.278467398405987
-0.0251256281407031 0.279171466394084
0.025125628140704 0.279171464536624
0.075376884422111 0.278467392200466
0.125628140703518 0.277064571553602
0.175879396984925 0.274973580448543
0.226130653266332 0.272210119526885
0.276381909547739 0.268794815860034
0.326633165829146 0.264752966548691
0.376884422110553 0.260114226731481
0.42713567839196 0.254912247379736
0.477386934673367 0.24918426906177
0.527638190954774 0.242970678530014
0.577889447236181 0.236314535505515
0.628140703517588 0.229261077397817
0.678391959798995 0.221857209899495
0.728643216080402 0.214150991432878
0.778894472361809 0.206191119305371
0.829145728643216 0.198026425156385
0.879396984924623 0.189705386864191
0.929648241206031 0.181275663538844
0.979899497487438 0.172783659574437
1.03015075376884 0.164274122988973
1.08040201005025 0.155789782463447
1.13065326633166 0.147371026624425
1.18090452261307 0.139055628218051
1.23115577889447 0.13087851491936
1.28140703517588 0.12287158762965
1.33165829145729 0.115063586257054
1.38190954773869 0.107480002172637
1.4321608040201 0.10014303581443
1.48241206030151 0.0930715973235079
1.53266331658291 0.0862813477457613
1.58291457286432 0.0797847784683799
1.63316582914573 0.0735913277496372
1.68341708542714 0.0677075366795053
1.73366834170854 0.062137255172913
1.78391959798995 0.0568819262754656
1.83417085427136 0.0519410118782927
1.88442211055276 0.0473126861089722
1.93467336683417 0.042995027239407
1.98492462311558 0.0389880937076741
2.03517587939699 0.0352974660056385
2.08542713567839 0.031940025887588
2.1356783919598 0.0289528153750819
2.18592964824121 0.0264055788182372
2.23618090452261 0.0244167919080825
2.28643216080402 0.0231714005868086
2.33668341708543 0.0229361048006881
2.38693467336683 0.0240652149071817
2.43718592964824 0.0269878502356168
2.48743718592965 0.0321670417338721
2.53768844221106 0.0400247664069747
2.58793969849246 0.0508350197125202
2.63819095477387 0.0645990140461274
2.68844221105528 0.0809294908159007
2.73869346733668 0.0989799850595056
2.78894472361809 0.117454287491008
2.8391959798995 0.134717967798718
2.8894472361809 0.149008765048064
2.93969849246231 0.158712390384148
2.98994974874372 0.162645109756901
3.04020100502513 0.1602746356486
3.09045226130653 0.151821986116238
3.14070351758794 0.138216986210477
3.19095477386935 0.120919405810523
3.24120603015075 0.101652943202475
3.29145728643216 0.0821186635153608
3.34170854271357 0.0637527665670481
3.39195979899498 0.0475734474939742
3.44221105527638 0.034132049410167
3.49246231155779 0.0235555143255341
3.5427135678392 0.0156486494678393
3.5929648241206 0.0100192129522735
3.64321608040201 0.00619448934116349
3.69346733668342 0.00370997360368805
3.74371859296483 0.00216361952711671
3.79396984924623 0.00123895411562144
3.84422110552764 0.000705628809406842
3.89447236180905 0.000407061837700499
3.94472361809045 0.000243186819551659
3.99497487437186 0.000153584262141068
4.04522613065327 0.000103639804873161
4.09547738693467 7.44537558891065e-05
4.14572864321608 5.61342400787938e-05
4.19597989949749 4.36738214429802e-05
4.2462311557789 3.45833220308346e-05
4.2964824120603 2.76162745363451e-05
4.34673366834171 2.21195270421219e-05
4.39698492462312 1.77192757757784e-05
4.44723618090452 1.4175814188257e-05
4.49748743718593 1.13183866923619e-05
4.54773869346734 9.01615204262845e-06
4.59798994974874 7.16473587852552e-06
4.64824120603015 5.67933504631681e-06
4.69849246231156 4.49059289637496e-06
4.74874371859297 3.54172783472277e-06
4.79899497487437 2.78631803255883e-06
4.84924623115578 2.18650126748012e-06
4.89949748743719 1.71148136919749e-06
4.94974874371859 1.3362815250375e-06
5 1.04070366233492e-06
};
\addplot [very thick, color1]
table {%
-5 1.4867195147343e-06
-4.94974874371859 1.90897359306982e-06
-4.89949748743719 2.44497331818117e-06
-4.84924623115578 3.12357293963311e-06
-4.79899497487437 3.98045303347538e-06
-4.74874371859296 5.05960578580913e-06
-4.69849246231156 6.41511109267784e-06
-4.64824120603015 8.11325277766369e-06
-4.59798994974874 1.02350308754708e-05
-4.54773869346734 1.28791331028226e-05
-4.49748743718593 1.61654362953512e-05
-4.44723618090452 2.02391166640527e-05
-4.39698492462312 2.52754561211026e-05
-4.34673366834171 3.14854405119587e-05
-4.2964824120603 3.91222541992946e-05
-4.24623115577889 4.84887838612558e-05
-4.19597989949749 5.99462523304611e-05
-4.14572864321608 7.39241104976097e-05
-4.09547738693467 9.09313213657291e-05
-4.04522613065327 0.000111569174841876
-3.99497487437186 0.000136545774308338
-3.94472361809045 0.000166692335884057
-3.89447236180905 0.000202981437974616
-3.84422110552764 0.000246547351573483
-3.79396984924623 0.000298708570137427
-3.74371859296482 0.000360992641007504
-3.69346733668342 0.000435163377565355
-3.64321608040201 0.000523250501893059
-3.5929648241206 0.000627581730965922
-3.5427135678392 0.00075081727473373
-3.49246231155779 0.000895986661310447
-3.44221105527638 0.00106652774249448
-3.39195979899498 0.00126632766174431
-3.34170854271357 0.00149976548650443
-3.29145728643216 0.00177175611762511
-3.24120603015075 0.00208779499104519
-3.19095477386935 0.00245400298173195
-3.14070351758794 0.00287717080828339
-3.09045226130653 0.00336480212017405
-3.04020100502513 0.00392515433036495
-2.98994974874372 0.00456727613633012
-2.93969849246231 0.00530104055534539
-2.8894472361809 0.00613717218843179
-2.8391959798995 0.00708726732533951
-2.78894472361809 0.00816380541445068
-2.73869346733668 0.00938015035081449
-2.68844221105528 0.0107505399872646
-2.63819095477387 0.0122900622523775
-2.58793969849246 0.0140146162695652
-2.53768844221106 0.0159408569183735
-2.48743718592965 0.0180861213662637
-2.43718592964824 0.020468336230535
-2.38693467336683 0.0231059042086428
-2.33668341708543 0.0260175692432377
-2.28643216080402 0.0292222595669734
-2.23618090452261 0.032738908301549
-2.18592964824121 0.0365862516641934
-2.1356783919598 0.0407826052600598
-2.08542713567839 0.0453456194063217
-2.03517587939698 0.0502920149370724
-1.98492462311558 0.0556373014696323
-1.93467336683417 0.0613954806631067
-1.88442211055276 0.0675787375580062
-1.83417085427136 0.0741971236389479
-1.78391959798995 0.0812582357972332
-1.73366834170854 0.0887668958717688
-1.68341708542714 0.0967248359000726
-1.63316582914573 0.105130394600424
-1.58291457286432 0.113978230916186
-1.53266331658291 0.123259060669182
-1.48241206030151 0.132959422477082
-1.4321608040201 0.143061479078016
-1.38190954773869 0.153542860064127
-1.33165829145729 0.164376551747098
-1.28140703517588 0.175530839458451
-1.23115577889447 0.186969307024586
-1.18090452261307 0.198650897453633
-1.13065326633166 0.210530038034727
-1.08040201005025 0.222556832090602
-1.03015075376884 0.234677318555666
-0.979899497487437 0.246833799392047
-0.92964824120603 0.258965233626909
-0.879396984924623 0.271007695520249
-0.829145728643216 0.282894893080498
-0.778894472361809 0.294558741864703
-0.728643216080402 0.305929987761015
-0.678391959798994 0.316938871284489
-0.628140703517587 0.327515824852975
-0.57788944723618 0.337592193577141
-0.527638190954773 0.347100969324167
-0.477386934673366 0.355977527222154
-0.427135678391959 0.364160353381718
-0.376884422110552 0.371591752438066
-0.326633165829145 0.378218523571396
-0.276381909547738 0.38399259395063
-0.226130653266331 0.388871599064357
-0.175879396984924 0.392819400146452
-0.125628140703517 0.395806529859628
-0.0753768844221101 0.397810558549711
-0.0251256281407031 0.398816374703268
0.025125628140704 0.398816374703268
0.075376884422111 0.397810558549711
0.125628140703518 0.395806529859628
0.175879396984925 0.392819400146452
0.226130653266332 0.388871599064357
0.276381909547739 0.38399259395063
0.326633165829146 0.378218523571396
0.376884422110553 0.371591752438066
0.42713567839196 0.364160353381717
0.477386934673367 0.355977527222154
0.527638190954774 0.347100969324167
0.577889447236181 0.337592193577141
0.628140703517588 0.327515824852974
0.678391959798995 0.316938871284488
0.728643216080402 0.305929987761015
0.778894472361809 0.294558741864703
0.829145728643216 0.282894893080497
0.879396984924623 0.271007695520249
0.929648241206031 0.258965233626909
0.979899497487438 0.246833799392046
1.03015075376884 0.234677318555666
1.08040201005025 0.222556832090602
1.13065326633166 0.210530038034727
1.18090452261307 0.198650897453633
1.23115577889447 0.186969307024586
1.28140703517588 0.175530839458451
1.33165829145729 0.164376551747097
1.38190954773869 0.153542860064127
1.4321608040201 0.143061479078016
1.48241206030151 0.132959422477082
1.53266331658291 0.123259060669182
1.58291457286432 0.113978230916186
1.63316582914573 0.105130394600424
1.68341708542714 0.0967248359000725
1.73366834170854 0.0887668958717687
1.78391959798995 0.0812582357972331
1.83417085427136 0.0741971236389478
1.88442211055276 0.0675787375580061
1.93467336683417 0.0613954806631066
1.98492462311558 0.0556373014696322
2.03517587939699 0.0502920149370724
2.08542713567839 0.0453456194063216
2.1356783919598 0.0407826052600598
2.18592964824121 0.0365862516641933
2.23618090452261 0.0327389083015489
2.28643216080402 0.0292222595669734
2.33668341708543 0.0260175692432376
2.38693467336683 0.0231059042086428
2.43718592964824 0.0204683362305349
2.48743718592965 0.0180861213662637
2.53768844221106 0.0159408569183735
2.58793969849246 0.0140146162695652
2.63819095477387 0.0122900622523775
2.68844221105528 0.0107505399872646
2.73869346733668 0.00938015035081447
2.78894472361809 0.00816380541445067
2.8391959798995 0.0070872673253395
2.8894472361809 0.00613717218843178
2.93969849246231 0.00530104055534538
2.98994974874372 0.00456727613633011
3.04020100502513 0.00392515433036494
3.09045226130653 0.00336480212017405
3.14070351758794 0.00287717080828338
3.19095477386935 0.00245400298173195
3.24120603015075 0.00208779499104518
3.29145728643216 0.00177175611762511
3.34170854271357 0.00149976548650442
3.39195979899498 0.00126632766174431
3.44221105527638 0.00106652774249447
3.49246231155779 0.000895986661310447
3.5427135678392 0.000750817274733728
3.5929648241206 0.000627581730965922
3.64321608040201 0.000523250501893057
3.69346733668342 0.000435163377565355
3.74371859296483 0.000360992641007503
3.79396984924623 0.000298708570137427
3.84422110552764 0.000246547351573482
3.89447236180905 0.000202981437974616
3.94472361809045 0.000166692335884057
3.99497487437186 0.000136545774308338
4.04522613065327 0.000111569174841876
4.09547738693467 9.09313213657291e-05
4.14572864321608 7.39241104976095e-05
4.19597989949749 5.99462523304611e-05
4.2462311557789 4.84887838612555e-05
4.2964824120603 3.91222541992946e-05
4.34673366834171 3.14854405119586e-05
4.39698492462312 2.52754561211026e-05
4.44723618090452 2.02391166640527e-05
4.49748743718593 1.61654362953512e-05
4.54773869346734 1.28791331028225e-05
4.59798994974874 1.02350308754708e-05
4.64824120603015 8.11325277766367e-06
4.69849246231156 6.41511109267784e-06
4.74874371859297 5.0596057858091e-06
4.79899497487437 3.98045303347538e-06
4.84924623115578 3.12357293963309e-06
4.89949748743719 2.44497331818117e-06
4.94974874371859 1.90897359306981e-06
5 1.4867195147343e-06
};
\end{axis}

\end{tikzpicture}

%% file: figs/fig1_lula_map.tex
\begin{tikzpicture}

\definecolor{color0}{rgb}{0.00392156862745098,0.450980392156863,0.698039215686274}
\definecolor{color1}{rgb}{0.00784313725490196,0.619607843137255,0.450980392156863}

\begin{axis}[
width=0.3\textwidth,
height=0.15\textheight,
tick align=outside,
axis line style={white!15!black},
xtick={0},
xticklabels={$\theta$},
xmin=-4, xmax=4,
xticklabel pos=top,
yticklabel pos=right,
ymajorticks=false,
ymin=-4, ymax=4,
xtick style={draw=none},
ytick style={draw=none}
]
\path [draw=color0, draw opacity=0.5, very thick]
(axis cs:-1.79310344827586,-3.27643488833298)
--(axis cs:-1.51724137931034,-3.31312805851177)
--(axis cs:-1.24137931034483,-3.28852850791781)
--(axis cs:-0.96551724137931,-3.19238664901851)
--(axis cs:-0.930309671979427,-3.17241379310345)
--(axis cs:-0.689655172413793,-3.04716427753686)
--(axis cs:-0.508607986092716,-2.89655172413793)
--(axis cs:-0.413793103448276,-2.80351982565681)
--(axis cs:-0.24162274728389,-2.62068965517241)
--(axis cs:-0.137931034482759,-2.4859332695263)
--(axis cs:0.000100097217464956,-2.3448275862069)
--(axis cs:0.137931034482759,-2.21296913569497)
--(axis cs:0.354031182715395,-2.06896551724138)
--(axis cs:0.413793103448276,-2.03808646127213)
--(axis cs:0.689655172413793,-1.91607490083948)
--(axis cs:0.909680216208484,-1.79310344827586)
--(axis cs:0.96551724137931,-1.76511080974629)
--(axis cs:1.24137931034483,-1.57927794765901)
--(axis cs:1.31276927256711,-1.51724137931034)
--(axis cs:1.51724137931035,-1.31160569243642)
--(axis cs:1.57780687215793,-1.24137931034483)
--(axis cs:1.75929892596985,-0.96551724137931)
--(axis cs:1.79310344827586,-0.889209866507151)
--(axis cs:1.89172241808357,-0.689655172413793)
--(axis cs:1.97210446590956,-0.413793103448276)
--(axis cs:2.00796011403471,-0.137931034482759)
--(axis cs:2.00819624829296,0.137931034482759)
--(axis cs:1.97432509443882,0.413793103448276)
--(axis cs:1.90794740215231,0.689655172413793)
--(axis cs:1.83685551116735,0.96551724137931)
--(axis cs:1.82970629240971,1.24137931034483)
--(axis cs:1.8853805403975,1.51724137931035)
--(axis cs:1.92720401614514,1.79310344827586)
--(axis cs:1.91465222992882,2.06896551724138)
--(axis cs:1.81269908035946,2.3448275862069)
--(axis cs:1.79310344827586,2.37377409892412)
--(axis cs:1.61783385416228,2.62068965517241)
--(axis cs:1.51724137931035,2.71070619451806)
--(axis cs:1.24137931034483,2.84629031079624)
--(axis cs:0.96551724137931,2.87769788205918)
--(axis cs:0.689655172413793,2.83961104594105)
--(axis cs:0.413793103448276,2.70530431988312)
--(axis cs:0.304870375454948,2.62068965517241)
--(axis cs:0.137931034482759,2.45981183626611)
--(axis cs:0.0138077324029467,2.3448275862069)
--(axis cs:-0.137931034482759,2.17761210919314)
--(axis cs:-0.303223927471234,2.06896551724138)
--(axis cs:-0.413793103448276,2.00924123412157)
--(axis cs:-0.689655172413793,1.89928813001945)
--(axis cs:-0.894475227174118,1.79310344827586)
--(axis cs:-0.96551724137931,1.76011628671656)
--(axis cs:-1.24137931034483,1.57771839797306)
--(axis cs:-1.31139972756321,1.51724137931035)
--(axis cs:-1.51724137931034,1.31140748945763)
--(axis cs:-1.5777196387296,1.24137931034483)
--(axis cs:-1.75965135050171,0.96551724137931)
--(axis cs:-1.79310344827586,0.890889017111966)
--(axis cs:-1.89356045119927,0.689655172413793)
--(axis cs:-1.97812273440623,0.413793103448276)
--(axis cs:-2.02667236268448,0.137931034482759)
--(axis cs:-2.06253501901335,-0.137931034482759)
--(axis cs:-2.06896551724138,-0.170416237092235)
--(axis cs:-2.1352559031417,-0.413793103448276)
--(axis cs:-2.26327751751752,-0.689655172413793)
--(axis cs:-2.3448275862069,-0.815413258309013)
--(axis cs:-2.44511951184953,-0.96551724137931)
--(axis cs:-2.60563230767101,-1.24137931034483)
--(axis cs:-2.62068965517241,-1.27616714864714)
--(axis cs:-2.73827999084083,-1.51724137931034)
--(axis cs:-2.80291810745284,-1.79310344827586)
--(axis cs:-2.81180243608653,-2.06896551724138)
--(axis cs:-2.76551466026988,-2.3448275862069)
--(axis cs:-2.6396173186174,-2.62068965517241)
--(axis cs:-2.62068965517241,-2.64829504536459)
--(axis cs:-2.44064293085068,-2.89655172413793)
--(axis cs:-2.3448275862069,-2.98869584953107)
--(axis cs:-2.06896551724138,-3.16383715253106)
--(axis cs:-2.04451053452854,-3.17241379310345)
--(axis cs:-1.79310344827586,-3.27643488833298);

\path [draw=color0, draw opacity=0.5, very thick]
(axis cs:-1.79310344827586,-2.95446848142117)
--(axis cs:-1.51724137931034,-3.00385650840572)
--(axis cs:-1.24137931034483,-2.97218195804289)
--(axis cs:-1.06130503415657,-2.89655172413793)
--(axis cs:-0.96551724137931,-2.856814164249)
--(axis cs:-0.689655172413793,-2.6431938424986)
--(axis cs:-0.666863696052563,-2.62068965517241)
--(axis cs:-0.448110762066248,-2.3448275862069)
--(axis cs:-0.413793103448276,-2.28558018081907)
--(axis cs:-0.254088921389867,-2.06896551724138)
--(axis cs:-0.137931034482759,-1.92156126120137)
--(axis cs:0.0279576029934633,-1.79310344827586)
--(axis cs:0.137931034482759,-1.73102724347958)
--(axis cs:0.413793103448276,-1.61223801616897)
--(axis cs:0.607011419161926,-1.51724137931034)
--(axis cs:0.689655172413793,-1.48042291542836)
--(axis cs:0.96551724137931,-1.30770588154009)
--(axis cs:1.04444943301118,-1.24137931034483)
--(axis cs:1.24137931034483,-1.04241355063593)
--(axis cs:1.3049988512694,-0.96551724137931)
--(axis cs:1.46972792820229,-0.689655172413793)
--(axis cs:1.51724137931035,-0.555751482700863)
--(axis cs:1.57190650831126,-0.413793103448276)
--(axis cs:1.62309954954113,-0.137931034482759)
--(axis cs:1.62365485111336,0.137931034482759)
--(axis cs:1.57691555386385,0.413793103448276)
--(axis cs:1.51724137931035,0.612000087617144)
--(axis cs:1.49647080757373,0.689655172413793)
--(axis cs:1.44002023379708,0.96551724137931)
--(axis cs:1.47895550789975,1.24137931034483)
--(axis cs:1.51724137931035,1.32273652851886)
--(axis cs:1.59489424681535,1.51724137931034)
--(axis cs:1.67132424515262,1.79310344827586)
--(axis cs:1.66607078698349,2.06896551724138)
--(axis cs:1.54570537564538,2.3448275862069)
--(axis cs:1.51724137931034,2.37908381735431)
--(axis cs:1.24137931034483,2.59162052919388)
--(axis cs:1.08621442607229,2.62068965517241)
--(axis cs:0.965517241379311,2.6461051507494)
--(axis cs:0.86491066818542,2.62068965517241)
--(axis cs:0.689655172413793,2.58090207050331)
--(axis cs:0.413793103448276,2.38085745922365)
--(axis cs:0.376961984189357,2.3448275862069)
--(axis cs:0.184011955243715,2.06896551724138)
--(axis cs:0.137931034482759,1.97493747345891)
--(axis cs:-0.0285145342945377,1.79310344827586)
--(axis cs:-0.137931034482759,1.70837514043923)
--(axis cs:-0.413793103448276,1.59106173295148)
--(axis cs:-0.578323711335177,1.51724137931035)
--(axis cs:-0.689655172413793,1.47228630907515)
--(axis cs:-0.96551724137931,1.30500078097686)
--(axis cs:-1.04211730196948,1.24137931034483)
--(axis cs:-1.24137931034483,1.04214288645396)
--(axis cs:-1.30498745408292,0.965517241379311)
--(axis cs:-1.47088845261321,0.689655172413793)
--(axis cs:-1.51724137931034,0.564012341015265)
--(axis cs:-1.57724563951746,0.413793103448276)
--(axis cs:-1.6401292672412,0.137931034482759)
--(axis cs:-1.67404395923747,-0.137931034482759)
--(axis cs:-1.7126924796037,-0.413793103448276)
--(axis cs:-1.79310344827586,-0.644119729445402)
--(axis cs:-1.8127071557297,-0.689655172413793)
--(axis cs:-2.01834295543486,-0.96551724137931)
--(axis cs:-2.06896551724138,-1.01907690575319)
--(axis cs:-2.24511708429417,-1.24137931034483)
--(axis cs:-2.3448275862069,-1.4080762891207)
--(axis cs:-2.40631880596225,-1.51724137931034)
--(axis cs:-2.49295295208701,-1.79310344827586)
--(axis cs:-2.50438424226847,-2.06896551724138)
--(axis cs:-2.43979409462367,-2.3448275862069)
--(axis cs:-2.3448275862069,-2.5127722125608)
--(axis cs:-2.27491591347154,-2.62068965517241)
--(axis cs:-2.06896551724138,-2.81486808857844)
--(axis cs:-1.91133380375019,-2.89655172413793)
--(axis cs:-1.79310344827586,-2.95446848142117);

\path [draw=color0, draw opacity=0.5, very thick]
(axis cs:-1.79310344827586,-2.70640442953945)
--(axis cs:-1.51724137931034,-2.76869841493727)
--(axis cs:-1.24137931034483,-2.73125454800639)
--(axis cs:-1.02631566836436,-2.62068965517241)
--(axis cs:-0.96551724137931,-2.58181093709974)
--(axis cs:-0.739011275567136,-2.3448275862069)
--(axis cs:-0.689655172413793,-2.26216037206866)
--(axis cs:-0.572179098016369,-2.06896551724138)
--(axis cs:-0.436290980039865,-1.79310344827586)
--(axis cs:-0.413793103448276,-1.75498353645517)
--(axis cs:-0.14720012080641,-1.51724137931034)
--(axis cs:-0.137931034482759,-1.51239577403392)
--(axis cs:0.137931034482759,-1.41025340898767)
--(axis cs:0.413793103448276,-1.30746236033128)
--(axis cs:0.542315367886409,-1.24137931034483)
--(axis cs:0.689655172413793,-1.16460776691189)
--(axis cs:0.939445817792684,-0.96551724137931)
--(axis cs:0.96551724137931,-0.938997182855979)
--(axis cs:1.15860281659737,-0.689655172413793)
--(axis cs:1.24137931034483,-0.51008240451691)
--(axis cs:1.28430140770359,-0.413793103448276)
--(axis cs:1.34507422536108,-0.137931034482759)
--(axis cs:1.34579270501696,0.137931034482759)
--(axis cs:1.29063173332148,0.413793103448276)
--(axis cs:1.24137931034483,0.556203675907401)
--(axis cs:1.19390049310495,0.689655172413793)
--(axis cs:1.10728040015887,0.96551724137931)
--(axis cs:1.15807193068235,1.24137931034483)
--(axis cs:1.24137931034483,1.32947960811686)
--(axis cs:1.35062357792595,1.51724137931035)
--(axis cs:1.46855845468331,1.79310344827586)
--(axis cs:1.46594407692175,2.06896551724138)
--(axis cs:1.29152100245357,2.3448275862069)
--(axis cs:1.24137931034483,2.38557556032652)
--(axis cs:0.96551724137931,2.45979079275472)
--(axis cs:0.689655172413793,2.37277497243938)
--(axis cs:0.653169119471836,2.3448275862069)
--(axis cs:0.432427163471666,2.06896551724138)
--(axis cs:0.413793103448276,2.0125110617997)
--(axis cs:0.32124248448583,1.79310344827586)
--(axis cs:0.172277419068605,1.51724137931035)
--(axis cs:0.137931034482759,1.49226178931909)
--(axis cs:-0.137931034482759,1.38386419582581)
--(axis cs:-0.413793103448276,1.29229193074404)
--(axis cs:-0.520330018447086,1.24137931034483)
--(axis cs:-0.689655172413793,1.15939651715543)
--(axis cs:-0.937883126426256,0.96551724137931)
--(axis cs:-0.96551724137931,0.937914942353329)
--(axis cs:-1.15928156597644,0.689655172413793)
--(axis cs:-1.24137931034483,0.516380027415658)
--(axis cs:-1.28832605674593,0.413793103448276)
--(axis cs:-1.35850074785598,0.137931034482759)
--(axis cs:-1.38603585918986,-0.137931034482759)
--(axis cs:-1.39933364691845,-0.413793103448276)
--(axis cs:-1.44956163654171,-0.689655172413793)
--(axis cs:-1.51724137931034,-0.800008550464315)
--(axis cs:-1.63444097091414,-0.96551724137931)
--(axis cs:-1.79310344827586,-1.09839356257866)
--(axis cs:-1.92938209218357,-1.24137931034483)
--(axis cs:-2.06896551724138,-1.40139709288681)
--(axis cs:-2.14776994266405,-1.51724137931034)
--(axis cs:-2.25446793187459,-1.79310344827586)
--(axis cs:-2.26772200787134,-2.06896551724138)
--(axis cs:-2.18255350385096,-2.3448275862069)
--(axis cs:-2.06896551724138,-2.49994672806033)
--(axis cs:-1.93084011034146,-2.62068965517241)
--(axis cs:-1.79310344827586,-2.70640442953945);

\path [draw=color0, draw opacity=0.5, very thick]
(axis cs:-1.79310344827586,-2.45378160753173)
--(axis cs:-1.51724137931034,-2.54646844649189)
--(axis cs:-1.24137931034483,-2.49631284285627)
--(axis cs:-1.02223392188907,-2.3448275862069)
--(axis cs:-0.96551724137931,-2.27895015427686)
--(axis cs:-0.834820245830476,-2.06896551724138)
--(axis cs:-0.745421339953298,-1.79310344827586)
--(axis cs:-0.689655172413793,-1.55613495969418)
--(axis cs:-0.674834249906753,-1.51724137931034)
--(axis cs:-0.413793103448276,-1.27325523104974)
--(axis cs:-0.314727422574443,-1.24137931034483)
--(axis cs:-0.137931034482759,-1.20688901012526)
--(axis cs:0.137931034482759,-1.14598092450082)
--(axis cs:0.413793103448276,-1.04742527038539)
--(axis cs:0.55461450346162,-0.96551724137931)
--(axis cs:0.689655172413793,-0.873722145234141)
--(axis cs:0.87122303740957,-0.689655172413793)
--(axis cs:0.96551724137931,-0.541374608135499)
--(axis cs:1.03397466876609,-0.413793103448276)
--(axis cs:1.10594514364234,-0.137931034482759)
--(axis cs:1.10669897408627,0.137931034482759)
--(axis cs:1.04036737912109,0.413793103448276)
--(axis cs:0.96551724137931,0.583995377898168)
--(axis cs:0.908575463952755,0.689655172413793)
--(axis cs:0.72521490517965,0.96551724137931)
--(axis cs:0.689655172413793,1.05337414346566)
--(axis cs:0.413793103448276,1.1743718267958)
--(axis cs:0.137931034482759,1.16253842182783)
--(axis cs:-0.137931034482759,1.12254231752541)
--(axis cs:-0.413793103448276,1.03705672463358)
--(axis cs:-0.543386742029848,0.96551724137931)
--(axis cs:-0.689655172413793,0.870656073642667)
--(axis cs:-0.87082248455377,0.689655172413793)
--(axis cs:-0.965517241379311,0.544132299520599)
--(axis cs:-1.03684401739942,0.413793103448276)
--(axis cs:-1.11644926718978,0.137931034482759)
--(axis cs:-1.13846570669527,-0.137931034482759)
--(axis cs:-1.12402867098864,-0.413793103448276)
--(axis cs:-1.10182706159166,-0.689655172413793)
--(axis cs:-1.15110103216728,-0.96551724137931)
--(axis cs:-1.24137931034483,-1.04024810688242)
--(axis cs:-1.51724137931034,-1.21650371297578)
--(axis cs:-1.54729183756317,-1.24137931034483)
--(axis cs:-1.79310344827586,-1.41702321030946)
--(axis cs:-1.88238096421459,-1.51724137931034)
--(axis cs:-2.03225567711304,-1.79310344827586)
--(axis cs:-2.04950767532261,-2.06896551724138)
--(axis cs:-1.91390572869514,-2.3448275862069)
--(axis cs:-1.79310344827586,-2.45378160753173);

\path [draw=color0, draw opacity=0.5, very thick]
(axis cs:0.689655172413793,1.45948748444915)
--(axis cs:0.965517241379311,1.46045245537394)
--(axis cs:1.01904688082479,1.51724137931035)
--(axis cs:1.24137931034483,1.73149787003076)
--(axis cs:1.27204676113801,1.79310344827586)
--(axis cs:1.26995634858152,2.06896551724138)
--(axis cs:1.24137931034483,2.11139327532436)
--(axis cs:0.96551724137931,2.25406006225524)
--(axis cs:0.689655172413793,2.11109871193862)
--(axis cs:0.65493135398259,2.06896551724138)
--(axis cs:0.589167935458026,1.79310344827586)
--(axis cs:0.665457145900377,1.51724137931035)
--(axis cs:0.689655172413793,1.45948748444915);

\path [draw=color0, draw opacity=0.5, very thick]
(axis cs:-1.79310344827586,-2.07828912670865)
--(axis cs:-1.51724137931034,-2.28902689643115)
--(axis cs:-1.24137931034483,-2.20078947346718)
--(axis cs:-1.12701949434608,-2.06896551724138)
--(axis cs:-1.06416952530339,-1.79310344827586)
--(axis cs:-1.22876772493786,-1.51724137931034)
--(axis cs:-1.24137931034483,-1.5096005890993)
--(axis cs:-1.30756176296695,-1.51724137931034)
--(axis cs:-1.51724137931034,-1.54257899277154)
--(axis cs:-1.77560321219779,-1.79310344827586)
--(axis cs:-1.79310344827586,-1.97170734627616)
--(axis cs:-1.79822056906959,-2.06896551724138)
--(axis cs:-1.79310344827586,-2.07828912670865);

\path [draw=color0, draw opacity=0.5, very thick]
(axis cs:-0.689655172413793,-0.751300102141513)
--(axis cs:-0.413793103448276,-0.911363494509766)
--(axis cs:-0.137931034482759,-0.939357067285924)
--(axis cs:0.137931034482759,-0.901835793808681)
--(axis cs:0.413793103448276,-0.792135763541136)
--(axis cs:0.555847777858546,-0.689655172413793)
--(axis cs:0.689655172413793,-0.552145002869159)
--(axis cs:0.784586665392198,-0.413793103448276)
--(axis cs:0.876298535604607,-0.137931034482759)
--(axis cs:0.876996741987598,0.137931034482759)
--(axis cs:0.790114385699942,0.413793103448276)
--(axis cs:0.689655172413793,0.577864337279515)
--(axis cs:0.585055424247477,0.689655172413793)
--(axis cs:0.413793103448276,0.823460736854761)
--(axis cs:0.137931034482759,0.897676886540842)
--(axis cs:-0.137931034482759,0.882930150899313)
--(axis cs:-0.413793103448276,0.785335551960308)
--(axis cs:-0.551173964261148,0.689655172413793)
--(axis cs:-0.689655172413793,0.551991573470934)
--(axis cs:-0.78628556299623,0.413793103448276)
--(axis cs:-0.884780515506703,0.137931034482759)
--(axis cs:-0.902986564846436,-0.137931034482759)
--(axis cs:-0.854816723099702,-0.413793103448276)
--(axis cs:-0.731831627118353,-0.689655172413793)
--(axis cs:-0.689655172413793,-0.751300102141513);

\path [draw=color0, draw opacity=0.5, very thick]
(axis cs:-0.413793103448276,-0.56275794258712)
--(axis cs:-0.137931034482759,-0.673468732088576)
--(axis cs:0.137931034482759,-0.649062436957618)
--(axis cs:0.413793103448276,-0.499213835936704)
--(axis cs:0.497212151926637,-0.413793103448276)
--(axis cs:0.632408496864155,-0.137931034482759)
--(axis cs:0.632834757857343,0.137931034482759)
--(axis cs:0.499789778709588,0.413793103448276)
--(axis cs:0.413793103448276,0.504560081527847)
--(axis cs:0.137931034482759,0.639518853185503)
--(axis cs:-0.137931034482759,0.634518121853579)
--(axis cs:-0.413793103448276,0.495983390693888)
--(axis cs:-0.496404684463309,0.413793103448276)
--(axis cs:-0.639373864031863,0.137931034482759)
--(axis cs:-0.655316868901903,-0.137931034482759)
--(axis cs:-0.549497463644388,-0.413793103448276)
--(axis cs:-0.413793103448276,-0.56275794258712);

\path [draw=color0, draw opacity=0.5, very thick]
(axis cs:-0.137931034482759,-0.31322244429571)
--(axis cs:0.137931034482759,-0.297582287735333)
--(axis cs:0.290253665338834,-0.137931034482759)
--(axis cs:0.288418181781778,0.137931034482759)
--(axis cs:0.137931034482759,0.289507596752308)
--(axis cs:-0.137931034482759,0.28968866154605)
--(axis cs:-0.293274216344163,0.137931034482759)
--(axis cs:-0.307659472540151,-0.137931034482759)
--(axis cs:-0.137931034482759,-0.31322244429571);

\path [draw=color1, very thick]
(axis cs:-2.06896551724138,-3.34034289669793)
--(axis cs:-1.79310344827586,-3.41824195607456)
--(axis cs:-1.51724137931034,-3.44204945059624)
--(axis cs:-1.24137931034483,-3.42371002997107)
--(axis cs:-0.96551724137931,-3.35408436988601)
--(axis cs:-0.689655172413793,-3.19493836905649)
--(axis cs:-0.661197153341402,-3.17241379310345)
--(axis cs:-0.413793103448276,-2.95143360156188)
--(axis cs:-0.365186052358299,-2.89655172413793)
--(axis cs:-0.189677447630937,-2.62068965517241)
--(axis cs:-0.137931034482759,-2.45710524431766)
--(axis cs:-0.0912906192155744,-2.3448275862069)
--(axis cs:-0.04345472678758,-2.06896551724138)
--(axis cs:-0.0587972992001968,-1.79310344827586)
--(axis cs:-0.137931034482759,-1.53599048759376)
--(axis cs:-0.142003316343,-1.51724137931034)
--(axis cs:-0.260437872201391,-1.24137931034483)
--(axis cs:-0.413793103448276,-1.04839966095288)
--(axis cs:-0.490951445842261,-0.96551724137931)
--(axis cs:-0.689655172413793,-0.795571825783462)
--(axis cs:-0.904126278986817,-0.689655172413793)
--(axis cs:-0.96551724137931,-0.655074210016447)
--(axis cs:-1.24137931034483,-0.563351574586663)
--(axis cs:-1.51724137931034,-0.53919180375014)
--(axis cs:-1.79310344827586,-0.570555041640334)
--(axis cs:-2.06896551724138,-0.673176790788508)
--(axis cs:-2.09704134279148,-0.689655172413793)
--(axis cs:-2.3448275862069,-0.819607623668553)
--(axis cs:-2.51087590101299,-0.96551724137931)
--(axis cs:-2.62068965517241,-1.08903935544798)
--(axis cs:-2.73943037628774,-1.24137931034483)
--(axis cs:-2.86478113628233,-1.51724137931034)
--(axis cs:-2.89655172413793,-1.66923721888789)
--(axis cs:-2.93412519616351,-1.79310344827586)
--(axis cs:-2.95075553486136,-2.06896551724138)
--(axis cs:-2.8989045758698,-2.3448275862069)
--(axis cs:-2.89655172413793,-2.35057450132766)
--(axis cs:-2.8143229862457,-2.62068965517241)
--(axis cs:-2.62856523312447,-2.89655172413793)
--(axis cs:-2.62068965517241,-2.90561767462374)
--(axis cs:-2.3448275862069,-3.16730861458475)
--(axis cs:-2.3365908588285,-3.17241379310345)
--(axis cs:-2.06896551724138,-3.34034289669793);

\path [draw=color1, very thick]
(axis cs:-1.51724137931034,-3.17569860487465)
--(axis cs:-1.49099704079183,-3.17241379310345)
--(axis cs:-1.24137931034483,-3.15008083143636)
--(axis cs:-0.96551724137931,-3.05695357354162)
--(axis cs:-0.740451527195606,-2.89655172413793)
--(axis cs:-0.689655172413793,-2.852654337817)
--(axis cs:-0.494914330431155,-2.62068965517241)
--(axis cs:-0.413793103448276,-2.43797658669721)
--(axis cs:-0.369436427998302,-2.3448275862069)
--(axis cs:-0.313617820738789,-2.06896551724138)
--(axis cs:-0.331520717086748,-1.79310344827586)
--(axis cs:-0.413793103448276,-1.55991626742437)
--(axis cs:-0.426802470459948,-1.51724137931034)
--(axis cs:-0.596009494614365,-1.24137931034483)
--(axis cs:-0.689655172413793,-1.14361651525111)
--(axis cs:-0.943392723273748,-0.96551724137931)
--(axis cs:-0.96551724137931,-0.951120153623654)
--(axis cs:-1.24137931034483,-0.840138351816053)
--(axis cs:-1.51724137931034,-0.810905710536915)
--(axis cs:-1.79310344827586,-0.848854343775203)
--(axis cs:-2.05564784273709,-0.96551724137931)
--(axis cs:-2.06896551724138,-0.971104488766812)
--(axis cs:-2.3448275862069,-1.17939710876123)
--(axis cs:-2.40261524769111,-1.24137931034483)
--(axis cs:-2.57698040291445,-1.51724137931034)
--(axis cs:-2.62068965517241,-1.66738194489474)
--(axis cs:-2.66419651768645,-1.79310344827586)
--(axis cs:-2.68314488665745,-2.06896551724138)
--(axis cs:-2.62406664896654,-2.3448275862069)
--(axis cs:-2.62068965517241,-2.35205779262509)
--(axis cs:-2.50679221128482,-2.62068965517241)
--(axis cs:-2.3448275862069,-2.81890213706612)
--(axis cs:-2.26021768096605,-2.89655172413793)
--(axis cs:-2.06896551724138,-3.038573773395)
--(axis cs:-1.79310344827586,-3.14276705189983)
--(axis cs:-1.53784625453406,-3.17241379310345)
--(axis cs:-1.51724137931034,-3.17569860487465);

\path [draw=color1, very thick]
(axis cs:-1.79310344827586,-2.94871754092115)
--(axis cs:-1.51724137931034,-2.99648283347508)
--(axis cs:-1.24137931034483,-2.95968821022595)
--(axis cs:-1.10164945523643,-2.89655172413793)
--(axis cs:-0.96551724137931,-2.8324958925594)
--(axis cs:-0.731807652595347,-2.62068965517241)
--(axis cs:-0.689655172413793,-2.55865189542851)
--(axis cs:-0.566921165741308,-2.3448275862069)
--(axis cs:-0.507110283108459,-2.06896551724138)
--(axis cs:-0.52629363584382,-1.79310344827586)
--(axis cs:-0.633964383315731,-1.51724137931034)
--(axis cs:-0.689655172413793,-1.43900317070343)
--(axis cs:-0.869340577244514,-1.24137931034483)
--(axis cs:-0.96551724137931,-1.16498641468024)
--(axis cs:-1.24137931034483,-1.04107764879023)
--(axis cs:-1.51724137931034,-1.00844004113196)
--(axis cs:-1.79310344827586,-1.05080886408477)
--(axis cs:-2.06896551724138,-1.18944132084795)
--(axis cs:-2.13172111697182,-1.24137931034483)
--(axis cs:-2.3448275862069,-1.49028512180925)
--(axis cs:-2.36350333674669,-1.51724137931034)
--(axis cs:-2.47445633464836,-1.79310344827586)
--(axis cs:-2.49422447544583,-2.06896551724138)
--(axis cs:-2.43259030933223,-2.3448275862069)
--(axis cs:-2.3448275862069,-2.5019174855658)
--(axis cs:-2.26883523370082,-2.62068965517241)
--(axis cs:-2.06896551724138,-2.80942732881176)
--(axis cs:-1.90029407355284,-2.89655172413793)
--(axis cs:-1.79310344827586,-2.94871754092115);

\path [draw=color1, very thick]
(axis cs:-1.79310344827586,-2.77783251362097)
--(axis cs:-1.51724137931034,-2.83112177380711)
--(axis cs:-1.24137931034483,-2.79007192051542)
--(axis cs:-0.96551724137931,-2.63422604128926)
--(axis cs:-0.950581027639721,-2.62068965517241)
--(axis cs:-0.745958999262332,-2.3448275862069)
--(axis cs:-0.689655172413793,-2.13568395217772)
--(axis cs:-0.671987495895151,-2.06896551724138)
--(axis cs:-0.689655172413793,-1.87619003836836)
--(axis cs:-0.69682929016789,-1.79310344827586)
--(axis cs:-0.827032443414557,-1.51724137931034)
--(axis cs:-0.96551724137931,-1.36921639681235)
--(axis cs:-1.17898811958153,-1.24137931034483)
--(axis cs:-1.24137931034483,-1.20995992351879)
--(axis cs:-1.51724137931034,-1.16644311330778)
--(axis cs:-1.79310344827586,-1.22293487724485)
--(axis cs:-1.82710923853123,-1.24137931034483)
--(axis cs:-2.06896551724138,-1.40037159214434)
--(axis cs:-2.17390041633119,-1.51724137931034)
--(axis cs:-2.30370708188628,-1.79310344827586)
--(axis cs:-2.32683432035366,-2.06896551724138)
--(axis cs:-2.25472697995926,-2.3448275862069)
--(axis cs:-2.06896551724138,-2.60139090524296)
--(axis cs:-2.04702715995115,-2.62068965517241)
--(axis cs:-1.79310344827586,-2.77783251362097);

\path [draw=color1, very thick]
(axis cs:-1.51724137931034,-2.68207585566644)
--(axis cs:-1.24137931034483,-2.63076353905183)
--(axis cs:-1.22384682202284,-2.62068965517241)
--(axis cs:-0.96551724137931,-2.41367610950323)
--(axis cs:-0.913576867212359,-2.3448275862069)
--(axis cs:-0.823167381580076,-2.06896551724138)
--(axis cs:-0.852164730844307,-1.79310344827586)
--(axis cs:-0.96551724137931,-1.59031086586708)
--(axis cs:-1.02812530988731,-1.51724137931034)
--(axis cs:-1.24137931034483,-1.37272360259437)
--(axis cs:-1.51724137931034,-1.32074880483863)
--(axis cs:-1.79310344827586,-1.38822038630925)
--(axis cs:-1.96986926589737,-1.51724137931034)
--(axis cs:-2.06896551724138,-1.64420035928474)
--(axis cs:-2.14884466040147,-1.79310344827586)
--(axis cs:-2.1777537084857,-2.06896551724138)
--(axis cs:-2.0876195329927,-2.3448275862069)
--(axis cs:-2.06896551724138,-2.37059146707049)
--(axis cs:-1.79310344827586,-2.61483406560071)
--(axis cs:-1.77311021204005,-2.62068965517241)
--(axis cs:-1.51724137931034,-2.68207585566644);

\path [draw=color1, very thick]
(axis cs:-1.79310344827586,-2.43288314713429)
--(axis cs:-1.51724137931034,-2.5224575944717)
--(axis cs:-1.24137931034483,-2.45345648788772)
--(axis cs:-1.11211915767167,-2.3448275862069)
--(axis cs:-0.974624437709419,-2.06896551724138)
--(axis cs:-1.0187235981996,-1.79310344827586)
--(axis cs:-1.24137931034483,-1.54055428630513)
--(axis cs:-1.31232987509426,-1.51724137931034)
--(axis cs:-1.51724137931034,-1.47171901945321)
--(axis cs:-1.67812087617012,-1.51724137931034)
--(axis cs:-1.79310344827586,-1.56628704973335)
--(axis cs:-1.97876601533845,-1.79310344827586)
--(axis cs:-2.02049662721797,-2.06896551724138)
--(axis cs:-1.89038669472833,-2.3448275862069)
--(axis cs:-1.79310344827586,-2.43288314713429);

\path [draw=color1, very thick]
(axis cs:-1.51724137931034,-2.35543575056151)
--(axis cs:-1.4787356658798,-2.3448275862069)
--(axis cs:-1.24137931034483,-2.20967197623971)
--(axis cs:-1.16413632252328,-2.06896551724138)
--(axis cs:-1.21558534309516,-1.79310344827586)
--(axis cs:-1.24137931034483,-1.76384641844249)
--(axis cs:-1.51724137931034,-1.66315685605329)
--(axis cs:-1.79162161255643,-1.79310344827586)
--(axis cs:-1.79310344827586,-1.79651865542999)
--(axis cs:-1.84116335139556,-2.06896551724138)
--(axis cs:-1.79310344827586,-2.163257739103)
--(axis cs:-1.54747286790652,-2.3448275862069)
--(axis cs:-1.51724137931034,-2.35543575056151);

\addplot [very thick, red, opacity=0.75, mark=*, mark size=3, mark options={solid}]
table {%
-1.5 -2
};
\end{axis}

\end{tikzpicture}

%% file: figs/toy_reg_map.tex
\begin{tikzpicture}

\begin{axis}[
width=0.3\textwidth,
height=0.15\textheight,
axis line style={white!15.0!black},
xmajorticks=false,
xmin=-8, xmax=8,
ymajorticks=false,
ymin=-100, ymax=100
]
\addplot [only marks, mark=x, draw=black, fill=black, colormap={mymap}{[1pt]
 rgb(0pt)=(0.01060815,0.01808215,0.10018654);
  rgb(1pt)=(0.01428972,0.02048237,0.10374486);
  rgb(2pt)=(0.01831941,0.0229766,0.10738511);
  rgb(3pt)=(0.02275049,0.02554464,0.11108639);
  rgb(4pt)=(0.02759119,0.02818316,0.11483751);
  rgb(5pt)=(0.03285175,0.03088792,0.11863035);
  rgb(6pt)=(0.03853466,0.03365771,0.12245873);
  rgb(7pt)=(0.04447016,0.03648425,0.12631831);
  rgb(8pt)=(0.05032105,0.03936808,0.13020508);
  rgb(9pt)=(0.05611171,0.04224835,0.13411624);
  rgb(10pt)=(0.0618531,0.04504866,0.13804929);
  rgb(11pt)=(0.06755457,0.04778179,0.14200206);
  rgb(12pt)=(0.0732236,0.05045047,0.14597263);
  rgb(13pt)=(0.0788708,0.05305461,0.14995981);
  rgb(14pt)=(0.08450105,0.05559631,0.15396203);
  rgb(15pt)=(0.09011319,0.05808059,0.15797687);
  rgb(16pt)=(0.09572396,0.06050127,0.16200507);
  rgb(17pt)=(0.10132312,0.06286782,0.16604287);
  rgb(18pt)=(0.10692823,0.06517224,0.17009175);
  rgb(19pt)=(0.1125315,0.06742194,0.17414848);
  rgb(20pt)=(0.11813947,0.06961499,0.17821272);
  rgb(21pt)=(0.12375803,0.07174938,0.18228425);
  rgb(22pt)=(0.12938228,0.07383015,0.18636053);
  rgb(23pt)=(0.13501631,0.07585609,0.19044109);
  rgb(24pt)=(0.14066867,0.0778224,0.19452676);
  rgb(25pt)=(0.14633406,0.07973393,0.1986151);
  rgb(26pt)=(0.15201338,0.08159108,0.20270523);
  rgb(27pt)=(0.15770877,0.08339312,0.20679668);
  rgb(28pt)=(0.16342174,0.0851396,0.21088893);
  rgb(29pt)=(0.16915387,0.08682996,0.21498104);
  rgb(30pt)=(0.17489524,0.08848235,0.2190294);
  rgb(31pt)=(0.18065495,0.09009031,0.22303512);
  rgb(32pt)=(0.18643324,0.09165431,0.22699705);
  rgb(33pt)=(0.19223028,0.09317479,0.23091409);
  rgb(34pt)=(0.19804623,0.09465217,0.23478512);
  rgb(35pt)=(0.20388117,0.09608689,0.23860907);
  rgb(36pt)=(0.20973515,0.09747934,0.24238489);
  rgb(37pt)=(0.21560818,0.09882993,0.24611154);
  rgb(38pt)=(0.22150014,0.10013944,0.2497868);
  rgb(39pt)=(0.22741085,0.10140876,0.25340813);
  rgb(40pt)=(0.23334047,0.10263737,0.25697736);
  rgb(41pt)=(0.23928891,0.10382562,0.2604936);
  rgb(42pt)=(0.24525608,0.10497384,0.26395596);
  rgb(43pt)=(0.25124182,0.10608236,0.26736359);
  rgb(44pt)=(0.25724602,0.10715148,0.27071569);
  rgb(45pt)=(0.26326851,0.1081815,0.27401148);
  rgb(46pt)=(0.26930915,0.1091727,0.2772502);
  rgb(47pt)=(0.27536766,0.11012568,0.28043021);
  rgb(48pt)=(0.28144375,0.11104133,0.2835489);
  rgb(49pt)=(0.2875374,0.11191896,0.28660853);
  rgb(50pt)=(0.29364846,0.11275876,0.2896085);
  rgb(51pt)=(0.29977678,0.11356089,0.29254823);
  rgb(52pt)=(0.30592213,0.11432553,0.29542718);
  rgb(53pt)=(0.31208435,0.11505284,0.29824485);
  rgb(54pt)=(0.31826327,0.1157429,0.30100076);
  rgb(55pt)=(0.32445869,0.11639585,0.30369448);
  rgb(56pt)=(0.33067031,0.11701189,0.30632563);
  rgb(57pt)=(0.33689808,0.11759095,0.3088938);
  rgb(58pt)=(0.34314168,0.11813362,0.31139721);
  rgb(59pt)=(0.34940101,0.11863987,0.3138355);
  rgb(60pt)=(0.355676,0.11910909,0.31620996);
  rgb(61pt)=(0.36196644,0.1195413,0.31852037);
  rgb(62pt)=(0.36827206,0.11993653,0.32076656);
  rgb(63pt)=(0.37459292,0.12029443,0.32294825);
  rgb(64pt)=(0.38092887,0.12061482,0.32506528);
  rgb(65pt)=(0.38727975,0.12089756,0.3271175);
  rgb(66pt)=(0.39364518,0.12114272,0.32910494);
  rgb(67pt)=(0.40002537,0.12134964,0.33102734);
  rgb(68pt)=(0.40642019,0.12151801,0.33288464);
  rgb(69pt)=(0.41282936,0.12164769,0.33467689);
  rgb(70pt)=(0.41925278,0.12173833,0.33640407);
  rgb(71pt)=(0.42569057,0.12178916,0.33806605);
  rgb(72pt)=(0.43214263,0.12179973,0.33966284);
  rgb(73pt)=(0.43860848,0.12177004,0.34119475);
  rgb(74pt)=(0.44508855,0.12169883,0.34266151);
  rgb(75pt)=(0.45158266,0.12158557,0.34406324);
  rgb(76pt)=(0.45809049,0.12142996,0.34540024);
  rgb(77pt)=(0.46461238,0.12123063,0.34667231);
  rgb(78pt)=(0.47114798,0.12098721,0.34787978);
  rgb(79pt)=(0.47769736,0.12069864,0.34902273);
  rgb(80pt)=(0.48426077,0.12036349,0.35010104);
  rgb(81pt)=(0.49083761,0.11998161,0.35111537);
  rgb(82pt)=(0.49742847,0.11955087,0.35206533);
  rgb(83pt)=(0.50403286,0.11907081,0.35295152);
  rgb(84pt)=(0.51065109,0.11853959,0.35377385);
  rgb(85pt)=(0.51728314,0.1179558,0.35453252);
  rgb(86pt)=(0.52392883,0.11731817,0.35522789);
  rgb(87pt)=(0.53058853,0.11662445,0.35585982);
  rgb(88pt)=(0.53726173,0.11587369,0.35642903);
  rgb(89pt)=(0.54394898,0.11506307,0.35693521);
  rgb(90pt)=(0.5506426,0.11420757,0.35737863);
  rgb(91pt)=(0.55734473,0.11330456,0.35775059);
  rgb(92pt)=(0.56405586,0.11235265,0.35804813);
  rgb(93pt)=(0.57077365,0.11135597,0.35827146);
  rgb(94pt)=(0.5774991,0.11031233,0.35841679);
  rgb(95pt)=(0.58422945,0.10922707,0.35848469);
  rgb(96pt)=(0.59096382,0.10810205,0.35847347);
  rgb(97pt)=(0.59770215,0.10693774,0.35838029);
  rgb(98pt)=(0.60444226,0.10573912,0.35820487);
  rgb(99pt)=(0.61118304,0.10450943,0.35794557);
  rgb(100pt)=(0.61792306,0.10325288,0.35760108);
  rgb(101pt)=(0.62466162,0.10197244,0.35716891);
  rgb(102pt)=(0.63139686,0.10067417,0.35664819);
  rgb(103pt)=(0.63812122,0.09938212,0.35603757);
  rgb(104pt)=(0.64483795,0.0980891,0.35533555);
  rgb(105pt)=(0.65154562,0.09680192,0.35454107);
  rgb(106pt)=(0.65824241,0.09552918,0.3536529);
  rgb(107pt)=(0.66492652,0.09428017,0.3526697);
  rgb(108pt)=(0.67159578,0.09306598,0.35159077);
  rgb(109pt)=(0.67824099,0.09192342,0.3504148);
  rgb(110pt)=(0.684863,0.09085633,0.34914061);
  rgb(111pt)=(0.69146268,0.0898675,0.34776864);
  rgb(112pt)=(0.69803757,0.08897226,0.3462986);
  rgb(113pt)=(0.70457834,0.0882129,0.34473046);
  rgb(114pt)=(0.71108138,0.08761223,0.3430635);
  rgb(115pt)=(0.7175507,0.08716212,0.34129974);
  rgb(116pt)=(0.72398193,0.08688725,0.33943958);
  rgb(117pt)=(0.73035829,0.0868623,0.33748452);
  rgb(118pt)=(0.73669146,0.08704683,0.33543669);
  rgb(119pt)=(0.74297501,0.08747196,0.33329799);
  rgb(120pt)=(0.74919318,0.08820542,0.33107204);
  rgb(121pt)=(0.75535825,0.08919792,0.32876184);
  rgb(122pt)=(0.76145589,0.09050716,0.32637117);
  rgb(123pt)=(0.76748424,0.09213602,0.32390525);
  rgb(124pt)=(0.77344838,0.09405684,0.32136808);
  rgb(125pt)=(0.77932641,0.09634794,0.31876642);
  rgb(126pt)=(0.78513609,0.09892473,0.31610488);
  rgb(127pt)=(0.79085854,0.10184672,0.313391);
  rgb(128pt)=(0.7965014,0.10506637,0.31063031);
  rgb(129pt)=(0.80205987,0.10858333,0.30783);
  rgb(130pt)=(0.80752799,0.11239964,0.30499738);
  rgb(131pt)=(0.81291606,0.11645784,0.30213802);
  rgb(132pt)=(0.81820481,0.12080606,0.29926105);
  rgb(133pt)=(0.82341472,0.12535343,0.2963705);
  rgb(134pt)=(0.82852822,0.13014118,0.29347474);
  rgb(135pt)=(0.83355779,0.13511035,0.29057852);
  rgb(136pt)=(0.83850183,0.14025098,0.2876878);
  rgb(137pt)=(0.84335441,0.14556683,0.28480819);
  rgb(138pt)=(0.84813096,0.15099892,0.281943);
  rgb(139pt)=(0.85281737,0.15657772,0.27909826);
  rgb(140pt)=(0.85742602,0.1622583,0.27627462);
  rgb(141pt)=(0.86196552,0.16801239,0.27346473);
  rgb(142pt)=(0.86641628,0.17387796,0.27070818);
  rgb(143pt)=(0.87079129,0.17982114,0.26797378);
  rgb(144pt)=(0.87507281,0.18587368,0.26529697);
  rgb(145pt)=(0.87925878,0.19203259,0.26268136);
  rgb(146pt)=(0.8833417,0.19830556,0.26014181);
  rgb(147pt)=(0.88731387,0.20469941,0.25769539);
  rgb(148pt)=(0.89116859,0.21121788,0.2553592);
  rgb(149pt)=(0.89490337,0.21785614,0.25314362);
  rgb(150pt)=(0.8985026,0.22463251,0.25108745);
  rgb(151pt)=(0.90197527,0.23152063,0.24918223);
  rgb(152pt)=(0.90530097,0.23854541,0.24748098);
  rgb(153pt)=(0.90848638,0.24568473,0.24598324);
  rgb(154pt)=(0.911533,0.25292623,0.24470258);
  rgb(155pt)=(0.9144225,0.26028902,0.24369359);
  rgb(156pt)=(0.91717106,0.26773821,0.24294137);
  rgb(157pt)=(0.91978131,0.27526191,0.24245973);
  rgb(158pt)=(0.92223947,0.28287251,0.24229568);
  rgb(159pt)=(0.92456587,0.29053388,0.24242622);
  rgb(160pt)=(0.92676657,0.29823282,0.24285536);
  rgb(161pt)=(0.92882964,0.30598085,0.24362274);
  rgb(162pt)=(0.93078135,0.31373977,0.24468803);
  rgb(163pt)=(0.93262051,0.3215093,0.24606461);
  rgb(164pt)=(0.93435067,0.32928362,0.24775328);
  rgb(165pt)=(0.93599076,0.33703942,0.24972157);
  rgb(166pt)=(0.93752831,0.34479177,0.25199928);
  rgb(167pt)=(0.93899289,0.35250734,0.25452808);
  rgb(168pt)=(0.94036561,0.36020899,0.25734661);
  rgb(169pt)=(0.94167588,0.36786594,0.2603949);
  rgb(170pt)=(0.94291042,0.37549479,0.26369821);
  rgb(171pt)=(0.94408513,0.3830811,0.26722004);
  rgb(172pt)=(0.94520419,0.39062329,0.27094924);
  rgb(173pt)=(0.94625977,0.39813168,0.27489742);
  rgb(174pt)=(0.94727016,0.4055909,0.27902322);
  rgb(175pt)=(0.94823505,0.41300424,0.28332283);
  rgb(176pt)=(0.94914549,0.42038251,0.28780969);
  rgb(177pt)=(0.95001704,0.42771398,0.29244728);
  rgb(178pt)=(0.95085121,0.43500005,0.29722817);
  rgb(179pt)=(0.95165009,0.44224144,0.30214494);
  rgb(180pt)=(0.9524044,0.44944853,0.3072105);
  rgb(181pt)=(0.95312556,0.45661389,0.31239776);
  rgb(182pt)=(0.95381595,0.46373781,0.31769923);
  rgb(183pt)=(0.95447591,0.47082238,0.32310953);
  rgb(184pt)=(0.95510255,0.47787236,0.32862553);
  rgb(185pt)=(0.95569679,0.48489115,0.33421404);
  rgb(186pt)=(0.95626788,0.49187351,0.33985601);
  rgb(187pt)=(0.95681685,0.49882008,0.34555431);
  rgb(188pt)=(0.9573439,0.50573243,0.35130912);
  rgb(189pt)=(0.95784842,0.51261283,0.35711942);
  rgb(190pt)=(0.95833051,0.51946267,0.36298589);
  rgb(191pt)=(0.95879054,0.52628305,0.36890904);
  rgb(192pt)=(0.95922872,0.53307513,0.3748895);
  rgb(193pt)=(0.95964538,0.53983991,0.38092784);
  rgb(194pt)=(0.96004345,0.54657593,0.3870292);
  rgb(195pt)=(0.96042097,0.55328624,0.39319057);
  rgb(196pt)=(0.96077819,0.55997184,0.39941173);
  rgb(197pt)=(0.9611152,0.5666337,0.40569343);
  rgb(198pt)=(0.96143273,0.57327231,0.41203603);
  rgb(199pt)=(0.96173392,0.57988594,0.41844491);
  rgb(200pt)=(0.96201757,0.58647675,0.42491751);
  rgb(201pt)=(0.96228344,0.59304598,0.43145271);
  rgb(202pt)=(0.96253168,0.5995944,0.43805131);
  rgb(203pt)=(0.96276513,0.60612062,0.44471698);
  rgb(204pt)=(0.96298491,0.6126247,0.45145074);
  rgb(205pt)=(0.96318967,0.61910879,0.45824902);
  rgb(206pt)=(0.96337949,0.6255736,0.46511271);
  rgb(207pt)=(0.96355923,0.63201624,0.47204746);
  rgb(208pt)=(0.96372785,0.63843852,0.47905028);
  rgb(209pt)=(0.96388426,0.64484214,0.4861196);
  rgb(210pt)=(0.96403203,0.65122535,0.4932578);
  rgb(211pt)=(0.96417332,0.65758729,0.50046894);
  rgb(212pt)=(0.9643063,0.66393045,0.5077467);
  rgb(213pt)=(0.96443322,0.67025402,0.51509334);
  rgb(214pt)=(0.96455845,0.67655564,0.52251447);
  rgb(215pt)=(0.96467922,0.68283846,0.53000231);
  rgb(216pt)=(0.96479861,0.68910113,0.53756026);
  rgb(217pt)=(0.96492035,0.69534192,0.5451917);
  rgb(218pt)=(0.96504223,0.7015636,0.5528892);
  rgb(219pt)=(0.96516917,0.70776351,0.5606593);
  rgb(220pt)=(0.96530224,0.71394212,0.56849894);
  rgb(221pt)=(0.96544032,0.72010124,0.57640375);
  rgb(222pt)=(0.96559206,0.72623592,0.58438387);
  rgb(223pt)=(0.96575293,0.73235058,0.59242739);
  rgb(224pt)=(0.96592829,0.73844258,0.60053991);
  rgb(225pt)=(0.96612013,0.74451182,0.60871954);
  rgb(226pt)=(0.96632832,0.75055966,0.61696136);
  rgb(227pt)=(0.96656022,0.75658231,0.62527295);
  rgb(228pt)=(0.96681185,0.76258381,0.63364277);
  rgb(229pt)=(0.96709183,0.76855969,0.64207921);
  rgb(230pt)=(0.96739773,0.77451297,0.65057302);
  rgb(231pt)=(0.96773482,0.78044149,0.65912731);
  rgb(232pt)=(0.96810471,0.78634563,0.66773889);
  rgb(233pt)=(0.96850919,0.79222565,0.6764046);
  rgb(234pt)=(0.96893132,0.79809112,0.68512266);
  rgb(235pt)=(0.96935926,0.80395415,0.69383201);
  rgb(236pt)=(0.9698028,0.80981139,0.70252255);
  rgb(237pt)=(0.97025511,0.81566605,0.71120296);
  rgb(238pt)=(0.97071849,0.82151775,0.71987163);
  rgb(239pt)=(0.97120159,0.82736371,0.72851999);
  rgb(240pt)=(0.97169389,0.83320847,0.73716071);
  rgb(241pt)=(0.97220061,0.83905052,0.74578903);
  rgb(242pt)=(0.97272597,0.84488881,0.75440141);
  rgb(243pt)=(0.97327085,0.85072354,0.76299805);
  rgb(244pt)=(0.97383206,0.85655639,0.77158353);
  rgb(245pt)=(0.97441222,0.86238689,0.78015619);
  rgb(246pt)=(0.97501782,0.86821321,0.78871034);
  rgb(247pt)=(0.97564391,0.87403763,0.79725261);
  rgb(248pt)=(0.97628674,0.87986189,0.8057883);
  rgb(249pt)=(0.97696114,0.88568129,0.81430324);
  rgb(250pt)=(0.97765722,0.89149971,0.82280948);
  rgb(251pt)=(0.97837585,0.89731727,0.83130786);
  rgb(252pt)=(0.97912374,0.90313207,0.83979337);
  rgb(253pt)=(0.979891,0.90894778,0.84827858);
  rgb(254pt)=(0.98067764,0.91476465,0.85676611);
  rgb(255pt)=(0.98137749,0.92061729,0.86536915)
}]
table{%
x                      y
2.58711259403494 16.6145794509798
-2.25117804118928 -10.6677824273625
-3.74831192878677 -44.1956273977353
0.36740933375087 -2.8577950566267
-2.76435503635587 -24.9228154322306
-2.5042936589832 -9.65264724554914
-0.621218450998576 3.08668172492519
0.84441999611125 0.35054848004516
-0.0611671039393054 -3.59663459453611
1.15295254167363 10.3277348244988
3.30439583340008 38.7283422450371
2.16883287430782 4.85513927305084
-1.69311024479576 -3.40761647582771
-0.676469247568152 5.49845899654717
-1.34807651788659 -2.42627166338337
-1.06938644206434 -1.18842811868548
-1.50243729666088 -4.94728956459902
0.643566740983422 1.0703771720973
-3.09638617327673 -30.7320979397704
-1.39300112330007 -2.73461477959883
};
\addplot [very thick, black]
table {%
-8 -137.616027832031
-7.98398398398398 -137.261383056641
-7.96796796796797 -136.90673828125
-7.95195195195195 -136.552093505859
-7.93593593593594 -136.19743347168
-7.91991991991992 -135.842788696289
-7.9039039039039 -135.488143920898
-7.88788788788789 -135.133483886719
-7.87187187187187 -134.778839111328
-7.85585585585586 -134.424194335938
-7.83983983983984 -134.069534301758
-7.82382382382382 -133.714889526367
-7.80780780780781 -133.360244750977
-7.79179179179179 -133.005584716797
-7.77577577577578 -132.650939941406
-7.75975975975976 -132.296295166016
-7.74374374374374 -131.941650390625
-7.72772772772773 -131.586990356445
-7.71171171171171 -131.232345581055
-7.6956956956957 -130.877700805664
-7.67967967967968 -130.523040771484
-7.66366366366366 -130.168395996094
-7.64764764764765 -129.813751220703
-7.63163163163163 -129.459106445312
-7.61561561561562 -129.104446411133
-7.5995995995996 -128.749801635742
-7.58358358358358 -128.395141601562
-7.56756756756757 -128.040496826172
-7.55155155155155 -127.685852050781
-7.53553553553554 -127.331199645996
-7.51951951951952 -126.976547241211
-7.5035035035035 -126.62190246582
-7.48748748748749 -126.26725769043
-7.47147147147147 -125.912605285645
-7.45545545545546 -125.557952880859
-7.43943943943944 -125.203315734863
-7.42342342342342 -124.848655700684
-7.40740740740741 -124.494010925293
-7.39139139139139 -124.139358520508
-7.37537537537538 -123.784706115723
-7.35935935935936 -123.430061340332
-7.34334334334334 -123.075408935547
-7.32732732732733 -122.720764160156
-7.31131131131131 -122.366111755371
-7.2952952952953 -122.01146697998
-7.27927927927928 -121.656814575195
-7.26326326326326 -121.302169799805
-7.24724724724725 -120.94750213623
-7.23123123123123 -120.592864990234
-7.21521521521522 -120.238220214844
-7.1991991991992 -119.883567810059
-7.18318318318318 -119.528915405273
-7.16716716716717 -119.174278259277
-7.15115115115115 -118.819633483887
-7.13513513513514 -118.464973449707
-7.11911911911912 -118.110328674316
-7.1031031031031 -117.755676269531
-7.08708708708709 -117.401023864746
-7.07107107107107 -117.046371459961
-7.05505505505506 -116.69172668457
-7.03903903903904 -116.337074279785
-7.02302302302302 -115.982421875
-7.00700700700701 -115.627784729004
-6.99099099099099 -115.273139953613
-6.97497497497497 -114.918479919434
-6.95895895895896 -114.563842773438
-6.94294294294294 -114.209175109863
-6.92692692692693 -113.854530334473
-6.91091091091091 -113.499885559082
-6.89489489489489 -113.145240783691
-6.87887887887888 -112.790580749512
-6.86286286286286 -112.435943603516
-6.84684684684685 -112.081298828125
-6.83083083083083 -111.72664642334
-6.81481481481481 -111.37198638916
-6.7987987987988 -111.017333984375
-6.78278278278278 -110.66268157959
-6.76676676676677 -110.308036804199
-6.75075075075075 -109.953384399414
-6.73473473473473 -109.598731994629
-6.71871871871872 -109.244094848633
-6.7027027027027 -108.889442443848
-6.68668668668669 -108.534782409668
-6.67067067067067 -108.180145263672
-6.65465465465465 -107.825492858887
-6.63863863863864 -107.470848083496
-6.62262262262262 -107.116203308105
-6.60660660660661 -106.761543273926
-6.59059059059059 -106.406890869141
-6.57457457457457 -106.052253723145
-6.55855855855856 -105.697601318359
-6.54254254254254 -105.342948913574
-6.52652652652653 -104.988311767578
-6.51051051051051 -104.633651733398
-6.49449449449449 -104.278999328613
-6.47847847847848 -103.924369812012
-6.46246246246246 -103.569709777832
-6.44644644644645 -103.215057373047
-6.43043043043043 -102.860412597656
-6.41441441441441 -102.505760192871
-6.3983983983984 -102.151107788086
-6.38238238238238 -101.796455383301
-6.36636636636637 -101.44181060791
-6.35035035035035 -101.087158203125
-6.33433433433433 -100.73250579834
-6.31831831831832 -100.377868652344
-6.3023023023023 -100.023208618164
-6.28628628628629 -99.6685638427734
-6.27027027027027 -99.3139114379883
-6.25425425425425 -98.9592666625977
-6.23823823823824 -98.604621887207
-6.22222222222222 -98.2499771118164
-6.20620620620621 -97.8953170776367
-6.19019019019019 -97.5406723022461
-6.17417417417417 -97.1860275268555
-6.15815815815816 -96.8313751220703
-6.14214214214214 -96.4767227172852
-6.12612612612613 -96.1220855712891
-6.11011011011011 -95.7674331665039
-6.09409409409409 -95.4127731323242
-6.07807807807808 -95.0581283569336
-6.06206206206206 -94.703483581543
-6.04604604604605 -94.3488311767578
-6.03003003003003 -93.9941940307617
-6.01401401401401 -93.639533996582
-5.997997997998 -93.2848739624023
-5.98198198198198 -92.9302368164062
-5.96596596596597 -92.5755767822266
-5.94994994994995 -92.2209243774414
-5.93393393393393 -91.8662796020508
-5.91791791791792 -91.5116348266602
-5.9019019019019 -91.1569747924805
-5.88588588588589 -90.8023452758789
-5.86986986986987 -90.4476852416992
-5.85385385385385 -90.0930328369141
-5.83783783783784 -89.738395690918
-5.82182182182182 -89.3837356567383
-5.80580580580581 -89.0290908813477
-5.78978978978979 -88.674446105957
-5.77377377377377 -88.3197860717773
-5.75775775775776 -87.9651412963867
-5.74174174174174 -87.6104965209961
-5.72572572572573 -87.2558517456055
-5.70970970970971 -86.9011917114258
-5.69369369369369 -86.5465469360352
-5.67767767767768 -86.1919021606445
-5.66166166166166 -85.8372421264648
-5.64564564564565 -85.4825973510742
-5.62962962962963 -85.1279525756836
-5.61361361361361 -84.7733001708984
-5.5975975975976 -84.4186477661133
-5.58158158158158 -84.0640029907227
-5.56556556556557 -83.7093505859375
-5.54954954954955 -83.3546981811523
-5.53353353353353 -83.0000610351562
-5.51751751751752 -82.6454086303711
-5.5015015015015 -82.2907485961914
-5.48548548548549 -81.9361038208008
-5.46946946946947 -81.5814590454102
-5.45345345345345 -81.226806640625
-5.43743743743744 -80.8721694946289
-5.42142142142142 -80.5175094604492
-5.40540540540541 -80.1628570556641
-5.38938938938939 -79.808219909668
-5.37337337337337 -79.4535675048828
-5.35735735735736 -79.0989151000977
-5.34134134134134 -78.744270324707
-5.32532532532533 -78.3896255493164
-5.30930930930931 -78.0349655151367
-5.29329329329329 -77.6803283691406
-5.27727727727728 -77.3256759643555
-5.26126126126126 -76.9710235595703
-5.24524524524525 -76.6163787841797
-5.22922922922923 -76.2617263793945
-5.21321321321321 -75.9070816040039
-5.1971971971972 -75.5524291992188
-5.18118118118118 -75.1977767944336
-5.16516516516517 -74.8431243896484
-5.14914914914915 -74.4884796142578
-5.13313313313313 -74.1338272094727
-5.11711711711712 -73.7791748046875
-5.1011011011011 -73.4245376586914
-5.08508508508509 -73.0698776245117
-5.06906906906907 -72.7152328491211
-5.05305305305305 -72.3605880737305
-5.03703703703704 -72.0059356689453
-5.02102102102102 -71.6512832641602
-5.00500500500501 -71.296630859375
-4.98898898898899 -70.9419937133789
-4.97297297297297 -70.5873336791992
-4.95695695695696 -70.2326889038086
-4.94094094094094 -69.878044128418
-4.92492492492492 -69.5233917236328
-4.90890890890891 -69.1687393188477
-4.89289289289289 -68.814094543457
-4.87687687687688 -68.4594497680664
-4.86086086086086 -68.1047897338867
-4.84484484484484 -67.7501602172852
-4.82882882882883 -67.3955001831055
-4.81281281281281 -67.0408554077148
-4.7967967967968 -66.6862106323242
-4.78078078078078 -66.3315505981445
-4.76476476476476 -65.9769058227539
-4.74874874874875 -65.6222610473633
-4.73273273273273 -65.2676162719727
-4.71671671671672 -64.9129638671875
-4.7007007007007 -64.5583190917969
-4.68468468468468 -64.2036666870117
-4.66866866866867 -63.8490142822266
-4.65265265265265 -63.4943695068359
-4.63663663663664 -63.1397171020508
-4.62062062062062 -62.7850685119629
-4.6046046046046 -62.4304275512695
-4.58858858858859 -62.0757751464844
-4.57257257257257 -61.7211227416992
-4.55655655655656 -61.3664817810059
-4.54054054054054 -61.0118255615234
-4.52452452452452 -60.6571731567383
-4.50850850850851 -60.3025360107422
-4.49249249249249 -59.947883605957
-4.47647647647648 -59.5932312011719
-4.46046046046046 -59.238582611084
-4.44444444444444 -58.8839378356934
-4.42842842842843 -58.5292892456055
-4.41241241241241 -58.1746368408203
-4.3963963963964 -57.8199806213379
-4.38038038038038 -57.46533203125
-4.36436436436436 -57.1106796264648
-4.34834834834835 -56.7560348510742
-4.33233233233233 -56.4013824462891
-4.31631631631632 -56.0467338562012
-4.3003003003003 -55.6920890808105
-4.28428428428428 -55.3374328613281
-4.26826826826827 -54.9827880859375
-4.25225225225225 -54.6281433105469
-4.23623623623624 -54.2734909057617
-4.22022022022022 -53.9188461303711
-4.2042042042042 -53.5641975402832
-4.18818818818819 -53.209545135498
-4.17217217217217 -52.8548965454102
-4.15615615615616 -52.500244140625
-4.14014014014014 -52.1455993652344
-4.12412412412412 -51.7909507751465
-4.10810810810811 -51.4363021850586
-4.09209209209209 -51.0816535949707
-4.07607607607608 -50.7270088195801
-4.06006006006006 -50.3723526000977
-4.04404404404404 -50.017707824707
-4.02802802802803 -49.6630630493164
-4.01201201201201 -49.3084106445312
-3.995995995996 -48.9537658691406
-3.97997997997998 -48.59912109375
-3.96396396396396 -48.2444648742676
-3.94794794794795 -47.889820098877
-3.93193193193193 -47.5351715087891
-3.91591591591592 -47.1805191040039
-3.8998998998999 -46.8258743286133
-3.88388388388388 -46.4712295532227
-3.86786786786787 -46.1165771484375
-3.85185185185185 -45.7619323730469
-3.83583583583584 -45.4072875976562
-3.81981981981982 -45.0526351928711
-3.8038038038038 -44.6979904174805
-3.78778778778779 -44.3433303833008
-3.77177177177177 -43.9886856079102
-3.75575575575576 -43.6340370178223
-3.73973973973974 -43.2793846130371
-3.72372372372372 -42.9247398376465
-3.70770770770771 -42.5700950622559
-3.69169169169169 -42.2154388427734
-3.67567567567568 -41.8607940673828
-3.65965965965966 -41.5061492919922
-3.64364364364364 -41.151496887207
-3.62762762762763 -40.7968521118164
-3.61161161161161 -40.4422035217285
-3.5955955955956 -40.0875511169434
-3.57957957957958 -39.73291015625
-3.56356356356356 -39.3782577514648
-3.54754754754755 -39.0236053466797
-3.53153153153153 -38.6689605712891
-3.51551551551552 -38.3143119812012
-3.4994994994995 -37.9596633911133
-3.48348348348348 -37.6050186157227
-3.46746746746747 -37.2503700256348
-3.45145145145145 -36.8957138061523
-3.43543543543544 -36.541072845459
-3.41941941941942 -36.1864242553711
-3.4034034034034 -35.8317718505859
-3.38738738738739 -35.477123260498
-3.37137137137137 -35.1224708557129
-3.35535535535536 -34.7678260803223
-3.33933933933934 -34.4131774902344
-3.32332332332332 -34.0585289001465
-3.30730730730731 -33.7038803100586
-3.29129129129129 -33.349235534668
-3.27527527527528 -32.9945831298828
-3.25925925925926 -32.6399383544922
-3.24324324324324 -32.285285949707
-3.22722722722723 -31.9306373596191
-3.21121121121121 -31.5759906768799
-3.1951951951952 -31.2213439941406
-3.17917917917918 -30.8666934967041
-3.16316316316316 -30.5120449066162
-3.14714714714715 -30.1574001312256
-3.13113113113113 -29.8027477264404
-3.11511511511512 -29.4481029510498
-3.0990990990991 -29.0934505462646
-3.08308308308308 -28.7388038635254
-3.06706706706707 -28.3841590881348
-3.05105105105105 -28.029504776001
-3.03503503503504 -27.6748600006104
-3.01901901901902 -27.3202152252197
-3.003003003003 -26.9655590057373
-2.98698698698699 -26.6109104156494
-2.97097097097097 -26.2562637329102
-2.95495495495495 -25.9016132354736
-2.93893893893894 -25.5469646453857
-2.92292292292292 -25.1923198699951
-2.90690690690691 -24.83766746521
-2.89089089089089 -24.4830207824707
-2.87487487487487 -24.1283760070801
-2.85885885885886 -23.7737236022949
-2.84284284284284 -23.4190769195557
-2.82682682682683 -23.064432144165
-2.81081081081081 -22.7097797393799
-2.79479479479479 -22.3551330566406
-2.77877877877878 -22.0004863739014
-2.76276276276276 -21.6458358764648
-2.74674674674675 -21.2911891937256
-2.73073073073073 -20.9365367889404
-2.71471471471471 -20.5818901062012
-2.6986986986987 -20.2272434234619
-2.68268268268268 -19.8725910186768
-2.66666666666667 -19.5179462432861
-2.65065065065065 -19.1632976531982
-2.63463463463463 -18.8086490631104
-2.61861861861862 -18.4540004730225
-2.6026026026026 -18.0993556976318
-2.58658658658659 -17.7446975708008
-2.57057057057057 -17.3900508880615
-2.55455455455455 -17.0354061126709
-2.53853853853854 -16.6807537078857
-2.52252252252252 -16.3261070251465
-2.50650650650651 -15.9714584350586
-2.49049049049049 -15.6168098449707
-2.47447447447447 -15.2621631622314
-2.45845845845846 -14.9075145721436
-2.44244244244244 -14.5528650283813
-2.42642642642643 -14.1982192993164
-2.41041041041041 -13.8435707092285
-2.39439439439439 -13.488920211792
-2.37837837837838 -13.1342735290527
-2.36236236236236 -12.7796249389648
-2.34634634634635 -12.424976348877
-2.33033033033033 -12.0703268051147
-2.31431431431431 -11.7156810760498
-2.2982982982983 -11.4174938201904
-2.28228228228228 -11.1398906707764
-2.26626626626627 -10.8622922897339
-2.25025025025025 -10.5846920013428
-2.23423423423423 -10.3070888519287
-2.21821821821822 -10.0294904708862
-2.2022022022022 -9.75189018249512
-2.18618618618619 -9.47428321838379
-2.17017017017017 -9.19668197631836
-2.15415415415415 -8.91908359527588
-2.13813813813814 -8.64148235321045
-2.12212212212212 -8.36388206481934
-2.10610610610611 -8.08628273010254
-2.09009009009009 -7.80868148803711
-2.07407407407407 -7.53108024597168
-2.05805805805806 -7.25348091125488
-2.04204204204204 -6.97587871551514
-2.02602602602603 -6.69827890396118
-2.01001001001001 -6.42067956924438
-1.99399399399399 -6.16137409210205
-1.97797797797798 -6.06929779052734
-1.96196196196196 -5.97855949401855
-1.94594594594595 -5.88782215118408
-1.92992992992993 -5.79708385467529
-1.91391391391391 -5.7063455581665
-1.8978978978979 -5.61560678482056
-1.88188188188188 -5.52486896514893
-1.86586586586587 -5.43413114547729
-1.84984984984985 -5.34339427947998
-1.83383383383383 -5.25265598297119
-1.81781781781782 -5.16191864013672
-1.8018018018018 -5.07117938995361
-1.78578578578579 -4.98044204711914
-1.76976976976977 -4.88970470428467
-1.75375375375375 -4.79896640777588
-1.73773773773774 -4.70822858810425
-1.72172172172172 -4.61749076843262
-1.70570570570571 -4.52675247192383
-1.68968968968969 -4.43601512908936
-1.67367367367367 -4.34527683258057
-1.65765765765766 -4.25453901290894
-1.64164164164164 -4.16380071640015
-1.62562562562563 -4.07306385040283
-1.60960960960961 -3.98232579231262
-1.59359359359359 -3.89158797264099
-1.57757757757758 -3.8008496761322
-1.56156156156156 -3.71011161804199
-1.54554554554555 -3.61937403678894
-1.52952952952953 -3.52863621711731
-1.51351351351351 -3.43789839744568
-1.4974974974975 -3.34715962409973
-1.48148148148148 -3.25642204284668
-1.46546546546547 -3.16568446159363
-1.44944944944945 -3.07494592666626
-1.43343343343343 -2.98420858383179
-1.41741741741742 -2.893470287323
-1.4014014014014 -2.80273246765137
-1.38538538538539 -2.71199488639832
-1.36936936936937 -2.62125706672668
-1.35335335335335 -2.53051900863647
-1.33733733733734 -2.43978095054626
-1.32132132132132 -2.34904336929321
-1.30530530530531 -2.25830578804016
-1.28928928928929 -2.16756701469421
-1.27327327327327 -2.076828956604
-1.25725725725726 -1.9860907793045
-1.24124124124124 -1.89535319805145
-1.22522522522523 -1.80461537837982
-1.20920920920921 -1.7138774394989
-1.19319319319319 -1.62313950061798
-1.17717717717718 -1.53240191936493
-1.16116116116116 -1.44166362285614
-1.14514514514515 -1.35092604160309
-1.12912912912913 -1.26018822193146
-1.11311311311311 -1.16945028305054
-1.0970970970971 -1.07871174812317
-1.08108108108108 -0.987973749637604
-1.06506506506507 -0.897235929965973
-1.04904904904905 -0.806498110294342
-1.03303303303303 -0.715760290622711
-1.01701701701702 -0.625022351741791
-1.001001001001 -0.53428441286087
-0.984984984984985 -0.443546831607819
-0.968968968968969 -0.352808892726898
-0.952952952952953 -0.262070924043655
-0.936936936936937 -0.171332716941833
-0.920920920920921 -0.0805947706103325
-0.904904904904905 0.0101429298520088
-0.888888888888889 0.100880779325962
-0.872872872872873 0.191618621349335
-0.856856856856857 0.282356679439545
-0.840840840840841 0.373094767332077
-0.824824824824825 0.463832706212997
-0.808808808808809 0.554570436477661
-0.792792792792793 0.645308434963226
-0.776776776776777 0.736046135425568
-0.76076076076076 0.826784133911133
-0.744744744744745 0.917522311210632
-0.728728728728729 1.00826013088226
-0.712712712712713 1.09899806976318
-0.696696696696697 1.18973588943481
-0.68068068068068 1.26108396053314
-0.664664664664665 1.24587500095367
-0.648648648648649 1.22689568996429
-0.632632632632633 1.20791637897491
-0.616616616616617 1.18893718719482
-0.6006006006006 1.16995775699615
-0.584584584584585 1.15097868442535
-0.568568568568569 1.13199937343597
-0.552552552552553 1.11302006244659
-0.536536536536537 1.09404075145721
-0.52052052052052 1.07506144046783
-0.504504504504505 1.05608224868774
-0.488488488488488 1.03710293769836
-0.472472472472472 1.01812362670898
-0.456456456456457 0.99914425611496
-0.44044044044044 0.980165004730225
-0.424424424424425 0.96118575334549
-0.408408408408408 0.94220644235611
-0.392392392392392 0.92322713136673
-0.376376376376377 0.904247879981995
-0.36036036036036 0.88526862859726
-0.344344344344345 0.86628931760788
-0.328328328328328 0.8473100066185
-0.312312312312312 0.828330755233765
-0.296296296296297 0.80935150384903
-0.28028028028028 0.79037219285965
-0.264264264264265 0.77139288187027
-0.248248248248248 0.752413630485535
-0.232232232232232 0.733434319496155
-0.216216216216217 0.71445506811142
-0.2002002002002 0.695475816726685
-0.184184184184184 0.676496565341949
-0.168168168168168 0.65751725435257
-0.152152152152152 0.63853794336319
-0.136136136136137 0.619558691978455
-0.12012012012012 0.600579440593719
-0.104104104104104 0.58160012960434
-0.0880880880880879 0.562620878219604
-0.0720720720720722 0.543641567230225
-0.0560560560560557 0.52466231584549
-0.04004004004004 0.505682945251465
-0.0240240240240244 0.48670369386673
-0.00800800800800783 0.467724442481995
0.00800800800800872 0.448745161294937
0.0240240240240244 0.429765909910202
0.04004004004004 0.410786658525467
0.0560560560560557 0.391807287931442
0.0720720720720713 0.37282806634903
0.0880880880880888 0.353848725557327
0.104104104104104 0.334869503974915
0.12012012012012 0.315890252590179
0.136136136136136 0.2969109416008
0.152152152152151 0.277931690216064
0.168168168168169 0.25895231962204
0.184184184184184 0.239973068237305
0.2002002002002 0.220993846654892
0.216216216216216 0.202014535665512
0.232232232232231 0.183035284280777
0.248248248248249 0.164055913686752
0.264264264264265 0.145076662302017
0.28028028028028 0.126097440719604
0.296296296296296 0.107118122279644
0.312312312312311 0.0881388783454895
0.328328328328329 0.0691595673561096
0.344344344344345 0.0501803271472454
0.36036036036036 0.0312010832130909
0.376376376376376 0.0302140768617392
0.392392392392392 0.0708837509155273
0.408408408408409 0.151372313499451
0.424424424424425 0.23204830288887
0.44044044044044 0.312725394964218
0.456456456456456 0.393401980400085
0.472472472472472 0.474078506231308
0.488488488488489 0.554754495620728
0.504504504504505 0.635431051254272
0.52052052052052 0.716107606887817
0.536536536536536 0.796783566474915
0.552552552552552 0.877460718154907
0.568568568568569 0.958137273788452
0.584584584584585 1.03881335258484
0.6006006006006 1.11948990821838
0.616616616616616 1.20016646385193
0.632632632632632 1.28084290027618
0.648648648648649 1.36151897907257
0.664664664664665 1.44219553470612
0.68068068068068 1.52287209033966
0.696696696696696 1.60354804992676
0.712712712712712 1.68422448635101
0.728728728728729 1.76490116119385
0.744744744744745 1.84557712078094
0.76076076076076 1.92625367641449
0.776776776776776 2.00693011283875
0.792792792792794 2.08760666847229
0.808808808808809 2.16828274726868
0.824824824824825 2.24895930290222
0.840840840840841 2.32963585853577
0.856856856856856 2.41031241416931
0.872872872872874 2.49098896980286
0.888888888888889 2.5716655254364
0.904904904904905 2.65234160423279
0.920920920920921 2.73301792144775
0.936936936936936 2.81369471549988
0.952952952952954 2.89437174797058
0.968968968968969 2.97504758834839
0.984984984984985 3.05572414398193
1.001001001001 3.1364004611969
1.01701701701702 3.21707677841187
1.03303303303303 3.29775309562683
1.04904904904905 3.37842965126038
1.06506506506507 3.45910573005676
1.08108108108108 3.53978252410889
1.0970970970971 3.62045884132385
1.11311311311311 3.70113492012024
1.12912912912913 3.78181147575378
1.14514514514515 3.86248826980591
1.16116116116116 3.94316458702087
1.17717717717718 4.02384042739868
1.19319319319319 4.10451745986938
1.20920920920921 4.18519401550293
1.22522522522523 4.26587009429932
1.24124124124124 4.3465461730957
1.25725725725726 4.42722272872925
1.27327327327327 4.50789880752563
1.28928928928929 4.58857536315918
1.30530530530531 4.6692533493042
1.32132132132132 4.74992990493774
1.33733733733734 4.83060550689697
1.35335335335335 4.91128253936768
1.36936936936937 4.99195861816406
1.38538538538539 5.07263469696045
1.4014014014014 5.15331172943115
1.41741741741742 5.23398780822754
1.43343343343343 5.31466388702393
1.44944944944945 5.39534091949463
1.46546546546547 5.47601699829102
1.48148148148148 5.5566930770874
1.4974974974975 5.63737010955811
1.51351351351351 5.71804618835449
1.52952952952953 5.79872226715088
1.54554554554555 5.87939929962158
1.56156156156156 5.96007537841797
1.57757757757758 6.0407509803772
1.59359359359359 6.12142848968506
1.60960960960961 6.20210456848145
1.62562562562563 6.28277969360352
1.64164164164164 6.36345720291138
1.65765765765766 6.44413375854492
1.67367367367367 6.52480936050415
1.68968968968969 6.60548686981201
1.70570570570571 6.68616390228271
1.72172172172172 6.76684093475342
1.73773773773774 6.84751653671265
1.75375375375375 6.92819261550903
1.76976976976977 7.00887012481689
1.78578578578579 7.08954572677612
1.8018018018018 7.17022132873535
1.81781781781782 7.25089883804321
1.83383383383383 7.33157444000244
1.84984984984985 7.41225099563599
1.86586586586587 7.49292755126953
1.88188188188188 7.57360363006592
1.8978978978979 7.6542797088623
1.91391391391391 7.73495674133301
1.92992992992993 7.81563329696655
1.94594594594595 7.89630889892578
1.96196196196196 7.97698593139648
1.97797797797798 8.05766201019287
1.99399399399399 8.13833808898926
2.01001001001001 8.21901512145996
2.02602602602603 8.29969120025635
2.04204204204204 8.38036727905273
2.05805805805806 8.46104431152344
2.07407407407407 8.54172039031982
2.09009009009009 8.62239646911621
2.10610610610611 8.70307350158691
2.12212212212212 8.78374862670898
2.13813813813814 8.86500835418701
2.15415415415415 9.03757953643799
2.17017017017017 9.27993583679199
2.18618618618619 9.53918647766113
2.2022022022022 9.83084487915039
2.21821821821822 10.181339263916
2.23423423423423 10.5457210540771
2.25025025025025 10.9100971221924
2.26626626626627 11.2744731903076
2.28228228228228 11.6388559341431
2.2982982982983 12.0032320022583
2.31431431431431 12.3676090240479
2.33033033033033 12.731990814209
2.34634634634635 13.0963668823242
2.36236236236236 13.4607429504395
2.37837837837838 13.8251256942749
2.39439439439439 14.1895017623901
2.41041041041041 14.5538787841797
2.42642642642643 14.9182596206665
2.44244244244244 15.2826366424561
2.45845845845846 15.6470184326172
2.47447447447447 16.0113945007324
2.49049049049049 16.3757705688477
2.50650650650651 16.7401542663574
2.52252252252252 17.104528427124
2.53853853853854 17.4689064025879
2.55455455455455 17.833288192749
2.57057057057057 18.1976642608643
2.58658658658659 18.5620403289795
2.6026026026026 18.9264278411865
2.61861861861862 19.2908039093018
2.63463463463463 19.655179977417
2.65065065065065 20.0195617675781
2.66666666666667 20.383939743042
2.68268268268268 20.7483158111572
2.6986986986987 21.1126976013184
2.71471471471471 21.4770736694336
2.73073073073073 21.8414516448975
2.74674674674675 22.20583152771
2.76276276276276 22.5702075958252
2.77877877877878 22.9345836639404
2.79479479479479 23.2989654541016
2.81081081081081 23.6633434295654
2.82682682682683 24.0277252197266
2.84284284284284 24.3921031951904
2.85885885885886 24.756477355957
2.87487487487487 25.1208591461182
2.89089089089089 25.4852352142334
2.90690690690691 25.8496131896973
2.92292292292292 26.2139949798584
2.93893893893894 26.5783710479736
2.95495495495495 26.9427452087402
2.97097097097097 27.30712890625
2.98698698698699 27.6715049743652
3.003003003003 28.035888671875
3.01901901901902 28.4002704620361
3.03503503503504 28.7646446228027
3.05105105105105 29.1290225982666
3.06706706706707 29.4934024810791
3.08308308308308 29.857780456543
3.0990990990991 30.2221565246582
3.11511511511512 30.586540222168
3.13113113113113 30.9509143829346
3.14714714714715 31.3152961730957
3.16316316316316 31.6796741485596
3.17917917917918 32.0440483093262
3.1951951951952 32.4084281921387
3.21121121121121 32.7728042602539
3.22722722722723 33.1371841430664
3.24324324324324 33.5015640258789
3.25925925925926 33.8659400939941
3.27527527527528 34.2303161621094
3.29129129129129 34.5946960449219
3.30730730730731 34.9590682983398
3.32332332332332 35.3234519958496
3.33933933933934 35.6878356933594
3.35535535535536 36.0522155761719
3.37137137137137 36.4165840148926
3.38738738738739 36.7809677124023
3.4034034034034 37.1453514099121
3.41941941941942 37.5097312927246
3.43543543543544 37.8741149902344
3.45145145145145 38.2384948730469
3.46746746746747 38.6028671264648
3.48348348348348 38.9672470092773
3.4994994994995 39.3316307067871
3.51551551551552 39.6959991455078
3.53153153153153 40.0603828430176
3.54754754754755 40.4247665405273
3.56356356356356 40.7891387939453
3.57957957957958 41.1535186767578
3.5955955955956 41.5178985595703
3.61161161161161 41.8822708129883
3.62762762762763 42.2466506958008
3.64364364364364 42.6110382080078
3.65965965965966 42.9754028320312
3.67567567567568 43.339786529541
3.69169169169169 43.704174041748
3.70770770770771 44.0685386657715
3.72372372372372 44.4329223632812
3.73973973973974 44.7973022460938
3.75575575575576 45.1616744995117
3.77177177177177 45.5260543823242
3.78778778778779 45.8904418945312
3.8038038038038 46.254810333252
3.81981981981982 46.6191902160645
3.83583583583584 46.9835739135742
3.85185185185185 47.3479461669922
3.86786786786787 47.7123260498047
3.88388388388388 48.0767059326172
3.8998998998999 48.4410781860352
3.91591591591592 48.8054618835449
3.93193193193193 49.1698417663574
3.94794794794795 49.5342178344727
3.96396396396396 49.8985977172852
3.97997997997998 50.2629776000977
3.995995995996 50.6273574829102
4.01201201201201 50.9917297363281
4.02802802802803 51.3561134338379
4.04404404404404 51.7204933166504
4.06006006006006 52.0848617553711
4.07607607607608 52.4492492675781
4.09209209209209 52.8136291503906
4.10810810810811 53.1779975891113
4.12412412412412 53.5423812866211
4.14014014014014 53.9067649841309
4.15615615615616 54.2711334228516
4.17217217217217 54.6355133056641
4.18818818818819 54.9998970031738
4.2042042042042 55.3642692565918
4.22022022022022 55.7286529541016
4.23623623623624 56.0930328369141
4.25225225225225 56.457405090332
4.26826826826827 56.8217849731445
4.28428428428428 57.186164855957
4.3003003003003 57.5505409240723
4.31631631631632 57.9149208068848
4.33233233233233 58.2793006896973
4.34834834834835 58.6436767578125
4.36436436436436 59.008056640625
4.38038038038038 59.3724365234375
4.3963963963964 59.7368087768555
4.41241241241241 60.1012001037598
4.42842842842843 60.4655799865723
4.44444444444444 60.8299560546875
4.46046046046046 61.1943359375
4.47647647647648 61.5587158203125
4.49249249249249 61.9230880737305
4.50850850850851 62.287467956543
4.52452452452452 62.6518516540527
4.54054054054054 63.0162239074707
4.55655655655656 63.3806037902832
4.57257257257257 63.744987487793
4.58858858858859 64.1093597412109
4.6046046046046 64.4737396240234
4.62062062062062 64.8381271362305
4.63663663663664 65.2024917602539
4.65265265265265 65.5668716430664
4.66866866866867 65.9312591552734
4.68468468468468 66.2956390380859
4.7007007007007 66.6600112915039
4.71671671671672 67.0243911743164
4.73273273273273 67.3887710571289
4.74874874874875 67.7531433105469
4.76476476476476 68.1175308227539
4.78078078078078 68.4819030761719
4.7967967967968 68.8462829589844
4.81281281281281 69.2106552124023
4.82882882882883 69.5750427246094
4.84484484484484 69.9394149780273
4.86086086086086 70.3037948608398
4.87687687687688 70.6681747436523
4.89289289289289 71.0325469970703
4.90890890890891 71.3969268798828
4.92492492492492 71.7613143920898
4.94094094094094 72.1256790161133
4.95695695695696 72.4900588989258
4.97297297297297 72.8544464111328
4.98898898898899 73.2188186645508
5.00500500500501 73.5831985473633
5.02102102102102 73.9475784301758
5.03703703703704 74.3119506835938
5.05305305305305 74.6763305664062
5.06906906906907 75.0407180786133
5.08508508508509 75.4050903320312
5.1011011011011 75.7694702148438
5.11711711711712 76.1338500976562
5.13313313313313 76.4982223510742
5.14914914914915 76.8626022338867
5.16516516516517 77.2269821166992
5.18118118118118 77.5913543701172
5.1971971971972 77.9557418823242
5.21321321321321 78.3201293945312
5.22922922922923 78.6845016479492
5.24524524524525 79.0488815307617
5.26126126126126 79.4132614135742
5.27727727727728 79.7776412963867
5.29329329329329 80.1420211791992
5.30930930930931 80.5064010620117
5.32532532532533 80.8707656860352
5.34134134134134 81.2351531982422
5.35735735735736 81.5995330810547
5.37337337337337 81.9639205932617
5.38938938938939 82.3282852172852
5.40540540540541 82.6926651000977
5.42142142142142 83.0570449829102
5.43743743743744 83.4214248657227
5.45345345345345 83.7858047485352
5.46946946946947 84.1501846313477
5.48548548548549 84.5145568847656
5.5015015015015 84.8789367675781
5.51751751751752 85.2433242797852
5.53353353353353 85.6076889038086
5.54954954954955 85.9720687866211
5.56556556556557 86.3364562988281
5.58158158158158 86.7008285522461
5.5975975975976 87.0652084350586
5.61361361361361 87.4295883178711
5.62962962962963 87.7939529418945
5.64564564564565 88.1583404541016
5.66166166166166 88.5227279663086
5.67767767767768 88.887092590332
5.69369369369369 89.2514801025391
5.70970970970971 89.6158676147461
5.72572572572573 89.9802322387695
5.74174174174174 90.344612121582
5.75775775775776 90.7089920043945
5.77377377377377 91.0733642578125
5.78978978978979 91.437744140625
5.80580580580581 91.802131652832
5.82182182182182 92.1664962768555
5.83783783783784 92.530876159668
5.85385385385385 92.895263671875
5.86986986986987 93.259635925293
5.88588588588589 93.6240158081055
5.9019019019019 93.9884033203125
5.91791791791792 94.3527679443359
5.93393393393393 94.717155456543
5.94994994994995 95.0815353393555
5.96596596596597 95.4458999633789
5.98198198198198 95.8102951049805
5.997997997998 96.1746673583984
6.01401401401401 96.5390472412109
6.03003003003003 96.903434753418
6.04604604604605 97.2678146362305
6.06206206206206 97.632194519043
6.07807807807808 97.9965667724609
6.09409409409409 98.360954284668
6.11011011011011 98.7253265380859
6.12612612612613 99.0896987915039
6.14214214214214 99.4540863037109
6.15815815815816 99.8184661865234
6.17417417417417 100.182838439941
6.19019019019019 100.547218322754
6.20620620620621 100.911598205566
6.22222222222222 101.275970458984
6.23823823823824 101.640350341797
6.25425425425425 102.004730224609
6.27027027027027 102.369102478027
6.28628628628629 102.73348236084
6.3023023023023 103.097869873047
6.31831831831832 103.462242126465
6.33433433433433 103.826622009277
6.35035035035035 104.19100189209
6.36636636636637 104.555381774902
6.38238238238238 104.919761657715
6.3983983983984 105.284141540527
6.41441441441441 105.648506164551
6.43043043043043 106.012886047363
6.44644644644645 106.37727355957
6.46246246246246 106.741645812988
6.47847847847848 107.106025695801
6.49449449449449 107.470405578613
6.51051051051051 107.834785461426
6.52652652652653 108.199165344238
6.54254254254254 108.563545227051
6.55855855855856 108.927909851074
6.57457457457457 109.292297363281
6.59059059059059 109.656677246094
6.60660660660661 110.021041870117
6.62262262262262 110.385429382324
6.63863863863864 110.749809265137
6.65465465465465 111.11417388916
6.67067067067067 111.478569030762
6.68668668668669 111.842948913574
6.7027027027027 112.207321166992
6.71871871871872 112.57169342041
6.73473473473473 112.936073303223
6.75075075075075 113.300437927246
6.76676676676677 113.664833068848
6.78278278278278 114.029220581055
6.7987987987988 114.393585205078
6.81481481481481 114.757972717285
6.83083083083083 115.122367858887
6.84684684684685 115.486747741699
6.86286286286286 115.851112365723
6.87887887887888 116.215492248535
6.89489489489489 116.579879760742
6.91091091091091 116.944244384766
6.92692692692693 117.308631896973
6.94294294294294 117.673011779785
6.95895895895896 118.037376403809
6.97497497497497 118.401756286621
6.99099099099099 118.766151428223
7.00700700700701 119.130516052246
7.02302302302302 119.494903564453
7.03903903903904 119.85929107666
7.05505505505506 120.223655700684
7.07107107107107 120.588035583496
7.08708708708709 120.952415466309
7.1031031031031 121.316780090332
7.11911911911912 121.681167602539
7.13513513513514 122.045555114746
7.15115115115115 122.40991973877
7.16716716716717 122.774314880371
7.18318318318318 123.138679504395
7.1991991991992 123.503074645996
7.21521521521522 123.86743927002
7.23123123123123 124.231819152832
7.24724724724725 124.596199035645
7.26326326326326 124.960571289062
7.27927927927928 125.32494354248
7.2952952952953 125.689338684082
7.31131131131131 126.053703308105
7.32732732732733 126.418083190918
7.34334334334334 126.78247833252
7.35935935935936 127.146842956543
7.37537537537538 127.511207580566
7.39139139139139 127.875602722168
7.40740740740741 128.239990234375
7.42342342342342 128.604354858398
7.43943943943944 128.96875
7.45545545545546 129.333129882812
7.47147147147147 129.697494506836
7.48748748748749 130.061889648438
7.5035035035035 130.426239013672
7.51951951951952 130.790618896484
7.53553553553554 131.155014038086
7.55155155155155 131.519393920898
7.56756756756757 131.883758544922
7.58358358358358 132.248153686523
7.5995995995996 132.612518310547
7.61561561561562 132.976913452148
7.63163163163163 133.34130859375
7.64764764764765 133.705688476562
7.66366366366366 134.070053100586
7.67967967967968 134.434448242188
7.6956956956957 134.798812866211
7.71171171171171 135.163192749023
7.72772772772773 135.527587890625
7.74374374374374 135.891937255859
7.75975975975976 136.25634765625
7.77577577577578 136.620712280273
7.79179179179179 136.985076904297
7.80780780780781 137.349472045898
7.82382382382382 137.713851928711
7.83983983983984 138.078216552734
7.85585585585586 138.442611694336
7.87187187187187 138.806976318359
7.88788788788789 139.171356201172
7.9039039039039 139.535736083984
7.91991991991992 139.900115966797
7.93593593593594 140.264495849609
7.95195195195195 140.628875732422
7.96796796796797 140.993255615234
7.98398398398398 141.357635498047
8 141.722015380859
};
\end{axis}

\end{tikzpicture}

%% file: contents/02_background.tex

\subsection{Bayesian Neural Networks}
\label{subsec:BNNs}

Let $f: \R^n \times \R^d \to \R^k$ defined by $(x, \theta) \mapsto f(x; \theta)$ be an $L$-layer neural network. Here, $\theta$ is the vector of all the parameters of $f$. Suppose that the size of each layer of $f$ is given by the sequence of $(n_l \in \mathbb{Z}_{> 0})_{l=1}^L$. Then, for each $l = 1, \dots, L$, the $l$-th layer of $f$ is defined by
\begin{equation} \label{eq:nn_units}
    a^{(l)} := W^{(l)} h^{(l-1)} + b^{(l)} \\
\end{equation}
with
\begin{equation*}
    h^{(l)} :=
            \begin{cases}
                \varphi(a^{(l)})    & \text{if } l < L \\
                a^{(l)}             & \text{if } l = L  \, ,
            \end{cases}
\end{equation*}
where $W^{(l)} \in \R^{n_l \times n_{l-1}}$ and $b^{(l)} \in \R^{n_l}$ are the weight matrix and bias vector of the layer, and $\varphi$ is a component-wise activation function. We call the vector $h^{(l)} \in \R^{n_l}$ the $l$-th hidden units of $f$. Note that by convention, we consider $n_0 := n$ and $n_L := k$, while $h^{(0)} := x$ and $h^{(L)} := f(x; \theta)$.

From the Bayesian perspective, the ubiquitous training formalism of neural networks amounts to MAP estimation: The empirical risk and the regularizer are interpretable as the negative log-likelihood under an i.i.d. dataset $\D := \{ x_i, y_i \}_{i = 1}^m$ and the negative log-prior, respectively. That is, the loss function is interpreted as
\begin{equation}\label{eq:map_obj}
    \begin{aligned}
        \mathcal{L}(\theta) &:= - \sum_{i=1}^m \log p(y_i \mid f(x_i; \theta)) - \log p(\theta) \\
            &= - \log p(\theta \mid \D) \, .
    \end{aligned}
\end{equation}
In this view, the \emph{de facto} weight decay regularizer amounts to a zero-mean isotropic Gaussian prior $p(\theta) = \N(0, \lambda^\inv I)$ \ with a scalar precision hyperparameter $\lambda$. Meanwhile, the usual softmax and quadratic output losses correspond to the Categorical and Gaussian distributions over $y_i$ in the case of classification and regression, respectively.

MAP-trained neural networks have been shown to be overconfident \citep{hein2019relu} and BNNs can mitigate this issue \citep{kristiadi2020being}. BNNs quantify epistemic uncertainty by inferring the full posterior distribution of the parameters $\theta$, instead of just a single point estimate in MAP training. Given that $p(\theta \mid \D)$ is the posterior, then the prediction for any test point $x \in \R^n$ is obtained via marginalization
\begin{equation}\label{eq:pred_dist}
    p(y \mid x, \D) = \int p(y \mid f(x; \theta)) \, p(\theta \mid \D) \, d\theta \, ,
\end{equation}
which captures the uncertainty encoded in the posterior.

\subsection{Laplace Approximations}
\label{subsec:laplace}

In deep learning, since the exact Bayesian posterior is intractable, approximate Bayesian inference methods are used. Laplace approximations (LAs) are an important family of such methods. Let $\theta_\map$ be the minimizer of \eqref{eq:map_obj}, which corresponds to a mode of the posterior distribution. A LA locally approximates the posterior using a Gaussian
\begin{equation*}
    p(\theta \mid \D) \approx \N(\theta_\map, \varSigma)  \, ,
\end{equation*}
where $\varSigma := (\nabla^2 \L \vert_{\theta_\map})^\inv$ is the inverse Hessian of the loss function, evaluated at the MAP estimate $\theta_\map$. Thus, LAs construct an approximate Gaussian posterior \emph{around} $\theta_\map$, whose precision equals to the Hessian of the loss at $\theta_\map$---the ``curvature'' of the loss landscape at $\theta_\map$, cf. \cref{fig:one} (top) for an illustration.

While the covariance of a LA is tied to the weight decay of the loss, a common practice in LAs is to tune the prior precision under some objective in a \emph{post-hoc} manner \citep{ritter_scalable_2018,kristiadi2020being}. In other words, the MAP estimation and the covariance inference are thought of as separate, independent processes. For example, given a fixed MAP estimate, one can maximize the log-likelihood of a LA w.r.t. the prior precision to obtain the covariance. This hyperparameter tuning can thus be thought of as an \emph{uncertainty tuning}.

A recent example of LAs is the Kronecker-factored Laplace (KFL) \citep{ritter_scalable_2018}. The key idea is to approximate the Hessian matrix with the layer-wise Kronecker factorization scheme proposed by \citet{heskes2000natural,martens2015optimizing}. That is, for each layer $l = 1, \dots, L$, KFL assumes that the Hessian corresponding to the $l$-th weight matrix $W^{(l)} \in \R^{n_l \times n_{l-1}}$ can be written as the Kronecker product $G^{(l)} \otimes A^{(l)}$ for some $G^{(l)} \in \R^{n_l \times n_l}$ and $A^{(l)} \in \R^{n_{l-1} \times n_{l-1}}$. This assumption brings the inversion cost of the Hessian down to $\Theta(n_l^3 + n_{l-1}^3)$, instead of the usual $\Theta(n_l^3 n_{l-1}^3)$ cost. Note that the approximate Hessian can easily be computed via tools such as BackPACK \citep{dangel2020backpack}.

Even in the case when a closed-form Laplace-approximated posterior can be obtained, the integral \eqref{eq:pred_dist} in general does not have an analytic solution since $f$ is nonlinear. To alleviate this, one can simply employ Monte-Carlo (MC) integration by sampling from the Gaussian:
\begin{align*}
    p(y \mid x, \D) &\approx \frac{1}{S} \sum_{s=1}^S p(y \mid f(x; \theta_s)) \\
         &\text{with}\enspace \theta_s \sim \N(\theta_\map, \varSigma) \, ,
\end{align*}
for $S$ number of samples.

Alternatively, a closed-form approximation to the predictive distribution---useful for analysis but has also been shown to be better than MC integration in practice \citep{foong2019between,immer2020improving}---can be obtained by linearizing the network w.r.t. its parameter at the MAP estimate.\footnote{The resulting network is still non-linear in its input, but linear in its parameters.} That is, given any input $x \in \R^n$ and the Jacobian matrix $J(x) := \nabla_\theta f(x; \theta) \vert_{\theta_\text{MAP}} \in \R^{d \times k}$, we Taylor-approximate the network as
\begin{equation} \label{eq:linearization}
    f(x; \theta) \approx f(x; \theta_\text{MAP}) + J(x)^\top (\theta - \theta_\text{MAP}) \, .
\end{equation}
Under this approximation, since $\theta$ is \emph{a posteriori} distributed as Gaussian $\N(\theta_\text{MAP}, \varSigma)$, it follows that the marginal distribution over the network output $f(x)$ is also a Gaussian \citep[Sec. 5.7.3]{bishop2006prml}, given by
\begin{equation} \label{eq:linearized_dist}
    p(f(x) \mid x, \D) \sim \N(f(x; \theta_\text{MAP}), J(x)^\top \varSigma \, J(x)) \, .
\end{equation}
For classification, one can then use the so-called probit approximation \citep{spiegelhalter1990sequential,mackay1992evidence} or its generalization \citep{gibbs1997bayesian} to obtain the predictive distribution. In the binary classification case, this is
\begin{equation} \label{eq:pred_dist_approx}
    \begin{aligned}
        p(y = 1 \mid x, \D) &= \int \sigma(f(x)) \, p(f(x) \mid x, \D) \, d(f(x)) \\
            &\approx \sigma \left( \frac{f(x; \theta_\map)}{\sqrt{1 + \pi/8 \, v(x)}} \right) \, ,
    \end{aligned}
\end{equation}
where $v(x) := J(x)^\top \varSigma \, J(x)$ is the variance of $f(x)$ under \eqref{eq:linearized_dist}. Using this approximation, we can clearly see the connection between output variance and predictive uncertainty: As $v(x)$ increases, the predictive probability becomes closer $0.5$ and therefore the predictive entropy increases.

%% file: contents/03_method.tex

In this section, we introduce \emph{uncertainty units}, which can be added to the layers of any MAP-trained network (\cref{subsec:lula_construction}) and trained via an uncertainty-aware loss (\cref{subsec:lula_training}) to improve uncertainty calibration under Laplace approximations. All proofs are in \cref{appendix:proofs}.

\subsection{Construction}
\label{subsec:lula_construction}

Let $f: \R^n \times \R^d \to \R^k$ be a MAP-trained $L$-layer neural network with parameters $\theta_\map = ( W_\map^{(l)}, b_\map^{(l)} )_{l=1}^{L}$. The premise of our method is simple: At each hidden layer $l = 1, \dots, L-1$, we add $m_l \in \mathbb{Z}_{\geq 0}$ additional hidden units (under the original activation function) to $h^{(l)}$---as a consequence, the $l$-th weight matrix and bias vector need to be extended to accommodate them. Our method augments these parameters in such a way that for any input $x \in \R^N$, the original network output $f(x; \theta_\map)$ is preserved, as follows.

For each layer $l = 1, \dots, L-1$ of the network $f$, we expand the MAP-estimated weight matrix $W_\map^{(l)} \in \R^{n_l \times n_{l-1}}$ and the bias vector $b_\map^{(l)} \in \R^{n_l}$ to obtain the following block matrix and vector:
\begin{equation} \label{eq:lula_structure}
  \begin{aligned}
      \widetilde{W}^{(l)} &:= \begin{pmatrix}
          W_\map^{(l)}       & 0 \\[5pt]
          \widehat{W}_1^{(l)} & \widehat{W}_2^{(l)} \\
      \end{pmatrix} \in \R^{(n_l + m_l) \times (n_{l-1} + m_{l-1})} \, , \\[1em]
      %
      %
      \widetilde{b}^{(l)} &:= \begin{pmatrix}
          b_\map^{(l)} \\[5pt]
          \widehat{b}^{(l)}
      \end{pmatrix} \in \R^{n_l + m_l} \, ,
  \end{aligned}
\end{equation}
to take into account the additional $m_l$ hidden units. We do not add additional units to the input layer, so $m_0 = 0$. Furthermore, for $l = L$, we define
\begin{equation} \label{eq:lula_structure_ll}
    \begin{aligned}
        \widetilde{W}^{(L)} &:= (W_\map^{(L)}, 0) \in \R^{k \times (n_{L-1} + m_{L-1})} \,; \\[0.5em]
        \widetilde{b}^{(L)} &:= b_\map^{(L)} \in \R^k \, ,
    \end{aligned}
\end{equation}
so that the output dimensionality is also unchanged. For brevity, we denote by $\widehat{\theta}^{(l)}$ the non-zero additional parameters in \eqref{eq:lula_structure}, i.e.~we define $\widehat{\theta}^{(l)}$ to be the tuple $(\widehat{W}_1^{(l)}, \widehat{W}_2^{(l)}, \widehat{b}^{(l)})$. Altogether, considering all layers $l = 1, \dots, L-1$, we denote
$$
    \widehat{\theta} := (\theta^{(l)})_{l=1}^{L-1} \, ,
$$
to be the tuple of all non-zero additional parameters of the network $f$. Furthermore, we write the resulting augmented network as $\widetilde{f}$ and the resulting overall parameter vector---consisting of $( \widetilde{W}^{(l)}, \widetilde{b}^{(l)} )_{l=1}^{L}$---as $\widetilde{\theta}_\map \in \R^{\widetilde{d}}$, where $\widetilde{d}$ is the resulting number of parameters. Refer to \cref{fig:lula_viz} for an illustration and \cref{algo:lula_construction} in \cref{appendix:implementation} for a step-by-step summary.  Note that we can easily extend this construction to convolutional networks by expanding the ``channel'' of hidden convolution layers.\footnote{E.g.~if the hidden units are a 3D array of (channel $\times$ height $\times$ width), then we expand the first dimension.}

Let us inspect the implication of this construction. Here for each $l=1, \dots, L-1$, the sub-matrices $\widehat{W}_1^{(l)}$, $\widehat{W}_2^{(l)}$ and the sub-vector $\widehat{b}^{(l)}$ contain parameters for the additional $m_{l}$ hidden units in the $l$-th layer. We are free to choose the values of these parameters since the upper-right quadrant of $\widetilde{W}^{(l)}$, i.e.~the zero part of the additional weights, \emph{deactivates} the $m_{l-1}$ additional hidden units in the previous layer, hence they do not contribute to the original hidden units in the $l$-th layer. Part (a) of the following proposition thus guarantees that the additional hidden units will not change the output of the network.

\begin{proposition}[Properties]
\label{prop:output_preservation}
    Let $f: \R^n \times \R^d \to \R^k$ be a MAP-trained $L$-layer network under dataset $\D$, and let $\theta_\map$ be the MAP estimate. Suppose $\widetilde{f}: \R^n \times \R^{\widetilde{d}} \to \R$ and $\widetilde{\theta}_\map \in \R^{\widetilde{d}}$ are obtained via the previous construction, and $\widetilde{\L}$ is the resulting loss function under $\widetilde{f}$.
    \begin{enumerate}[label=(\alph*)]
        \item For an arbitrary input $x \in \R^n$, we have $\widetilde{f}(x; \widetilde{\theta}_\map) = f(x; \theta_\map)$.
        \item The gradient of $\widetilde{\L}$ w.r.t. the additional weights in $\widetilde{W}^{(L)}$ is non-linear in $\widehat{\theta}$.
    \end{enumerate}
\end{proposition}
\begin{proof}[Proof Sketch]
    Part (a) is straightforward. For part (b), we can show that the gradient of the network output w.r.t. the additional zero weight in \eqref{eq:lula_structure_ll} is given by the additional hidden units of the previous layer. Note that these hidden units are nonlinear in the additional weights induced by LULA, due to the structure \eqref{eq:lula_structure}. The result then follows immediately by the chain rule. The full proof is in \cref{appendix:proofs}.
\end{proof}

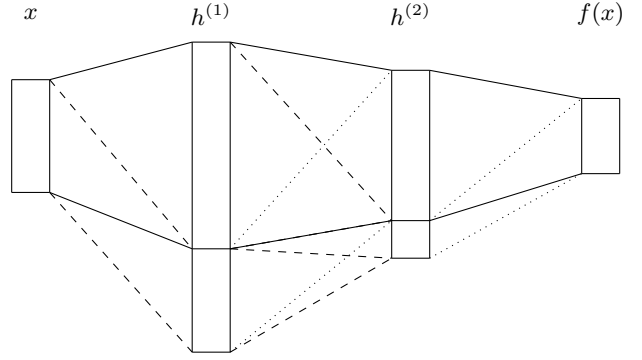
\begin{figure}
    \centering
    \input{figs/lula_viz.tikz}
    \caption{An illustration of the proposed construction. \textbf{Rectangles} represent layers, \textbf{solid lines} represent connection between layers, given by the original weight matrices $W_\text{MAP}^{(1)}, \dots, W_\text{MAP}^{(L)}$. The additional units are represented by the additional block at the bottom of each layer. \textbf{Dashed lines} correspond to the free parameters $\widehat{\theta}$, while \textbf{dotted lines} to the zero weights.}
    \label{fig:lula_viz}
\end{figure}

Part (b) of the last proposition tells us that the additional non-zero weights $\widehat{\theta}$ affect the loss landscape in a non-trivial way, and they, in general, induce non-trivial curvatures along the additional dimensions in the last-layer weight matrix \eqref{eq:lula_structure_ll} of the network. Therefore this construction non-trivially affects the covariance matrix in a LA. The implication of this insight to predictive uncertainty can be seen clearly in real-valued networks with diagonal LA posteriors, as the following proposition shows. (The usage of the network linearization below is necessary for analytical tractability.)

\begin{proposition}[Predictive Uncertainty]
\label{prop:variance_guarantee}
  Suppose $f: \R^n \times \R^d \to \R$ is a real-valued network and $\widetilde{f}$ is as constructed above. Suppose further that diagonal Laplace-approximated posteriors $\N(\theta_\map, \diag{\sigma})$, $\N(\widetilde{\theta}_\map, \diag{\widetilde{\sigma}})$ are employed for $f$ and $\widetilde{f}$, respectively. Under the linearization \eqref{eq:linearization}, for any input $x \in \R^n$, the variance over the output $\widetilde{f}(x; \widetilde{\theta})$ is at least that of $f(x; \theta)$.
\end{proposition}

In summary, the construction along with \cref{prop:output_preservation,prop:variance_guarantee} imply that the additional hidden units we have added to the original network are \emph{uncertainty units} under Laplace approximations, i.e.~hidden units that \emph{only} contribute to the Laplace-approximated uncertainty and not the predictions. Furthermore, by part (b) of \cref{prop:output_preservation}, the values of $\widehat{\theta}$---which can be set freely without affecting the output---influence the loss-landscape Hessian in a non-trivial way. They are thus learnable and so we call these units \emph{Learnable Uncertainty under Laplace Approximations (LULA)} units.

\subsection{Training}
\label{subsec:lula_training}

In this section, we discuss a way to train LULA units to improve predictive uncertainty under Laplace approximations. We follow a contemporary technique from the \emph{non}-Bayesian robust learning literature which has been shown to be effective in improving uncertainty calibration of non-Bayesian networks \citep[etc.]{lee2018training,hendrycks2018deep,bitterwolf2020certifiably}.

Let $f: \R^n \times \R^d \to \R^k$ be an $L$-layer neural network with a MAP-trained parameters $\theta_\text{MAP}$ and let $\widetilde{f}: \R^n \times \R^{\widetilde{d}} \to \R^k$ along with $\widetilde{\theta}_\text{MAP}$ be obtained by adding LULA units. Let $q(\widetilde{\theta}) := \N(\widetilde{\theta}_\map, \widetilde{\varSigma})$ be the Laplace-approximated posterior and $p(y \mid x, \D; \widetilde{\theta}_\text{MAP})$ be the (approximate) predictive distribution under the LA. Furthermore, let us denote the dataset sampled i.i.d. from the data distribution as $\D_\text{in}$ and that from some outlier distribution as $\D_\text{out}$, and let $H$ be the entropy functional. We construct the following loss function to induce high uncertainty on outliers while maintaining high confidence over the data (inliers):
\begin{equation} \label{eq:obj_exact}
    \begin{aligned}
        \L_\text{LULA}(&\widetilde{\theta}_\map) := \frac{1}{\abs{\D_\text{in}}} \sum_{x_\text{in} \in \D_\text{in}} H[p(y \mid x_\text{in}, \D; \widetilde{\theta}_\text{MAP})] \\
            &\quad- \frac{1}{\abs{\D_\text{out}}} \sum_{x_\text{out} \in \D_\text{out}} H[p(y \mid x_\text{out}, \D; \widetilde{\theta}_\text{MAP})] \, ,
    \end{aligned}
\end{equation}
and minimize it w.r.t. the free parameters $\widehat{\theta}$. This objective is task agnostic---it can be used in regression and classification networks alike. Furthermore, the first term of this objective can alternatively be replaced with the standard negative log-likelihood loss. In our case, since by \cref{prop:output_preservation}, predictions do not change under LULA, using the negative log-likelihood yields the same result as predictive entropy: they both only affect uncertainty and keep predictions over $\D_\text{in}$ confident. In any case, without this term, $\L_\text{LULA}$ potentially assigns the trivial solution of maximum uncertainty prediction everywhere in the input space.

The intuition of LULA training is as follows. By adding LULA units, we obtain a non-trivially augmented version of the network’s loss landscape (\cref{prop:output_preservation}(b)). The goal of LULA training is then to exploit the weight-space symmetry (i.e.~different parameters that induce the same output) arising from the construction as shown by \cref{prop:output_preservation}(a), and pick a point in the extended parameter space that is symmetric to the original parameters but has ``better'' curvatures, in the sense that they induce lower loss \eqref{eq:obj_exact}. These parameters, then, when used in a LA, improve the predictive uncertainty of standard non-LULA-augmented LAs.

\subsubsection{Practical Matters}

\paragraph*{Datasets} We can simply set $\D_\text{in}$ to be the validation set of the dataset $\D$. Meanwhile, $\D_\text{out}$ can be chosen depending on the task at hand, e.g. noise and large-scale natural image datasets can be used for regression and image classification tasks, respectively \citep{hendrycks2018deep}.

\paragraph*{Maintaining Weight Structures} Since our aim is to improve predictive uncertainty by exploiting weight-space symmetries given by the structure of LULA weights, we must maintain the structure of all weights and biases in $\widetilde{\theta}_\text{MAP}$, in accordance to \eqref{eq:lula_structure} and \eqref{eq:lula_structure_ll}. This can be enforced by gradient masking: For all $l = 1, \dots, L-1$, set the gradients of the blocks of $\widetilde{W}^{(l)}$ and $\widetilde{b}^{(l)}$ not corresponding to $\widehat{W}_1^{(l)}$, $\widehat{W}_2^{(l)}$, and $\widehat{b}^{(l)}$, to zero. Under this scheme, \cref{prop:output_preservation}(a) will still hold for trained LULA units.

\begin{algorithm}[t]
    \caption{Training LULA units.}
    \label{algo:lula_training}
    \begin{algorithmic}[1]
        \Require
            \Statex MAP-trained network $f$. Dataset $\D_\text{in}$, OOD dataset $\D_\text{out}$. Learning rate $\alpha$. Number of epochs $E$.

        \vspace{1em}

        \State Construct $\widetilde{f}$ from $f$ by following \cref{subsec:lula_construction}.
        \For{$i=1, \dots, E$}
            \State $q(\widetilde{\theta}) = \N(\widetilde{\theta}_\map, \widetilde{\varSigma}(\widetilde{\theta}_\map))$    
            \State Compute $\L_\text{LULA}(\widetilde{\theta}_\map)$ via \eqref{eq:obj_exact} with $q(\widetilde{\theta})$, $\D$, $\D_\text{out}$
            \State $g = \nabla \L_\text{LULA}(\widetilde{\theta}_\map)$
            \State $\widehat{g} = \mathtt{mask\_gradient}(g)$     
            \State $\widetilde{\theta}_\map = \widetilde{\theta}_\map - \alpha \widehat{g}$
        \EndFor
        \State $p(\widetilde{\theta} \mid \D) \approx \N(\widetilde{\theta}_\map, \widetilde{\varSigma}(\widetilde{\theta}_\map))$        
        \State \Return $\widetilde{f}$ and $p(\widetilde{\theta} \mid \D)$
    \end{algorithmic}
\end{algorithm}

\paragraph*{Laplace Approximations During Training} Since the covariance matrix $\widetilde{\varSigma}$ of the Laplace-approximated posterior depends on $\widetilde{\theta}_\text{MAP}$, it needs to be updated at every iteration during the optimization of $\L_\text{LULA}$. This can be expensive for large networks depending on the Laplace approximation used, not to mention that one must use the entire dataset $\D_\text{in}$ to obtain this matrix. As a simple and much cheaper proxy to the true covariance, we employ a simple diagonal Fisher information matrix \citep{amari1998natural,martens2014new}, obtained from a single minibatch, irrespective of the Laplace approximation variant employed at test time---we show in \cref{sec:experiments} that this training scheme is both effective and efficient.\footnote{The actual Laplace approximations used in all experiments are \emph{non}-diagonal.} Finally, we note that backpropagation through this diagonal matrix, which is fully determined by the network's gradient, does not pose a difficulty since modern deep learning libraries such as PyTorch and TensorFlow support ``double backprop'' efficiently. \Cref{algo:lula_training} provides a summary of LULA training in pseudocode. Code can be found in \url{https://github.com/wiseodd/lula}.

\begin{figure*}[t]
  \centering

  \subfloat[MAP]{\includegraphics[width=0.3\textwidth, height=0.10\textheight]{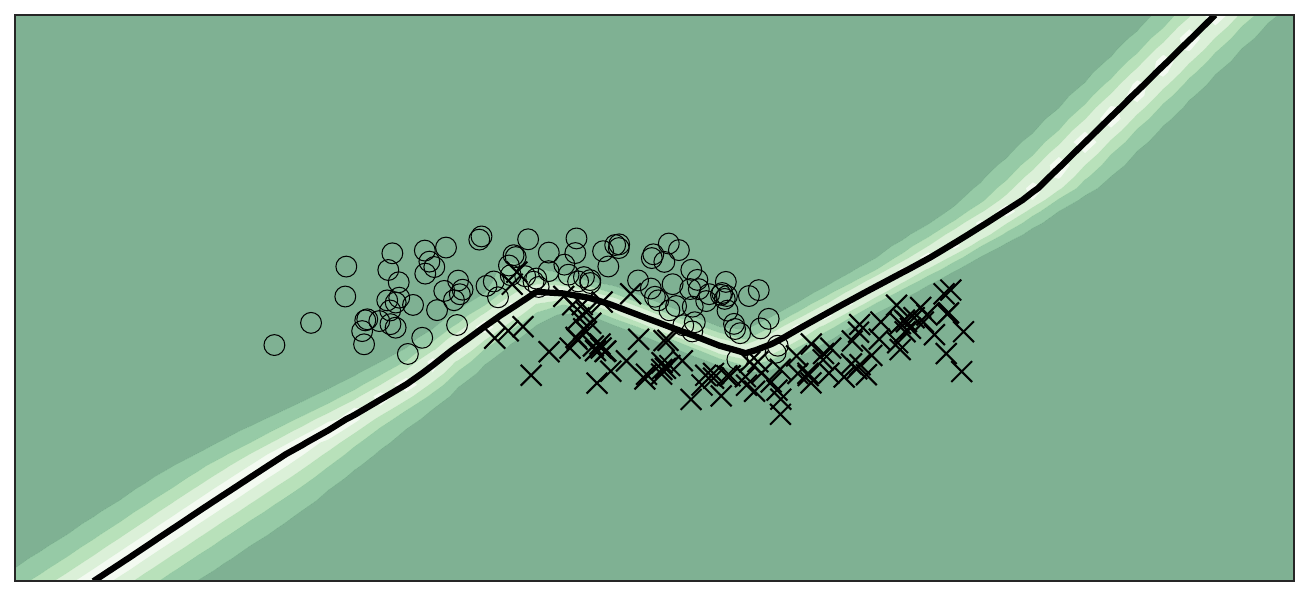}}
  \hfill
  \subfloat[LA]{\includegraphics[width=0.3\textwidth, height=0.10\textheight]{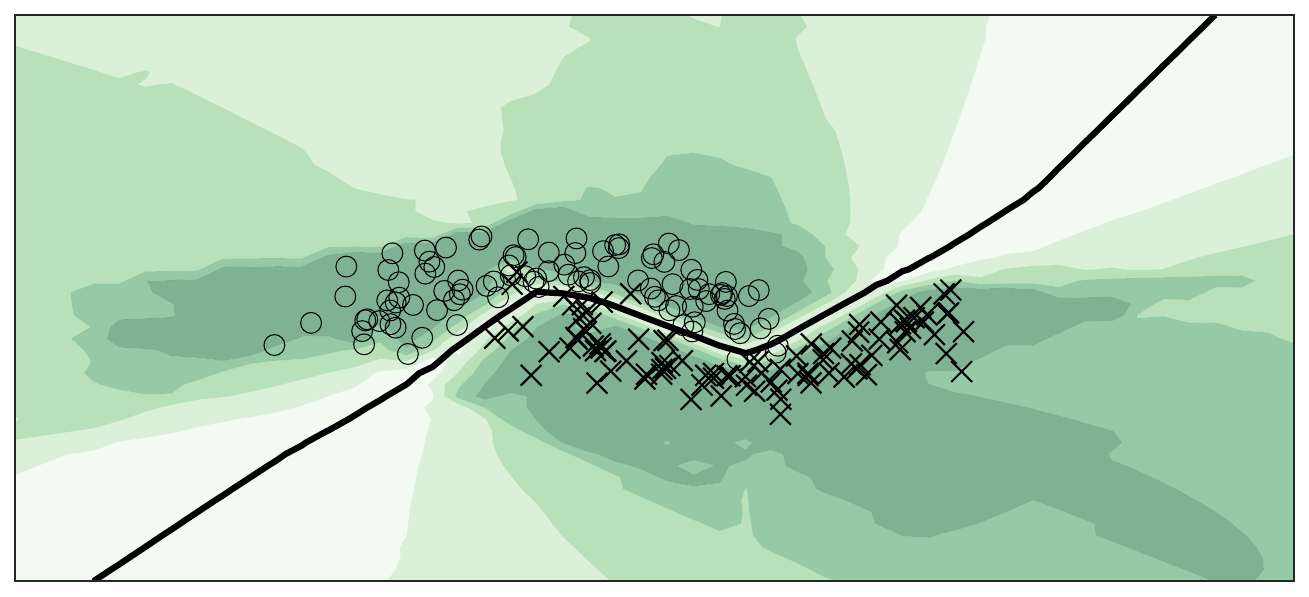}}
  \hfill
  \subfloat[LA-LULA]{\includegraphics[width=0.3\textwidth, height=0.10\textheight]{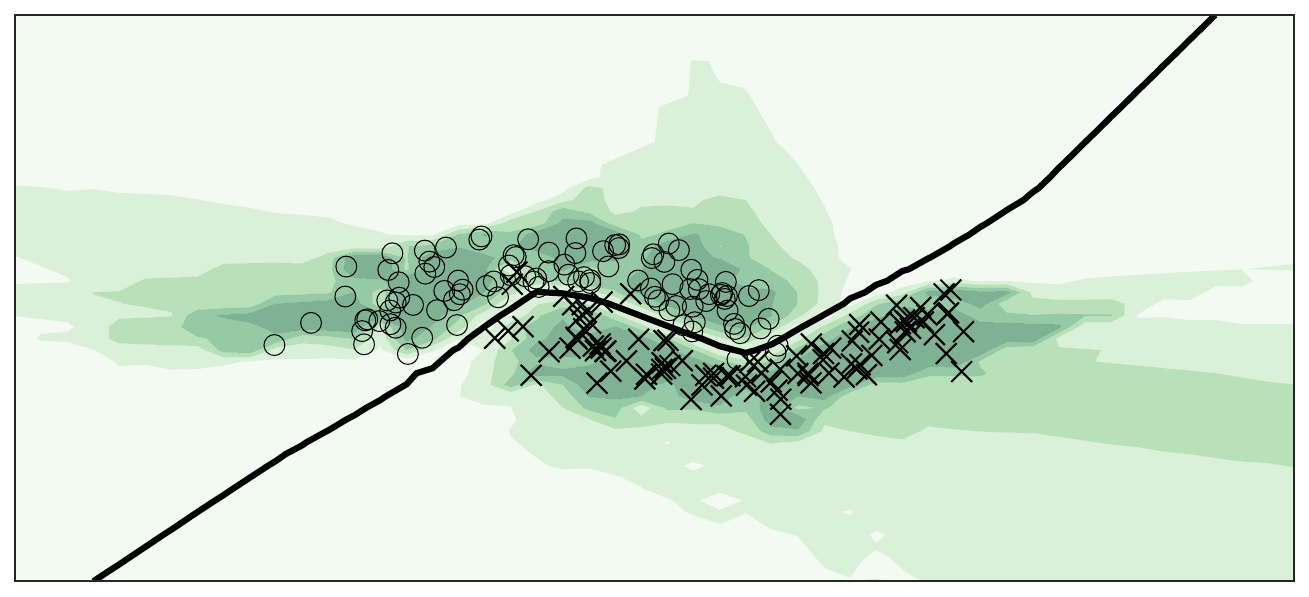}}
  \quad
  \subfloat{\includegraphics[width=0.04\textwidth, height=0.10\textheight]{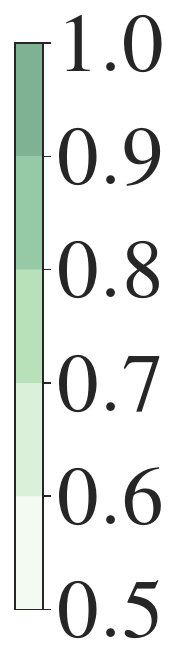}}

  \caption{Predictive uncertainty estimates of a standard LA and the LULA-augmented LA. Black curves and shades are decision boundaries and confidence estimates, respectively.}

  \label{fig:toy_exp}
\end{figure*}

%% file: figs/lula_viz.tikz
\begin{tikzpicture}
	\small
	\begin{pgfonlayer}{nodelayer}
		\node [style=none] (0) at (-8, 6) {};
		\node [style=none] (1) at (-8, 3) {};
		\node [style=none] (2) at (-7, 3) {};
		\node [style=none] (3) at (-7, 6) {};
		\node [style=none] (4) at (-3.25, 7) {};
		\node [style=none] (5) at (-3.25, -1.25) {};
		\node [style=none] (6) at (-2.25, -1.25) {};
		\node [style=none] (7) at (-2.25, 7) {};
		\node [style=none] (8) at (2, 6.25) {};
		\node [style=none] (9) at (3, 6.25) {};
		\node [style=none] (10) at (2, 1.25) {};
		\node [style=none] (11) at (3, 1.25) {};
		\node [style=none] (12) at (7, 5.5) {};
		\node [style=none] (13) at (8, 5.5) {};
		\node [style=none] (14) at (7, 3.5) {};
		\node [style=none] (15) at (8, 3.5) {};
		\node [style=none] (16) at (-3.25, 1.5) {};
		\node [style=none] (17) at (-2.25, 1.5) {};
		\node [style=none] (18) at (2, 2.25) {};
		\node [style=none] (19) at (3, 2.25) {};
		\node [style=none] (20) at (-7.5, 7.75) {$x$};
		\node [style=none] (21) at (-2.75, 7.75) {$h^{(1)}$};
		\node [style=none] (22) at (2.5, 7.75) {$h^{(2)}$};
		\node [style=none] (23) at (7.5, 7.75) {$f(x)$};
		\node [style=none] (25) at (-4.5, 4.75) {};
		\node [style=none] (26) at (-4, 1) {};
		\node [style=none] (28) at (1, 4.75) {};
		\node [style=none] (30) at (5, 4.75) {};
		\node [style=none] (31) at (-0.25, 1) {};
		\node [style=none] (32) at (3.5, 2) {};
		\node [style=none] (33) at (-6, 3.5) {};
		\node [style=none] (34) at (-1, 3.5) {};
	\end{pgfonlayer}
	\begin{pgfonlayer}{edgelayer}
		\draw (0.center) to (3.center);
		\draw (3.center) to (2.center);
		\draw (2.center) to (1.center);
		\draw (1.center) to (0.center);
		\draw (4.center) to (5.center);
		\draw (5.center) to (6.center);
		\draw (6.center) to (7.center);
		\draw (7.center) to (4.center);
		\draw (8.center) to (10.center);
		\draw (10.center) to (11.center);
		\draw (11.center) to (9.center);
		\draw (9.center) to (8.center);
		\draw (12.center) to (14.center);
		\draw (14.center) to (15.center);
		\draw (15.center) to (13.center);
		\draw (13.center) to (12.center);
		\draw (3.center) to (4.center);
		\draw [dashed] (2.center) to (5.center);
		\draw (2.center) to (16.center);
		\draw (16.center) to (17.center);
		\draw (17.center) to (18.center);
		\draw [dashed] (17.center) to (18.center);
		\draw (18.center) to (19.center);
		\draw (7.center) to (8.center);
		\draw (9.center) to (12.center);
		\draw (19.center) to (14.center);
		\draw [dashed] (6.center) to (10.center);
		\draw [dotted] (11.center) to (14.center);
		\draw [dashed] (3.center) to (16.center);
		\draw [dashed] (7.center) to (18.center);
		\draw [dashed] (17.center) to (10.center);
		\draw [dotted] (17.center) to (8.center);
		\draw [dotted] (19.center) to (12.center);
		\draw [dotted] (6.center) to (18.center);
	\end{pgfonlayer}
\end{tikzpicture}

%% file: contents/04_related.tex

While traditionally hyperparameter optimization in LAs requires re-training the network (under type-II maximum likelihood or the evidence framework \citep{mackay1992practical} or empirical Bayes \citep{robbins1956empirical}), tuning it in a \emph{post-hoc} manner has become increasingly common. \citet{ritter2018online,ritter_scalable_2018} tune the prior precision of a LA by maximizing the predictive log-likelihood. \citet{kristiadi2020being} extend this procedure by also using outliers to better calibrate the uncertainty. However, they are limited in terms of flexibility since the prior precision of the LAs constitutes a single scalar parameter. LULA can be seen as an extension of these approaches with greater flexibility and is complementary to them since it does not modify the prior precision used.

Confidence calibration via outliers has achieved state-of-the-art performance in non-Bayesian outlier detection. \citet{hendrycks2018deep,hein2019relu,meinke2020towards} use outliers to regularize the standard maximum-likelihood training. \citet{malinin2018predictive,malinin2019reverse} use outliers to train probabilistic models based on the Dirichlet distribution. In contrast to our approach, all these methods are neither Bayesian nor \emph{post-hoc}.

%% file: contents/05_experiments.tex

We empirically validate that LULA does improve vanilla LAs via toy and image classification experiments---results on UCI regression tasks are in the appendix. We expand the image classification experiment into \emph{dataset shift robustness} and \emph{out-of-distribution} (OOD) experiments to show LULA's performance over standard benchmark suites.

\subsection{Setup}
\label{subsec:setup}

\paragraph{Toy experiments} We use the ``cubic'' \citep{hernandez2015probabilistic} and ``two moons'' datasets for regression and classification, respectively. For classification, we use a full Laplace with generalized Gauss-Newton Hessian approximation on a three-layer FC network. For regression, we apply the Kronecker-factored Laplace (KFL) \citep{ritter_scalable_2018} on a two-layer fully-connected network. In this particular case, we directly use the predictive variance instead of (differential) entropy for \cref{eq:obj_exact}. The two are closely related, but in the case of regression with continuous output, the variance is easier to work with since it is lower-bounded by zero. Finally, the corresponding numbers of additional LULA units are $30$ and $50$, respectively.

\paragraph{Image classification} We use the following standard datasets: MNIST, SVHN, CIFAR-10, and CIFAR-100. For each dataset, we split its test set to obtain a validation set of size $2000$. On all datasets and all methods, we use the WideResNet-16-4 architecture \citep{zagoruyko2016wide} and optimize the network with Nesterov-SGD with weight decay \num{5e-4} and initial learning rate $0.1$ for $100$ epochs. We anneal the learning rate with the cosine decay method \citep{loshchilov2016sgdr}.

\paragraph{Baselines} We use the vanilla MAP-trained network (abbreviated as \textbf{MAP}), a last-layer KFL (\textbf{LA}), and Deep Ensemble (\textbf{DE}) \citep{lakshminarayanan2017simple} as baselines. For MAP and DE, we additionally use the temperature scaling post-processing scheme to improve their calibration (\textbf{Temp}) \citep{guo17calibration}. Specifically for DE, a single temperature hyperparameter is used for all ensemble members \citep{rahaman2020uncertainty}. Note that DE is used to represent the state-of-the-art uncertainty-quantification methods \citep{ovadia2019can}. For the Bayesian baseline (LA), we use a last-layer Laplace since it has been shown to be competitive to its all-layer counterpart while being much cheaper and thus more suitable for large networks \citep{kristiadi2020being}. We do not tune the prior variance of LA---it is obtained from the weight decay used during MAP training. Nevertheless, to show that LULA is also applicable to and can improve methods which their uncertainty is already explicitly tuned, we additionally use two OOD-trained/tuned baselines for the OOD-detection benchmark: (i) the last-layer Laplace where the prior variance is tuned via an OOD validation set (\textbf{LLLA}) \citep{kristiadi2020being}, and (ii) the outlier exposure method (\textbf{OE}) \citep{hendrycks2018deep} where OOD data is used during the MAP training itself. For the latter, we apply a standard last-layer KFL post-training (see \citep[Appendix D.6]{kristiadi2020being}).

\paragraph{LULA} For the toy experiments, we use uniform noise as $\D_\text{out}$. We add $50$ and $30$ LULA units to each layer of the toy regression and classification networks, respectively. Meanwhile, we use the downscaled ImageNet dataset \citep{chrabaszcz2017downsampled} as $\D_\text{out}$ for the image classification experiments. We do not use the 80 Million Tiny Images dataset \citep{torralba2008eightymillion} as used by \citet{hendrycks2018deep,meinke2020towards,bitterwolf2020certifiably} since it is not available anymore. We use the aforementioned ImageNet dataset as the OOD dataset for training/tuning the LLLA and OE baselines. We put LULA units on top of the pre-trained LA baseline and optimize them using Adam for $10$ epochs using the validation set. To pick the number of additional (last-layer) LULA units, we employ a grid search over the set $\{ 32, 64, 128, 256, 512, 1024 \}$ and pick the one minimizing validation LULA loss $\L_\text{LULA}$ under the LA. Finally, note that we implement LULA on top of the KFL discussed above, thus by doing so, we show that LULA is generally applicable even though it is specifically trained via a proxy diagonal LA.

\paragraph{Benchmark} For the dataset shift robustness experiment, we use the standard rotated-MNIST (MNIST-R) and corrupted-CIFAR-10 (CIFAR-10-C) datasets, which contain corrupted MNIST and CIFAR-10 test images with varying severity levels, respectively. Meanwhile, for the OOD experiment, we use $6$ OOD datasets for each \emph{in-distribution dataset} (i.e.~the dataset the model is trained on).

\paragraph{Metrics} First, we denote with ``$\downarrow$'' next to the name of a metric to indicate that lower values are better, and vice versa for ``$\uparrow$''. We use the standard uncertainty metrics: expected calibration error (ECE $\downarrow$) \citep{naeini2015obtaining}, Brier score ($\downarrow$) \citep{brier1950verification}, test log-likelihood ($\uparrow$), and average confidence (MMC $\downarrow$) \citep{hendrycks2018deep}. Additionally, for OOD detection, we use the FPR95 ($\downarrow$) metric which measures the false positive rate at a fixed true positive rate of $95\%$ when discriminating between in- and out-of-distribution data, based on their confidence (maximum predictive probability) estimates.

\begin{table}[t]
  \caption{Calibration and generalization performance. All values are in percent and averages over five prediction runs. Best ECE values among each pair of the vanilla and LULA-equipped methods (e.g. LA and LA-LULA) are in bold. Best overall values are underlined.}
  \label{tab:calib}

  \centering
  \setlength{\tabcolsep}{5pt}
  \fontsize{8}{9}\selectfont

  \begin{tabular}{lrrrrr}
    \toprule

     & \textbf{MNIST} & \textbf{SVHN} & \textbf{CIFAR-10} & \textbf{CIFAR-100}  \\

    \midrule

    \textbf{ECE} $\downarrow$ \\
    MAP & 13.8$\pm$0.0 & 9.7$\pm$0.0 & 12.2$\pm$0.0 & 16.6$\pm$0.0 \\
    MAP-Temp & 14.8$\pm$0.0 & \underline{2.0}$\pm$0.0 & 4.5$\pm$0.0 & \underline{4.1}$\pm$0.0 \\
    DE & \underline{13.2}$\pm$0.0 & 4.3$\pm$0.0 & 6.1$\pm$0.0 & 5.4$\pm$0.0 \\
    DE-Temp & 16.9$\pm$0.0 & 2.2$\pm$0.0 & \underline{3.8}$\pm$0.0 & 4.5$\pm$0.0 \\
    \midrule
    LA & \textbf{12.6}$\pm$0.1 & 9.3$\pm$0.0 & 10.9$\pm$0.3 & 7.0$\pm$0.1 \\
    \rowcolor{tablegray}
    LA-LULA & 14.8$\pm$0.3 & \textbf{3.3}$\pm$0.1 & \textbf{7.5}$\pm$0.1 & \textbf{5.3}$\pm$0.2 \\

    \midrule
    \midrule

    \textbf{Acc.} $\uparrow$ \\
    MAP & 99.7$\pm$0.0 & 97.1$\pm$0.0 & 95.0$\pm$0.0 & 75.8$\pm$0.0 \\
    MAP-Temp & 99.7$\pm$0.0 & 97.1$\pm$0.0 & 95.0$\pm$0.0 & 75.8$\pm$0.0 \\
    DE & 99.7$\pm$0.0 & 97.6$\pm$0.0 & 95.5$\pm$0.0 & 79.0$\pm$0.0 \\
    DE-Temp & 99.7$\pm$0.0 & 97.6$\pm$0.0 & 95.5$\pm$0.0 & 79.1$\pm$0.0 \\
    \midrule
    LA & 99.7$\pm$0.0 & 97.1$\pm$0.0 & 95.0$\pm$0.0 & 75.8$\pm$0.0 \\
    \rowcolor{tablegray}
    LA-LULA & 99.6$\pm$0.0 & 97.1$\pm$0.0 & 94.9$\pm$0.0 & 75.6$\pm$0.1 \\

    \bottomrule
  \end{tabular}

\end{table}

\subsection{Toy Experiments}
\label{subsec:toy}

We begin with toy regression and classification results in \cref{fig:one} (bottom) and \cref{fig:toy_exp}, respectively. As expected, the MAP-trained networks produce overconfident predictions in both cases. While LA provides meaningful uncertainty estimates, it can still be overconfident near the data. The same can be seen in the regression case: LA's uncertainty outside the data region grows slowly. LULA improves both cases: it makes (i) the regression uncertainty grow faster far from the data and (ii) the classification confidence more compact around the data region. Notice that in both cases LULA does not change the prediction of LA.

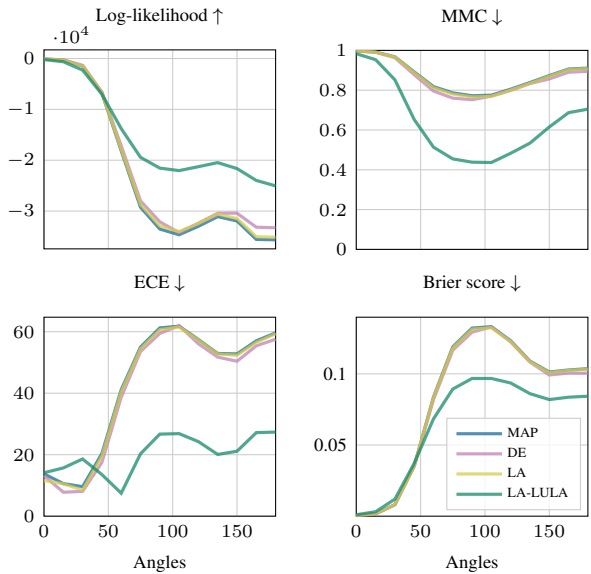
\begin{figure}[t]
  \centering

  \def\figfourwidth{0.27\textwidth}
  \def\figfourheight{0.18\textheight}


  \vspace{-1em}

  \hspace{1.5em}
  \subfloat{\input{figs/mnistr_loglik}}
  \hspace{2.5em}
  \subfloat{\input{figs/mnistr_mmc}}

  \vspace{-1em}

  \hspace{1.5em}
  \subfloat{\input{figs/mnistr_ece}}
  \hspace{2.5em}
  \subfloat{\input{figs/mnistr_brier}}


  \caption{Values of various uncertainty metrics as the rotation angle on MNIST images increases.}
  \label{fig:rotated_mnist}
\end{figure}

\subsection{Image Classifications}
\label{subsec:image_clf}

\subsubsection{Calibration}
\label{subsubsec:calib}

\Cref{tab:calib} summarizes the calibration and generalization performance of LULA in terms of ECE and test accuracy, respectively. We found that on ``harder'' datasets (SVHN, CIFAR-10, CIFAR-100), LULA consistently improves the vanilla LA's calibration, often even better than DE. However, on MNIST, both DE and LULA attain worse calibration than the vanilla LA. This might be because the accuracy of the network on MNIST is already almost perfect, thus even an overconfident classifier could yield a good ECE value---DE and LULA generally reduce confidence estimates (cf. \cref{tab:mmc} in the appendix) and thus yielding higher ECE values. Nevertheless, as we shall see in the next section, LULA is in general better calibrated to outliers than the other baselines on MNIST. As a final note, we emphasize that LULA preserves the predictive performance of the base LA and thus MAP's. This is important in practice: The allure of deep networks is their high predictive performance, thus, ``non-destructive'' \emph{post-hoc} methods are desirable.

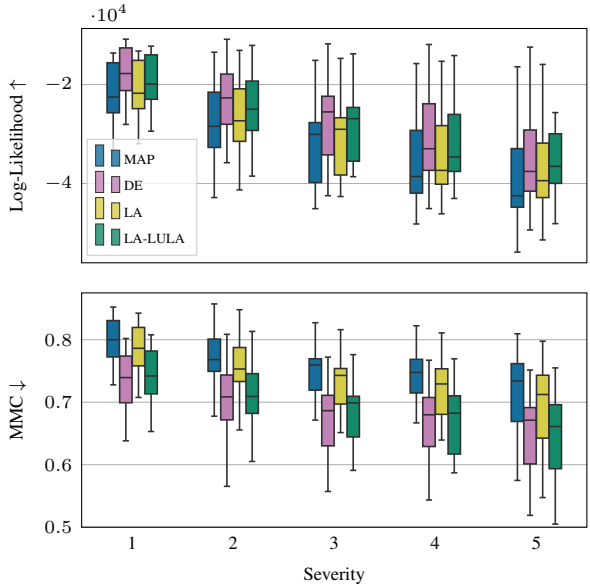
\begin{figure}[t]
  \centering

  \def\figfivewidth{0.48\textwidth}
  \def\figfiveheight{0.2\textheight}

  \hspace{2em}
  \subfloat{\input{figs/cifar10c_loglik}}

  \hspace{2em}
  \subfloat{\input{figs/cifar10c_mmc}}

  \caption{Summarized uncertainty quantification performance at each severity level of the CIFAR-10-C dataset.}
  \label{fig:cifar10c}
\end{figure}

\subsubsection{Dataset Shift Robustness}
\label{subsubsec:dataset_shift}

Dataset shift robustness tasks benchmark uncertainty calibration of a predictive model on corruptions or perturbations of the true dataset. To this end, we present various uncertainty metrics of LULA on the MNIST-R dataset in \cref{fig:rotated_mnist}. In all metrics considered, LULA improves not only the vanilla LA upon which LULA is implemented but also the state-of-the-art baseline in DE. Thus, even though LULA reduces calibration on the true MNIST dataset, it excels in making the network robust to outliers.

We furthermore present the results on the corrupted CIFAR-10 dataset in \cref{fig:cifar10c}. It can be seen that on average, LULA improves the vanilla LA, making it competitive to DE. In fact, on higher corruption levels, LULA can achieve better performance than DE, albeit marginally so. Nevertheless, this is important since standard BNNs have been shown to underperform compared to DE \citep{ovadia2019can}.

\begin{table}[t]
  \caption{OOD detection performance for each in-distribution dataset in terms of MMC and FPR95. Values are averages over six OOD test sets and five prediction runs. Best values among each pair of the vanilla and LULA-equipped methods are in bold. Best overall values are underlined.}
  \label{tab:ood}

  \centering
  \setlength{\tabcolsep}{5pt}
  \fontsize{8}{9}\selectfont

  \begin{tabular}{lrrrrr}
    \toprule

     & \textbf{MNIST} & \textbf{SVHN} & \textbf{CIFAR-10} & \textbf{CIFAR-100}  \\

    \midrule

    \textbf{MMC} $\downarrow$ \\
    MAP & 80.4$\pm$0.0 & 72.9$\pm$0.1 & 74.2$\pm$0.1 & 64.5$\pm$0.1 \\
    MAP-Temp & 82.2$\pm$0.0 & 63.4$\pm$0.0 & 60.5$\pm$0.0 & 48.2$\pm$0.1 \\
    DE & 73.8$\pm$0.0 & 58.3$\pm$0.1 & 66.3$\pm$0.0 & 46.8$\pm$0.0 \\
    DE-Temp & 84.1$\pm$0.0 & 59.0$\pm$0.1 & 62.0$\pm$0.0 & 46.5$\pm$0.1 \\
    \midrule
    LA & 78.7$\pm$0.1 & 72.1$\pm$0.1 & 70.7$\pm$0.2 & 53.4$\pm$0.2 \\
    \rowcolor{tablegray}
    LA-LULA & \textbf{46.0}$\pm$0.8 & \textbf{60.9}$\pm$0.2 & \textbf{63.8}$\pm$0.4 & \textbf{41.0}$\pm$0.5 \\
    \midrule
    LLLA & 61.0$\pm$0.4 & \textbf{47.3}$\pm$0.3 & 42.8$\pm$0.4 & 46.5$\pm$0.5 \\
    \rowcolor{tablegray}
    LLLA-LULA & \textbf{56.9}$\pm$0.8 & 52.1$\pm$0.4 & \textbf{\underline{35.1}}$\pm$0.3 & \textbf{\underline{33.1}}$\pm$0.7 \\
    \midrule
    OE & 35.2$\pm$0.0 & \textbf{\underline{18.0}}$\pm$0.0 & 53.4$\pm$0.0 & 51.8$\pm$0.0 \\
    \rowcolor{tablegray}
    OE-LULA & \textbf{\underline{22.6}}$\pm$0.2 & 20.1$\pm$0.2 & \textbf{52.0}$\pm$0.1 & \textbf{44.5}$\pm$0.2 \\

    \midrule
    \midrule

    \textbf{FPR95} $\downarrow$ \\
    MAP & 5.0$\pm$0.0 & 25.9$\pm$0.1 & 53.1$\pm$0.2 & 80.1$\pm$0.1 \\
    MAP-Temp & 5.0$\pm$0.0 & 25.6$\pm$0.1 & 47.0$\pm$0.2 & 77.1$\pm$0.1 \\
    DE & \underline{4.2}$\pm$0.0 & 11.9$\pm$0.1 & 47.6$\pm$0.0 & 59.3$\pm$0.1 \\
    DE-Temp & 4.5$\pm$0.0 & 16.4$\pm$0.1 & 44.8$\pm$0.0 & 72.3$\pm$0.1 \\
    \midrule
    LA & 4.9$\pm$0.0 & 25.5$\pm$0.2 & 48.5$\pm$0.5 & 78.3$\pm$0.5 \\
    \rowcolor{tablegray}
    LA-LULA & \textbf{5.8}$\pm$0.5 & \textbf{21.1}$\pm$0.4 & \textbf{39.5}$\pm$1.4 & \textbf{71.9}$\pm$1.3 \\
    \midrule
    LLLA & 5.8$\pm$0.5 & 22.0$\pm$1.8 & \textbf{\underline{23.7}}$\pm$0.5 & 75.4$\pm$0.9 \\
    \rowcolor{tablegray}
    LLLA-LULA & \textbf{4.5}$\pm$0.1 & \textbf{19.4}$\pm$0.5 & \textbf{\underline{22.9}}$\pm$0.8 & \textbf{68.4}$\pm$1.7 \\
    \midrule
    OE & 5.5$\pm$0.0 & \textbf{\underline{1.7}}$\pm$0.0 & 27.4$\pm$0.0 & 59.6$\pm$0.1 \\
    \rowcolor{tablegray}
    OE-LULA & \textbf{5.1}$\pm$0.3 & \textbf{\underline{1.7}}$\pm$0.0 & \textbf{26.7}$\pm$0.2 & \textbf{\underline{58.5}}$\pm$0.4 \\

    \bottomrule
  \end{tabular}

  \vspace{1em}
\end{table}

\subsubsection{OOD Detection}
\label{subsubsec:ood}

While dataset shift robustness tasks measure performance over outliers that are close to the true data, OOD detection tasks test performance on outliers that are far away from the data (e.g.~SVHN images as outliers for the CIFAR-10 dataset). \Cref{tab:ood} summarizes results. For each in-distribution dataset, LULA consistently improves the base LA, both in terms of its confidence estimates on OOD data (MMC) and its detection performance (FPR95). Furthermore, LULA in general assigns lower confidence to OOD data than DE. This suggests that, far from the data, LULA is more calibrated than DE. While LULA is better than DE in the detection of OOD data on CIFAR-10, DE yields a stronger FPR95 performance than LULA in general. Nevertheless, we stress that LULA is more cost-efficient than DE since it can be applied to any MAP-trained network \emph{post-hoc}. Moreover, unlike DE which requires us to train multiple (in our case, $5$) independent networks, LULA training is far cheaper than even the training time of a single network---see next section.

As stated in \cref{sec:related}, LULA is orthogonal to prior variance tuning methods commonly done in Laplace approximations. Hence, in \cref{tab:ood} we also show the OOD detection performance of LULA when applied to the LLLA baseline. We observe that LULA consistently improves LLLA. The same observation can also be seen when LULA is applied on top of a Laplace-approximated state-of-the-art OOD detector (OE): LULA also consistently improves OE even further.

\begin{table}[h]
  \caption{Wall-clock time in seconds for augmenting the WideResNet-16-4 network with $512$ LULA units and training them for ten epochs with a validation set of size $2000$.}
  \label{tab:cost_comp}

  \centering
  \small
  \setlength{\tabcolsep}{5pt}

  \begin{tabular}{lrrrrr}
    \toprule

     & \textbf{MNIST} & \textbf{SVHN} & \textbf{CIFAR-10} & \textbf{CIFAR-100}  \\

    \midrule

    Construction & 0.005 & 0.005 & 0.004 & 0.006 \\
    Training & 20.898 & 22.856 & 22.222 & 21.648 \\

    \bottomrule
  \end{tabular}
\end{table}

\subsection{Cost}

\Cref{tab:cost_comp} shows the computational overhead of LULA (wall-clock time, in seconds) on a single NVIDIA V100 GPU. The cost of augmenting the WideResNet-16-4 network with $512$ LULA units is negligible. The training time of these units is around $20$ seconds, which is also negligible compared to the time needed to do MAP training.

%% file: figs/mnistr_loglik.tex

\begin{tikzpicture}[trim axis left]

\definecolor{color0}{rgb}{0.0906862745098039,0.425980392156863,0.611274509803922}
\definecolor{color1}{rgb}{0.758823529411765,0.511764705882353,0.711764705882353}
\definecolor{color2}{rgb}{0.834803921568628,0.802450980392157,0.290686274509804}
\definecolor{color3}{rgb}{0.084313725490196,0.543137254901961,0.416666666666667}

\tikzstyle{every node}=[font=\scriptsize]

\begin{axis}[
width=\figfourwidth,
height=\figfourheight,
legend cell align={left},
legend style={fill opacity=0.8, draw opacity=1, text opacity=1, at={(0.97,0.03)}, anchor=south east, draw=white!80!black},
tick align=inside,
x grid style={white!80!black},
xmajorgrids,
xmajorticks=false,
xmin=0, xmax=180,
xtick style={draw=none},
title={Log-likelihood $\uparrow$},
y grid style={white!80!black},
ymajorgrids,
ymin=-37426.376064682, ymax=1699.87392082214,
ytick style={draw=none}
]
\addplot [very thick, color0, opacity=0.75]
table {%
0 -95.3966674804688
15 -315.192016601562
30 -1457.41564941406
45 -6787.3193359375
60 -18134.828125
75 -29230.8125
90 -33531.72265625
105 -34655.1796875
120 -32994.39453125
135 -31089.55859375
150 -31946.00390625
165 -35574.89453125
180 -35647.91015625
};
\addplot [very thick, color1, opacity=0.75]
table {%
0 -78.5919876098633
15 -260.830902099609
30 -1345.72241210938
45 -6579.921875
60 -17171.318359375
75 -27986.87109375
90 -32106.646484375
105 -34241.46875
120 -32375.66015625
135 -30427.037109375
150 -30409.4296875
165 -33169.0390625
180 -33275.828125
};
\addplot [very thick, color2, opacity=0.75]
table {%
0 -95.7052459716797
15 -314.789215087891
30 -1448.72131347656
45 -6703.03369140625
60 -17881.2421875
75 -28711.98828125
90 -32911.52734375
105 -34036.57421875
120 -32439.392578125
135 -30605.76171875
150 -31464.43359375
165 -35044.9609375
180 -35135.8359375
};
\addplot [very thick, color3, opacity=0.75]
table {%
0 -210.276245117188
15 -631.755615234375
30 -2279.24853515625
45 -7022.4794921875
60 -13767.6669921875
75 -19429.94921875
90 -21552.609375
105 -22039.642578125
120 -21267.701171875
135 -20472.42578125
150 -21641.279296875
165 -23978.970703125
180 -25042.580078125
};
\end{axis}

\end{tikzpicture}

%% file: figs/mnistr_mmc.tex

\begin{tikzpicture}[trim axis left, baseline]

\definecolor{color0}{rgb}{0.0906862745098039,0.425980392156863,0.611274509803922}
\definecolor{color1}{rgb}{0.758823529411765,0.511764705882353,0.711764705882353}
\definecolor{color2}{rgb}{0.834803921568628,0.802450980392157,0.290686274509804}
\definecolor{color3}{rgb}{0.084313725490196,0.543137254901961,0.416666666666667}

\tikzstyle{every node}=[font=\scriptsize]

\begin{axis}[
width=\figfourwidth,
height=\figfourheight,
legend cell align={left},
legend style={nodes={scale=0.75, transform shape}, fill opacity=0.8, draw opacity=1, text opacity=1, at={(1,0)}, anchor=south east, draw=white!80!black},
tick align=inside,
x grid style={white!80!black},
xmajorgrids,
xmajorticks=false,
xmin=0, xmax=180,
xtick style={draw=none},
title={MMC $\downarrow$},
y grid style={white!80!black},
ymajorgrids,
ymin=0, ymax=1,
ytick style={draw=none}
]
\addplot [very thick, color0, opacity=0.75]
table {%
0 0.997649550437927
15 0.991663217544556
30 0.967933475971222
45 0.892230629920959
60 0.818107604980469
75 0.786609590053558
90 0.772106468677521
105 0.774988055229187
120 0.805238425731659
135 0.838551044464111
150 0.87365198135376
165 0.906567275524139
180 0.910977303981781
};
\addplot [very thick, color1, opacity=0.75]
table {%
0 0.997144401073456
15 0.990667939186096
30 0.96347850561142
45 0.87811815738678
60 0.795869767665863
75 0.760099053382874
90 0.752240478992462
105 0.769990086555481
120 0.798616826534271
135 0.833298325538635
150 0.856808722019196
165 0.890261590480804
180 0.894849836826324
};
\addplot [very thick, color2, opacity=0.75]
table {%
0 0.997505724430084
15 0.991364359855652
30 0.966686546802521
45 0.888957381248474
60 0.813007533550262
75 0.780657410621643
90 0.766079246997833
105 0.769232571125031
120 0.800632953643799
135 0.83445006608963
150 0.870229244232178
165 0.903988540172577
180 0.908478617668152
};
\addplot [very thick, color3, opacity=0.75]
table {%
0 0.983301877975464
15 0.953443467617035
30 0.852357506752014
45 0.652787625789642
60 0.51375275850296
75 0.454696834087372
90 0.438228517770767
105 0.436648666858673
120 0.483070611953735
135 0.534101188182831
150 0.615113377571106
165 0.687286376953125
180 0.704646706581116
};
\end{axis}

\end{tikzpicture}

%% file: figs/mnistr_ece.tex

\begin{tikzpicture}[trim axis left]

\definecolor{color0}{rgb}{0.0906862745098039,0.425980392156863,0.611274509803922}
\definecolor{color1}{rgb}{0.758823529411765,0.511764705882353,0.711764705882353}
\definecolor{color2}{rgb}{0.834803921568628,0.802450980392157,0.290686274509804}
\definecolor{color3}{rgb}{0.084313725490196,0.543137254901961,0.416666666666667}

\tikzstyle{every node}=[font=\scriptsize]

\begin{axis}[
width=\figfourwidth,
height=\figfourheight,
legend cell align={left},
legend style={fill opacity=0.8, draw opacity=1, text opacity=1, at={(0.97,0.03)}, anchor=south east, draw=white!80!black},
tick align=inside,
x grid style={white!80!black},
xmajorgrids,
xmin=0, xmax=180,
xtick style={draw=none},
xlabel=Angles,
title={ECE $\downarrow$},
y grid style={white!80!black},
ymajorgrids,
ymin=0, ymax=64.7155269291683,
ytick style={draw=none}
]
\addplot [very thick, color0, opacity=0.75]
table {%
0 13.7654192307416
15 10.609529925836
30 9.58944537170382
45 20.3768040797829
60 41.1165283099784
75 54.9915317636612
90 61.2407633642692
105 61.8028865848097
120 57.4721657882408
135 52.9303164160444
150 52.7865097056409
165 57.0386593447825
180 59.6093997960944
};
\addplot [very thick, color1, opacity=0.75]
table {%
0 13.1904781400087
15 7.79837748268619
30 8.06872942762806
45 17.4878195484812
60 38.7091598868987
75 53.5640448188638
90 59.4319831363489
105 61.9897305360863
120 56.079884432467
135 51.7359565418135
150 50.3835132737447
165 55.4110242135632
180 57.5314466546613
};
\addplot [very thick, color2, opacity=0.75]
table {%
0 11.6027946628275
15 10.5479011183669
30 8.47868768861126
45 19.4961321136424
60 40.4462674960915
75 54.444416919052
90 60.660557770266
105 61.5030179392963
120 57.3103401926939
135 52.7640569191726
150 52.3605231984313
165 56.5200940373866
180 59.3658661682477
};
\addplot [very thick, color3, opacity=0.75]
table {%
0 14.1259244748984
15 15.6947514880505
30 18.6032153348764
45 13.5371404020602
60 7.47380267444642
75 20.2720651614546
90 26.6957288410095
105 26.8455984537238
120 24.2508361886198
135 20.0870581102358
150 21.1194050045574
165 27.1860058681646
180 27.3473668481525
};
\end{axis}

\end{tikzpicture}

%% file: figs/mnistr_brier.tex

\begin{tikzpicture}[trim axis left]

\definecolor{color0}{rgb}{0.0906862745098039,0.425980392156863,0.611274509803922}
\definecolor{color1}{rgb}{0.758823529411765,0.511764705882353,0.711764705882353}
\definecolor{color2}{rgb}{0.834803921568628,0.802450980392157,0.290686274509804}
\definecolor{color3}{rgb}{0.084313725490196,0.543137254901961,0.416666666666667}

\tikzstyle{every node}=[font=\scriptsize]

\begin{axis}[
width=\figfourwidth,
height=\figfourheight,
legend cell align={left},
legend style={nodes={scale=0.75, transform shape}, fill opacity=0.8, draw opacity=1, text opacity=1, at={(0.97,0.03)}, anchor=south east, draw=white!80!black},
tick align=inside,
x grid style={white!80!black},
xmajorgrids,
xmin=0, xmax=180,
xtick style={draw=none},
xlabel=Angles,
title={Brier score $\downarrow$},
y grid style={white!80!black},
ymajorgrids,
ymin=0, ymax=0.139772949801409,
ytick={0.05, 0.1},
yticklabels={0.05, 0.1},
ytick style={draw=none}
]
\addplot [very thick, color0, opacity=0.75]
table {%
0 0.000526922871358693
15 0.00180937675759196
30 0.00809390377253294
45 0.0352413691580296
60 0.083653524518013
75 0.118994385004044
90 0.132124483585358
105 0.133139386773109
120 0.12324646115303
135 0.108906224370003
150 0.101296946406364
165 0.102711282670498
180 0.103628344833851
};
\addlegendentry{MAP}
\addplot [very thick, color1, opacity=0.75]
table {%
0 0.000468126207124442
15 0.00147315359208733
30 0.00804297253489494
45 0.0351866260170937
60 0.0819162204861641
75 0.116586595773697
90 0.129292041063309
105 0.132888272404671
120 0.122374877333641
135 0.108223341405392
150 0.0991991609334946
165 0.100511401891708
180 0.100442603230476
};
\addlegendentry{DE}
\addplot [very thick, color2, opacity=0.75]
table {%
0 0.000528346165083349
15 0.00181854167021811
30 0.00809805002063513
45 0.0350452549755573
60 0.0832517370581627
75 0.118296921253204
90 0.131336867809296
105 0.132384821772575
120 0.122651219367981
135 0.108458705246449
150 0.100893676280975
165 0.102416843175888
180 0.103318065404892
};
\addlegendentry{LA}
\addplot [very thick, color3, opacity=0.75]
table {%
0 0.00100485153961927
15 0.00312011688947678
30 0.011970360763371
45 0.0371068269014359
60 0.0683690533041954
75 0.0893096402287483
90 0.0967507362365723
105 0.0967360734939575
120 0.0935560911893845
135 0.0860546603798866
150 0.08192478120327
165 0.0835964381694794
180 0.0842526480555534
};
\addlegendentry{LA-LULA}
\end{axis}

\end{tikzpicture}

%% file: figs/cifar10c_loglik.tex

\begin{tikzpicture}[baseline, trim axis left]

\definecolor{color0}{rgb}{0.0906862745098039,0.425980392156863,0.611274509803922}
\definecolor{color1}{rgb}{0.758823529411765,0.511764705882353,0.711764705882353}
\definecolor{color2}{rgb}{0.834803921568628,0.802450980392157,0.290686274509804}
\definecolor{color3}{rgb}{0.084313725490196,0.543137254901961,0.416666666666667}

\tikzstyle{every node}=[font=\scriptsize]

\begin{axis}[
width=\figfivewidth,
height=\figfiveheight,
legend cell align={left},
legend style={nodes={scale=0.75, transform shape}, fill opacity=0.8, draw opacity=1, text opacity=1, at={(0.01,0.03)}, anchor=south west, draw=white!80!black},
tick align=inside,
tick pos=left,
x grid style={white!69.0196078431373!black},
xmajorticks=false,
xmin=-0.5, xmax=4.5,
xtick style={draw=none},
xtick={0,1,2,3,4},
xticklabels={1,2,3,4,5},
y grid style={white!69.0196078431373!black},
ylabel={Log-Likelihood $\uparrow$},
ymin=-56024.9943359375, ymax=-8667.7126953125,
ytick style={draw=none},
ymajorgrids,
]
\path [draw=white!18.8235294117647!black, fill=color0, semithick]
(axis cs:-0.24875,-25704.376953125)
--(axis cs:-0.12625,-25704.376953125)
--(axis cs:-0.12625,-15590.6479492188)
--(axis cs:-0.24875,-15590.6479492188)
--(axis cs:-0.24875,-25704.376953125)
--cycle;
\path [draw=white!18.8235294117647!black, fill=color1, semithick]
(axis cs:-0.12375,-21213.130859375)
--(axis cs:-0.00125,-21213.130859375)
--(axis cs:-0.00125,-12589.328125)
--(axis cs:-0.12375,-12589.328125)
--(axis cs:-0.12375,-21213.130859375)
--cycle;
\path [draw=white!18.8235294117647!black, fill=color2, semithick]
(axis cs:0.00125,-24866.587890625)
--(axis cs:0.12375,-24866.587890625)
--(axis cs:0.12375,-15111.4467773438)
--(axis cs:0.00125,-15111.4467773438)
--(axis cs:0.00125,-24866.587890625)
--cycle;
\path [draw=white!18.8235294117647!black, fill=color3, semithick]
(axis cs:0.12625,-22976.3837890625)
--(axis cs:0.24875,-22976.3837890625)
--(axis cs:0.24875,-13991.841796875)
--(axis cs:0.12625,-13991.841796875)
--(axis cs:0.12625,-22976.3837890625)
--cycle;
\path [draw=white!18.8235294117647!black, fill=color0, semithick]
(axis cs:0.75125,-32713.39453125)
--(axis cs:0.87375,-32713.39453125)
--(axis cs:0.87375,-21542.376953125)
--(axis cs:0.75125,-21542.376953125)
--(axis cs:0.75125,-32713.39453125)
--cycle;
\path [draw=white!18.8235294117647!black, fill=color1, semithick]
(axis cs:0.87625,-28030.005859375)
--(axis cs:0.99875,-28030.005859375)
--(axis cs:0.99875,-17881.2421875)
--(axis cs:0.87625,-17881.2421875)
--(axis cs:0.87625,-28030.005859375)
--cycle;
\path [draw=white!18.8235294117647!black, fill=color2, semithick]
(axis cs:1.00125,-31476.5537109375)
--(axis cs:1.12375,-31476.5537109375)
--(axis cs:1.12375,-20863.3046875)
--(axis cs:1.00125,-20863.3046875)
--(axis cs:1.00125,-31476.5537109375)
--cycle;
\path [draw=white!18.8235294117647!black, fill=color3, semithick]
(axis cs:1.12625,-29272.3017578125)
--(axis cs:1.24875,-29272.3017578125)
--(axis cs:1.24875,-19293.853515625)
--(axis cs:1.12625,-19293.853515625)
--(axis cs:1.12625,-29272.3017578125)
--cycle;
\path [draw=white!18.8235294117647!black, fill=color0, semithick]
(axis cs:1.75125,-39787.248046875)
--(axis cs:1.87375,-39787.248046875)
--(axis cs:1.87375,-27684.20703125)
--(axis cs:1.75125,-27684.20703125)
--(axis cs:1.75125,-39787.248046875)
--cycle;
\path [draw=white!18.8235294117647!black, fill=color1, semithick]
(axis cs:1.87625,-34209.591796875)
--(axis cs:1.99875,-34209.591796875)
--(axis cs:1.99875,-22364.416015625)
--(axis cs:1.87625,-22364.416015625)
--(axis cs:1.87625,-34209.591796875)
--cycle;
\path [draw=white!18.8235294117647!black, fill=color2, semithick]
(axis cs:2.00125,-38280.142578125)
--(axis cs:2.12375,-38280.142578125)
--(axis cs:2.12375,-26702.4345703125)
--(axis cs:2.00125,-26702.4345703125)
--(axis cs:2.00125,-38280.142578125)
--cycle;
\path [draw=white!18.8235294117647!black, fill=color3, semithick]
(axis cs:2.12625,-35492.494140625)
--(axis cs:2.24875,-35492.494140625)
--(axis cs:2.24875,-24619.5908203125)
--(axis cs:2.12625,-24619.5908203125)
--(axis cs:2.12625,-35492.494140625)
--cycle;
\path [draw=white!18.8235294117647!black, fill=color0, semithick]
(axis cs:2.75125,-41949.208984375)
--(axis cs:2.87375,-41949.208984375)
--(axis cs:2.87375,-29258.9755859375)
--(axis cs:2.75125,-29258.9755859375)
--(axis cs:2.75125,-41949.208984375)
--cycle;
\path [draw=white!18.8235294117647!black, fill=color1, semithick]
(axis cs:2.87625,-37337.830078125)
--(axis cs:2.99875,-37337.830078125)
--(axis cs:2.99875,-23876.689453125)
--(axis cs:2.87625,-23876.689453125)
--(axis cs:2.87625,-37337.830078125)
--cycle;
\path [draw=white!18.8235294117647!black, fill=color2, semithick]
(axis cs:3.00125,-40141.345703125)
--(axis cs:3.12375,-40141.345703125)
--(axis cs:3.12375,-28299.4150390625)
--(axis cs:3.00125,-28299.4150390625)
--(axis cs:3.00125,-40141.345703125)
--cycle;
\path [draw=white!18.8235294117647!black, fill=color3, semithick]
(axis cs:3.12625,-37560.673828125)
--(axis cs:3.24875,-37560.673828125)
--(axis cs:3.24875,-26032.6474609375)
--(axis cs:3.12625,-26032.6474609375)
--(axis cs:3.12625,-37560.673828125)
--cycle;
\path [draw=white!18.8235294117647!black, fill=color0, semithick]
(axis cs:3.75125,-44780.080078125)
--(axis cs:3.87375,-44780.080078125)
--(axis cs:3.87375,-32941.982421875)
--(axis cs:3.75125,-32941.982421875)
--(axis cs:3.75125,-44780.080078125)
--cycle;
\path [draw=white!18.8235294117647!black, fill=color1, semithick]
(axis cs:3.87625,-41551.236328125)
--(axis cs:3.99875,-41551.236328125)
--(axis cs:3.99875,-29180.3623046875)
--(axis cs:3.87625,-29180.3623046875)
--(axis cs:3.87625,-41551.236328125)
--cycle;
\path [draw=white!18.8235294117647!black, fill=color2, semithick]
(axis cs:4.00125,-42845.482421875)
--(axis cs:4.12375,-42845.482421875)
--(axis cs:4.12375,-31828.8271484375)
--(axis cs:4.00125,-31828.8271484375)
--(axis cs:4.00125,-42845.482421875)
--cycle;
\path [draw=white!18.8235294117647!black, fill=color3, semithick]
(axis cs:4.12625,-39966.095703125)
--(axis cs:4.24875,-39966.095703125)
--(axis cs:4.24875,-29947.8271484375)
--(axis cs:4.12625,-29947.8271484375)
--(axis cs:4.12625,-39966.095703125)
--cycle;
\draw[draw=white!18.8235294117647!black,fill=color0,line width=0.3pt] (axis cs:0,0) rectangle (axis cs:0,0);
\addlegendimage{ybar,ybar legend,draw=white!18.8235294117647!black,fill=color0,line width=0.3pt};
\addlegendentry{MAP}

\draw[draw=white!18.8235294117647!black,fill=color1,line width=0.3pt] (axis cs:0,0) rectangle (axis cs:0,0);
\addlegendimage{ybar,ybar legend,draw=white!18.8235294117647!black,fill=color1,line width=0.3pt};
\addlegendentry{DE}

\draw[draw=white!18.8235294117647!black,fill=color2,line width=0.3pt] (axis cs:0,0) rectangle (axis cs:0,0);
\addlegendimage{ybar,ybar legend,draw=white!18.8235294117647!black,fill=color2,line width=0.3pt};
\addlegendentry{LA}

\draw[draw=white!18.8235294117647!black,fill=color3,line width=0.3pt] (axis cs:0,0) rectangle (axis cs:0,0);
\addlegendimage{ybar,ybar legend,draw=white!18.8235294117647!black,fill=color3,line width=0.3pt};
\addlegendentry{LA-LULA}

\addplot [semithick, white!18.8235294117647!black, forget plot]
table {%
-0.1875 -25704.376953125
-0.1875 -33187.93359375
};
\addplot [semithick, white!18.8235294117647!black, forget plot]
table {%
-0.1875 -15590.6479492188
-0.1875 -13615.935546875
};
\addplot [semithick, white!18.8235294117647!black, forget plot]
table {%
-0.218125 -33187.93359375
-0.156875 -33187.93359375
};
\addplot [semithick, white!18.8235294117647!black, forget plot]
table {%
-0.218125 -13615.935546875
-0.156875 -13615.935546875
};
\addplot [semithick, white!18.8235294117647!black, forget plot]
table {%
-0.0625 -21213.130859375
-0.0625 -28054.80859375
};
\addplot [semithick, white!18.8235294117647!black, forget plot]
table {%
-0.0625 -12589.328125
-0.0625 -10820.31640625
};
\addplot [semithick, white!18.8235294117647!black, forget plot]
table {%
-0.093125 -28054.80859375
-0.031875 -28054.80859375
};
\addplot [semithick, white!18.8235294117647!black, forget plot]
table {%
-0.093125 -10820.31640625
-0.031875 -10820.31640625
};
\addplot [semithick, white!18.8235294117647!black, forget plot]
table {%
0.0625 -24866.587890625
0.0625 -31960.154296875
};
\addplot [semithick, white!18.8235294117647!black, forget plot]
table {%
0.0625 -15111.4467773438
0.0625 -13226.697265625
};
\addplot [semithick, white!18.8235294117647!black, forget plot]
table {%
0.031875 -31960.154296875
0.093125 -31960.154296875
};
\addplot [semithick, white!18.8235294117647!black, forget plot]
table {%
0.031875 -13226.697265625
0.093125 -13226.697265625
};
\addplot [semithick, white!18.8235294117647!black, forget plot]
table {%
0.1875 -22976.3837890625
0.1875 -29427.796875
};
\addplot [semithick, white!18.8235294117647!black, forget plot]
table {%
0.1875 -13991.841796875
0.1875 -12234.88671875
};
\addplot [semithick, white!18.8235294117647!black, forget plot]
table {%
0.156875 -29427.796875
0.218125 -29427.796875
};
\addplot [semithick, white!18.8235294117647!black, forget plot]
table {%
0.156875 -12234.88671875
0.218125 -12234.88671875
};
\addplot [semithick, white!18.8235294117647!black, forget plot]
table {%
0.8125 -32713.39453125
0.8125 -42839
};
\addplot [semithick, white!18.8235294117647!black, forget plot]
table {%
0.8125 -21542.376953125
0.8125 -13459.8232421875
};
\addplot [semithick, white!18.8235294117647!black, forget plot]
table {%
0.781875 -42839
0.843125 -42839
};
\addplot [semithick, white!18.8235294117647!black, forget plot]
table {%
0.781875 -13459.8232421875
0.843125 -13459.8232421875
};
\addplot [semithick, white!18.8235294117647!black, forget plot]
table {%
0.9375 -28030.005859375
0.9375 -35798.2265625
};
\addplot [semithick, white!18.8235294117647!black, forget plot]
table {%
0.9375 -17881.2421875
0.9375 -10842.849609375
};
\addplot [semithick, white!18.8235294117647!black, forget plot]
table {%
0.906875 -35798.2265625
0.968125 -35798.2265625
};
\addplot [semithick, white!18.8235294117647!black, forget plot]
table {%
0.906875 -10842.849609375
0.968125 -10842.849609375
};
\addplot [semithick, white!18.8235294117647!black, forget plot]
table {%
1.0625 -31476.5537109375
1.0625 -41281.1328125
};
\addplot [semithick, white!18.8235294117647!black, forget plot]
table {%
1.0625 -20863.3046875
1.0625 -13084.728515625
};
\addplot [semithick, white!18.8235294117647!black, forget plot]
table {%
1.031875 -41281.1328125
1.093125 -41281.1328125
};
\addplot [semithick, white!18.8235294117647!black, forget plot]
table {%
1.031875 -13084.728515625
1.093125 -13084.728515625
};
\addplot [semithick, white!18.8235294117647!black, forget plot]
table {%
1.1875 -29272.3017578125
1.1875 -38506.3203125
};
\addplot [semithick, white!18.8235294117647!black, forget plot]
table {%
1.1875 -19293.853515625
1.1875 -12083.521484375
};
\addplot [semithick, white!18.8235294117647!black, forget plot]
table {%
1.156875 -38506.3203125
1.218125 -38506.3203125
};
\addplot [semithick, white!18.8235294117647!black, forget plot]
table {%
1.156875 -12083.521484375
1.218125 -12083.521484375
};
\addplot [semithick, white!18.8235294117647!black, forget plot]
table {%
1.8125 -39787.248046875
1.8125 -45084.6171875
};
\addplot [semithick, white!18.8235294117647!black, forget plot]
table {%
1.8125 -27684.20703125
1.8125 -15106.1796875
};
\addplot [semithick, white!18.8235294117647!black, forget plot]
table {%
1.781875 -45084.6171875
1.843125 -45084.6171875
};
\addplot [semithick, white!18.8235294117647!black, forget plot]
table {%
1.781875 -15106.1796875
1.843125 -15106.1796875
};
\addplot [semithick, white!18.8235294117647!black, forget plot]
table {%
1.9375 -34209.591796875
1.9375 -42460.59375
};
\addplot [semithick, white!18.8235294117647!black, forget plot]
table {%
1.9375 -22364.416015625
1.9375 -11775.107421875
};
\addplot [semithick, white!18.8235294117647!black, forget plot]
table {%
1.906875 -42460.59375
1.968125 -42460.59375
};
\addplot [semithick, white!18.8235294117647!black, forget plot]
table {%
1.906875 -11775.107421875
1.968125 -11775.107421875
};
\addplot [semithick, white!18.8235294117647!black, forget plot]
table {%
2.0625 -38280.142578125
2.0625 -42633.9609375
};
\addplot [semithick, white!18.8235294117647!black, forget plot]
table {%
2.0625 -26702.4345703125
2.0625 -14706.5927734375
};
\addplot [semithick, white!18.8235294117647!black, forget plot]
table {%
2.031875 -42633.9609375
2.093125 -42633.9609375
};
\addplot [semithick, white!18.8235294117647!black, forget plot]
table {%
2.031875 -14706.5927734375
2.093125 -14706.5927734375
};
\addplot [semithick, white!18.8235294117647!black, forget plot]
table {%
2.1875 -35492.494140625
2.1875 -38619.89453125
};
\addplot [semithick, white!18.8235294117647!black, forget plot]
table {%
2.1875 -24619.5908203125
2.1875 -13782.01171875
};
\addplot [semithick, white!18.8235294117647!black, forget plot]
table {%
2.156875 -38619.89453125
2.218125 -38619.89453125
};
\addplot [semithick, white!18.8235294117647!black, forget plot]
table {%
2.156875 -13782.01171875
2.218125 -13782.01171875
};
\addplot [semithick, white!18.8235294117647!black, forget plot]
table {%
2.8125 -41949.208984375
2.8125 -48180.28125
};
\addplot [semithick, white!18.8235294117647!black, forget plot]
table {%
2.8125 -29258.9755859375
2.8125 -15757.845703125
};
\addplot [semithick, white!18.8235294117647!black, forget plot]
table {%
2.781875 -48180.28125
2.843125 -48180.28125
};
\addplot [semithick, white!18.8235294117647!black, forget plot]
table {%
2.781875 -15757.845703125
2.843125 -15757.845703125
};
\addplot [semithick, white!18.8235294117647!black, forget plot]
table {%
2.9375 -37337.830078125
2.9375 -45067.2890625
};
\addplot [semithick, white!18.8235294117647!black, forget plot]
table {%
2.9375 -23876.689453125
2.9375 -11899.7197265625
};
\addplot [semithick, white!18.8235294117647!black, forget plot]
table {%
2.906875 -45067.2890625
2.968125 -45067.2890625
};
\addplot [semithick, white!18.8235294117647!black, forget plot]
table {%
2.906875 -11899.7197265625
2.968125 -11899.7197265625
};
\addplot [semithick, white!18.8235294117647!black, forget plot]
table {%
3.0625 -40141.345703125
3.0625 -46128.28515625
};
\addplot [semithick, white!18.8235294117647!black, forget plot]
table {%
3.0625 -28299.4150390625
3.0625 -15290.921875
};
\addplot [semithick, white!18.8235294117647!black, forget plot]
table {%
3.031875 -46128.28515625
3.093125 -46128.28515625
};
\addplot [semithick, white!18.8235294117647!black, forget plot]
table {%
3.031875 -15290.921875
3.093125 -15290.921875
};
\addplot [semithick, white!18.8235294117647!black, forget plot]
table {%
3.1875 -37560.673828125
3.1875 -43019.515625
};
\addplot [semithick, white!18.8235294117647!black, forget plot]
table {%
3.1875 -26032.6474609375
3.1875 -14132.87109375
};
\addplot [semithick, white!18.8235294117647!black, forget plot]
table {%
3.156875 -43019.515625
3.218125 -43019.515625
};
\addplot [semithick, white!18.8235294117647!black, forget plot]
table {%
3.156875 -14132.87109375
3.218125 -14132.87109375
};
\addplot [semithick, white!18.8235294117647!black, forget plot]
table {%
3.8125 -44780.080078125
3.8125 -53872.390625
};
\addplot [semithick, white!18.8235294117647!black, forget plot]
table {%
3.8125 -32941.982421875
3.8125 -16404.427734375
};
\addplot [semithick, white!18.8235294117647!black, forget plot]
table {%
3.781875 -53872.390625
3.843125 -53872.390625
};
\addplot [semithick, white!18.8235294117647!black, forget plot]
table {%
3.781875 -16404.427734375
3.843125 -16404.427734375
};
\addplot [semithick, white!18.8235294117647!black, forget plot]
table {%
3.9375 -41551.236328125
3.9375 -49396.03125
};
\addplot [semithick, white!18.8235294117647!black, forget plot]
table {%
3.9375 -29180.3623046875
3.9375 -12428.77734375
};
\addplot [semithick, white!18.8235294117647!black, forget plot]
table {%
3.906875 -49396.03125
3.968125 -49396.03125
};
\addplot [semithick, white!18.8235294117647!black, forget plot]
table {%
3.906875 -12428.77734375
3.968125 -12428.77734375
};
\addplot [semithick, white!18.8235294117647!black, forget plot]
table {%
4.0625 -42845.482421875
4.0625 -51404.91796875
};
\addplot [semithick, white!18.8235294117647!black, forget plot]
table {%
4.0625 -31828.8271484375
4.0625 -15930.1904296875
};
\addplot [semithick, white!18.8235294117647!black, forget plot]
table {%
4.031875 -51404.91796875
4.093125 -51404.91796875
};
\addplot [semithick, white!18.8235294117647!black, forget plot]
table {%
4.031875 -15930.1904296875
4.093125 -15930.1904296875
};
\addplot [semithick, white!18.8235294117647!black, forget plot]
table {%
4.1875 -39966.095703125
4.1875 -48113.984375
};
\addplot [semithick, white!18.8235294117647!black, forget plot]
table {%
4.1875 -29947.8271484375
4.1875 -25662.736328125
};
\addplot [semithick, white!18.8235294117647!black, forget plot]
table {%
4.156875 -48113.984375
4.218125 -48113.984375
};
\addplot [semithick, white!18.8235294117647!black, forget plot]
table {%
4.156875 -25662.736328125
4.218125 -25662.736328125
};
\addplot [semithick, white!18.8235294117647!black, forget plot]
table {%
-0.24875 -22537.17578125
-0.12625 -22537.17578125
};
\addplot [semithick, white!18.8235294117647!black, forget plot]
table {%
-0.12375 -17769.88671875
-0.00125 -17769.88671875
};
\addplot [semithick, white!18.8235294117647!black, forget plot]
table {%
0.00125 -21755.986328125
0.12375 -21755.986328125
};
\addplot [semithick, white!18.8235294117647!black, forget plot]
table {%
0.12625 -19905.02734375
0.24875 -19905.02734375
};
\addplot [semithick, white!18.8235294117647!black, forget plot]
table {%
0.75125 -28435.744140625
0.87375 -28435.744140625
};
\addplot [semithick, white!18.8235294117647!black, forget plot]
table {%
0.87625 -22694.376953125
0.99875 -22694.376953125
};
\addplot [semithick, white!18.8235294117647!black, forget plot]
table {%
1.00125 -27314.837890625
1.12375 -27314.837890625
};
\addplot [semithick, white!18.8235294117647!black, forget plot]
table {%
1.12625 -24972.359375
1.24875 -24972.359375
};
\addplot [semithick, white!18.8235294117647!black, forget plot]
table {%
1.75125 -30079.521484375
1.87375 -30079.521484375
};
\addplot [semithick, white!18.8235294117647!black, forget plot]
table {%
1.87625 -25526.9921875
1.99875 -25526.9921875
};
\addplot [semithick, white!18.8235294117647!black, forget plot]
table {%
2.00125 -29041.853515625
2.12375 -29041.853515625
};
\addplot [semithick, white!18.8235294117647!black, forget plot]
table {%
2.12625 -26900.1875
2.24875 -26900.1875
};
\addplot [semithick, white!18.8235294117647!black, forget plot]
table {%
2.75125 -38591.359375
2.87375 -38591.359375
};
\addplot [semithick, white!18.8235294117647!black, forget plot]
table {%
2.87625 -32973.14453125
2.99875 -32973.14453125
};
\addplot [semithick, white!18.8235294117647!black, forget plot]
table {%
3.00125 -37344.84765625
3.12375 -37344.84765625
};
\addplot [semithick, white!18.8235294117647!black, forget plot]
table {%
3.12625 -34637.87890625
3.24875 -34637.87890625
};
\addplot [semithick, white!18.8235294117647!black, forget plot]
table {%
3.75125 -42466.58203125
3.87375 -42466.58203125
};
\addplot [semithick, white!18.8235294117647!black, forget plot]
table {%
3.87625 -37556.46875
3.99875 -37556.46875
};
\addplot [semithick, white!18.8235294117647!black, forget plot]
table {%
4.00125 -39424.140625
4.12375 -39424.140625
};
\addplot [semithick, white!18.8235294117647!black, forget plot]
table {%
4.12625 -36532.328125
4.24875 -36532.328125
};
\end{axis}

\end{tikzpicture}

%% file: figs/cifar10c_mmc.tex

\begin{tikzpicture}[baseline, trim axis left]

\definecolor{color0}{rgb}{0.0906862745098039,0.425980392156863,0.611274509803922}
\definecolor{color1}{rgb}{0.758823529411765,0.511764705882353,0.711764705882353}
\definecolor{color2}{rgb}{0.834803921568628,0.802450980392157,0.290686274509804}
\definecolor{color3}{rgb}{0.084313725490196,0.543137254901961,0.416666666666667}

\tikzstyle{every node}=[font=\scriptsize]

\begin{axis}[
width=\figfivewidth,
height=\figfiveheight,
legend cell align={left},
legend style={nodes={scale=0.75, transform shape}, fill opacity=0.8, draw opacity=1, text opacity=1, at={(0,0)}, anchor=south west, draw=white!80!black},
tick align=inside,
tick pos=left,
x grid style={white!69.0196078431373!black},
xlabel={Severity},
xmin=-0.5, xmax=4.5,
xtick style={draw=none},
xtick={0,1,2,3,4},
xticklabels={1,2,3,4,5},
y grid style={white!69.0196078431373!black},
ylabel={MMC $\downarrow$},
ymin=0.5, ymax=0.874877294898033,
ytick style={draw=none},
ymajorgrids,
]
\path [draw=white!18.8235294117647!black, fill=color0, semithick]
(axis cs:-0.24875,0.772416710853577)
--(axis cs:-0.12625,0.772416710853577)
--(axis cs:-0.12625,0.830604821443558)
--(axis cs:-0.24875,0.830604821443558)
--(axis cs:-0.24875,0.772416710853577)
--cycle;
\path [draw=white!18.8235294117647!black, fill=color1, semithick]
(axis cs:-0.12375,0.699147790670395)
--(axis cs:-0.00125,0.699147790670395)
--(axis cs:-0.00125,0.773896336555481)
--(axis cs:-0.12375,0.773896336555481)
--(axis cs:-0.12375,0.699147790670395)
--cycle;
\path [draw=white!18.8235294117647!black, fill=color2, semithick]
(axis cs:0.00125,0.758135139942169)
--(axis cs:0.12375,0.758135139942169)
--(axis cs:0.12375,0.819619834423065)
--(axis cs:0.00125,0.819619834423065)
--(axis cs:0.00125,0.758135139942169)
--cycle;
\path [draw=white!18.8235294117647!black, fill=color3, semithick]
(axis cs:0.12625,0.713327586650848)
--(axis cs:0.24875,0.713327586650848)
--(axis cs:0.24875,0.781869620084763)
--(axis cs:0.12625,0.781869620084763)
--(axis cs:0.12625,0.713327586650848)
--cycle;
\path [draw=white!18.8235294117647!black, fill=color0, semithick]
(axis cs:0.75125,0.749334692955017)
--(axis cs:0.87375,0.749334692955017)
--(axis cs:0.87375,0.801014214754105)
--(axis cs:0.75125,0.801014214754105)
--(axis cs:0.75125,0.749334692955017)
--cycle;
\path [draw=white!18.8235294117647!black, fill=color1, semithick]
(axis cs:0.87625,0.671752274036407)
--(axis cs:0.99875,0.671752274036407)
--(axis cs:0.99875,0.74355810880661)
--(axis cs:0.87625,0.74355810880661)
--(axis cs:0.87625,0.671752274036407)
--cycle;
\path [draw=white!18.8235294117647!black, fill=color2, semithick]
(axis cs:1.00125,0.733120203018188)
--(axis cs:1.12375,0.733120203018188)
--(axis cs:1.12375,0.787484765052795)
--(axis cs:1.00125,0.787484765052795)
--(axis cs:1.00125,0.733120203018188)
--cycle;
\path [draw=white!18.8235294117647!black, fill=color3, semithick]
(axis cs:1.12625,0.682084739208221)
--(axis cs:1.24875,0.682084739208221)
--(axis cs:1.24875,0.745677918195724)
--(axis cs:1.12625,0.745677918195724)
--(axis cs:1.12625,0.682084739208221)
--cycle;
\path [draw=white!18.8235294117647!black, fill=color0, semithick]
(axis cs:1.75125,0.719272881746292)
--(axis cs:1.87375,0.719272881746292)
--(axis cs:1.87375,0.769477516412735)
--(axis cs:1.75125,0.769477516412735)
--(axis cs:1.75125,0.719272881746292)
--cycle;
\path [draw=white!18.8235294117647!black, fill=color1, semithick]
(axis cs:1.87625,0.630230993032455)
--(axis cs:1.99875,0.630230993032455)
--(axis cs:1.99875,0.711030513048172)
--(axis cs:1.87625,0.711030513048172)
--(axis cs:1.87625,0.630230993032455)
--cycle;
\path [draw=white!18.8235294117647!black, fill=color2, semithick]
(axis cs:2.00125,0.69707328081131)
--(axis cs:2.12375,0.69707328081131)
--(axis cs:2.12375,0.753904789686203)
--(axis cs:2.00125,0.753904789686203)
--(axis cs:2.00125,0.69707328081131)
--cycle;
\path [draw=white!18.8235294117647!black, fill=color3, semithick]
(axis cs:2.12625,0.644421130418777)
--(axis cs:2.24875,0.644421130418777)
--(axis cs:2.24875,0.709492892026901)
--(axis cs:2.12625,0.709492892026901)
--(axis cs:2.12625,0.644421130418777)
--cycle;
\path [draw=white!18.8235294117647!black, fill=color0, semithick]
(axis cs:2.75125,0.714768916368484)
--(axis cs:2.87375,0.714768916368484)
--(axis cs:2.87375,0.768564850091934)
--(axis cs:2.75125,0.768564850091934)
--(axis cs:2.75125,0.714768916368484)
--cycle;
\path [draw=white!18.8235294117647!black, fill=color1, semithick]
(axis cs:2.87625,0.629082053899765)
--(axis cs:2.99875,0.629082053899765)
--(axis cs:2.99875,0.707579404115677)
--(axis cs:2.87625,0.707579404115677)
--(axis cs:2.87625,0.629082053899765)
--cycle;
\path [draw=white!18.8235294117647!black, fill=color2, semithick]
(axis cs:3.00125,0.680557519197464)
--(axis cs:3.12375,0.680557519197464)
--(axis cs:3.12375,0.753449857234955)
--(axis cs:3.00125,0.753449857234955)
--(axis cs:3.00125,0.680557519197464)
--cycle;
\path [draw=white!18.8235294117647!black, fill=color3, semithick]
(axis cs:3.12625,0.617027103900909)
--(axis cs:3.24875,0.617027103900909)
--(axis cs:3.24875,0.71022042632103)
--(axis cs:3.12625,0.71022042632103)
--(axis cs:3.12625,0.617027103900909)
--cycle;
\path [draw=white!18.8235294117647!black, fill=color0, semithick]
(axis cs:3.75125,0.669158279895782)
--(axis cs:3.87375,0.669158279895782)
--(axis cs:3.87375,0.761750340461731)
--(axis cs:3.75125,0.761750340461731)
--(axis cs:3.75125,0.669158279895782)
--cycle;
\path [draw=white!18.8235294117647!black, fill=color1, semithick]
(axis cs:3.87625,0.60126057267189)
--(axis cs:3.99875,0.60126057267189)
--(axis cs:3.99875,0.69141098856926)
--(axis cs:3.87625,0.69141098856926)
--(axis cs:3.87625,0.60126057267189)
--cycle;
\path [draw=white!18.8235294117647!black, fill=color2, semithick]
(axis cs:4.00125,0.642528027296066)
--(axis cs:4.12375,0.642528027296066)
--(axis cs:4.12375,0.743139088153839)
--(axis cs:4.00125,0.743139088153839)
--(axis cs:4.00125,0.642528027296066)
--cycle;
\path [draw=white!18.8235294117647!black, fill=color3, semithick]
(axis cs:4.12625,0.593625158071518)
--(axis cs:4.24875,0.593625158071518)
--(axis cs:4.24875,0.696036100387573)
--(axis cs:4.12625,0.696036100387573)
--(axis cs:4.12625,0.593625158071518)
--cycle;
\draw[draw=white!18.8235294117647!black,fill=color0,line width=0.3pt] (axis cs:0,0) rectangle (axis cs:0,0);

\draw[draw=white!18.8235294117647!black,fill=color1,line width=0.3pt] (axis cs:0,0) rectangle (axis cs:0,0);

\draw[draw=white!18.8235294117647!black,fill=color2,line width=0.3pt] (axis cs:0,0) rectangle (axis cs:0,0);

\draw[draw=white!18.8235294117647!black,fill=color3,line width=0.3pt] (axis cs:0,0) rectangle (axis cs:0,0);

\addplot [semithick, white!18.8235294117647!black, forget plot]
table {%
-0.1875 0.772416710853577
-0.1875 0.727737724781036
};
\addplot [semithick, white!18.8235294117647!black, forget plot]
table {%
-0.1875 0.830604821443558
-0.1875 0.852262079715729
};
\addplot [semithick, white!18.8235294117647!black, forget plot]
table {%
-0.218125 0.727737724781036
-0.156875 0.727737724781036
};
\addplot [semithick, white!18.8235294117647!black, forget plot]
table {%
-0.218125 0.852262079715729
-0.156875 0.852262079715729
};
\addplot [semithick, white!18.8235294117647!black, forget plot]
table {%
-0.0625 0.699147790670395
-0.0625 0.638179898262024
};
\addplot [semithick, white!18.8235294117647!black, forget plot]
table {%
-0.0625 0.773896336555481
-0.0625 0.801881611347198
};
\addplot [semithick, white!18.8235294117647!black, forget plot]
table {%
-0.093125 0.638179898262024
-0.031875 0.638179898262024
};
\addplot [semithick, white!18.8235294117647!black, forget plot]
table {%
-0.093125 0.801881611347198
-0.031875 0.801881611347198
};
\addplot [semithick, white!18.8235294117647!black, forget plot]
table {%
0.0625 0.758135139942169
0.0625 0.707632660865784
};
\addplot [semithick, white!18.8235294117647!black, forget plot]
table {%
0.0625 0.819619834423065
0.0625 0.842484056949615
};
\addplot [semithick, white!18.8235294117647!black, forget plot]
table {%
0.031875 0.707632660865784
0.093125 0.707632660865784
};
\addplot [semithick, white!18.8235294117647!black, forget plot]
table {%
0.031875 0.842484056949615
0.093125 0.842484056949615
};
\addplot [semithick, white!18.8235294117647!black, forget plot]
table {%
0.1875 0.713327586650848
0.1875 0.653121650218964
};
\addplot [semithick, white!18.8235294117647!black, forget plot]
table {%
0.1875 0.781869620084763
0.1875 0.80782550573349
};
\addplot [semithick, white!18.8235294117647!black, forget plot]
table {%
0.156875 0.653121650218964
0.218125 0.653121650218964
};
\addplot [semithick, white!18.8235294117647!black, forget plot]
table {%
0.156875 0.80782550573349
0.218125 0.80782550573349
};
\addplot [semithick, white!18.8235294117647!black, forget plot]
table {%
0.8125 0.749334692955017
0.8125 0.677623510360718
};
\addplot [semithick, white!18.8235294117647!black, forget plot]
table {%
0.8125 0.801014214754105
0.8125 0.857260525226593
};
\addplot [semithick, white!18.8235294117647!black, forget plot]
table {%
0.781875 0.677623510360718
0.843125 0.677623510360718
};
\addplot [semithick, white!18.8235294117647!black, forget plot]
table {%
0.781875 0.857260525226593
0.843125 0.857260525226593
};
\addplot [semithick, white!18.8235294117647!black, forget plot]
table {%
0.9375 0.671752274036407
0.9375 0.565369069576263
};
\addplot [semithick, white!18.8235294117647!black, forget plot]
table {%
0.9375 0.74355810880661
0.9375 0.808627247810364
};
\addplot [semithick, white!18.8235294117647!black, forget plot]
table {%
0.906875 0.565369069576263
0.968125 0.565369069576263
};
\addplot [semithick, white!18.8235294117647!black, forget plot]
table {%
0.906875 0.808627247810364
0.968125 0.808627247810364
};
\addplot [semithick, white!18.8235294117647!black, forget plot]
table {%
1.0625 0.733120203018188
1.0625 0.655643284320831
};
\addplot [semithick, white!18.8235294117647!black, forget plot]
table {%
1.0625 0.787484765052795
1.0625 0.847956538200378
};
\addplot [semithick, white!18.8235294117647!black, forget plot]
table {%
1.031875 0.655643284320831
1.093125 0.655643284320831
};
\addplot [semithick, white!18.8235294117647!black, forget plot]
table {%
1.031875 0.847956538200378
1.093125 0.847956538200378
};
\addplot [semithick, white!18.8235294117647!black, forget plot]
table {%
1.1875 0.682084739208221
1.1875 0.605243861675262
};
\addplot [semithick, white!18.8235294117647!black, forget plot]
table {%
1.1875 0.745677918195724
1.1875 0.81328672170639
};
\addplot [semithick, white!18.8235294117647!black, forget plot]
table {%
1.156875 0.605243861675262
1.218125 0.605243861675262
};
\addplot [semithick, white!18.8235294117647!black, forget plot]
table {%
1.156875 0.81328672170639
1.218125 0.81328672170639
};
\addplot [semithick, white!18.8235294117647!black, forget plot]
table {%
1.8125 0.719272881746292
1.8125 0.671376824378967
};
\addplot [semithick, white!18.8235294117647!black, forget plot]
table {%
1.8125 0.769477516412735
1.8125 0.827174305915833
};
\addplot [semithick, white!18.8235294117647!black, forget plot]
table {%
1.781875 0.671376824378967
1.843125 0.671376824378967
};
\addplot [semithick, white!18.8235294117647!black, forget plot]
table {%
1.781875 0.827174305915833
1.843125 0.827174305915833
};
\addplot [semithick, white!18.8235294117647!black, forget plot]
table {%
1.9375 0.630230993032455
1.9375 0.557209074497223
};
\addplot [semithick, white!18.8235294117647!black, forget plot]
table {%
1.9375 0.711030513048172
1.9375 0.772097587585449
};
\addplot [semithick, white!18.8235294117647!black, forget plot]
table {%
1.906875 0.557209074497223
1.968125 0.557209074497223
};
\addplot [semithick, white!18.8235294117647!black, forget plot]
table {%
1.906875 0.772097587585449
1.968125 0.772097587585449
};
\addplot [semithick, white!18.8235294117647!black, forget plot]
table {%
2.0625 0.69707328081131
2.0625 0.651367604732513
};
\addplot [semithick, white!18.8235294117647!black, forget plot]
table {%
2.0625 0.753904789686203
2.0625 0.816057324409485
};
\addplot [semithick, white!18.8235294117647!black, forget plot]
table {%
2.031875 0.651367604732513
2.093125 0.651367604732513
};
\addplot [semithick, white!18.8235294117647!black, forget plot]
table {%
2.031875 0.816057324409485
2.093125 0.816057324409485
};
\addplot [semithick, white!18.8235294117647!black, forget plot]
table {%
2.1875 0.644421130418777
2.1875 0.590901732444763
};
\addplot [semithick, white!18.8235294117647!black, forget plot]
table {%
2.1875 0.709492892026901
2.1875 0.776102125644684
};
\addplot [semithick, white!18.8235294117647!black, forget plot]
table {%
2.156875 0.590901732444763
2.218125 0.590901732444763
};
\addplot [semithick, white!18.8235294117647!black, forget plot]
table {%
2.156875 0.776102125644684
2.218125 0.776102125644684
};
\addplot [semithick, white!18.8235294117647!black, forget plot]
table {%
2.8125 0.714768916368484
2.8125 0.666869223117828
};
\addplot [semithick, white!18.8235294117647!black, forget plot]
table {%
2.8125 0.768564850091934
2.8125 0.822218775749207
};
\addplot [semithick, white!18.8235294117647!black, forget plot]
table {%
2.781875 0.666869223117828
2.843125 0.666869223117828
};
\addplot [semithick, white!18.8235294117647!black, forget plot]
table {%
2.781875 0.822218775749207
2.843125 0.822218775749207
};
\addplot [semithick, white!18.8235294117647!black, forget plot]
table {%
2.9375 0.629082053899765
2.9375 0.543520510196686
};
\addplot [semithick, white!18.8235294117647!black, forget plot]
table {%
2.9375 0.707579404115677
2.9375 0.76716148853302
};
\addplot [semithick, white!18.8235294117647!black, forget plot]
table {%
2.906875 0.543520510196686
2.968125 0.543520510196686
};
\addplot [semithick, white!18.8235294117647!black, forget plot]
table {%
2.906875 0.76716148853302
2.968125 0.76716148853302
};
\addplot [semithick, white!18.8235294117647!black, forget plot]
table {%
3.0625 0.680557519197464
3.0625 0.63948667049408
};
\addplot [semithick, white!18.8235294117647!black, forget plot]
table {%
3.0625 0.753449857234955
3.0625 0.810884237289429
};
\addplot [semithick, white!18.8235294117647!black, forget plot]
table {%
3.031875 0.63948667049408
3.093125 0.63948667049408
};
\addplot [semithick, white!18.8235294117647!black, forget plot]
table {%
3.031875 0.810884237289429
3.093125 0.810884237289429
};
\addplot [semithick, white!18.8235294117647!black, forget plot]
table {%
3.1875 0.617027103900909
3.1875 0.587015211582184
};
\addplot [semithick, white!18.8235294117647!black, forget plot]
table {%
3.1875 0.71022042632103
3.1875 0.769392728805542
};
\addplot [semithick, white!18.8235294117647!black, forget plot]
table {%
3.156875 0.587015211582184
3.218125 0.587015211582184
};
\addplot [semithick, white!18.8235294117647!black, forget plot]
table {%
3.156875 0.769392728805542
3.218125 0.769392728805542
};
\addplot [semithick, white!18.8235294117647!black, forget plot]
table {%
3.8125 0.669158279895782
3.8125 0.574838578701019
};
\addplot [semithick, white!18.8235294117647!black, forget plot]
table {%
3.8125 0.761750340461731
3.8125 0.809472620487213
};
\addplot [semithick, white!18.8235294117647!black, forget plot]
table {%
3.781875 0.574838578701019
3.843125 0.574838578701019
};
\addplot [semithick, white!18.8235294117647!black, forget plot]
table {%
3.781875 0.809472620487213
3.843125 0.809472620487213
};
\addplot [semithick, white!18.8235294117647!black, forget plot]
table {%
3.9375 0.60126057267189
3.9375 0.51898330450058
};
\addplot [semithick, white!18.8235294117647!black, forget plot]
table {%
3.9375 0.69141098856926
3.9375 0.751423835754395
};
\addplot [semithick, white!18.8235294117647!black, forget plot]
table {%
3.906875 0.51898330450058
3.968125 0.51898330450058
};
\addplot [semithick, white!18.8235294117647!black, forget plot]
table {%
3.906875 0.751423835754395
3.968125 0.751423835754395
};
\addplot [semithick, white!18.8235294117647!black, forget plot]
table {%
4.0625 0.642528027296066
4.0625 0.54739785194397
};
\addplot [semithick, white!18.8235294117647!black, forget plot]
table {%
4.0625 0.743139088153839
4.0625 0.797571301460266
};
\addplot [semithick, white!18.8235294117647!black, forget plot]
table {%
4.031875 0.54739785194397
4.093125 0.54739785194397
};
\addplot [semithick, white!18.8235294117647!black, forget plot]
table {%
4.031875 0.797571301460266
4.093125 0.797571301460266
};
\addplot [semithick, white!18.8235294117647!black, forget plot]
table {%
4.1875 0.593625158071518
4.1875 0.504925131797791
};
\addplot [semithick, white!18.8235294117647!black, forget plot]
table {%
4.1875 0.696036100387573
4.1875 0.754909157752991
};
\addplot [semithick, white!18.8235294117647!black, forget plot]
table {%
4.156875 0.504925131797791
4.218125 0.504925131797791
};
\addplot [semithick, white!18.8235294117647!black, forget plot]
table {%
4.156875 0.754909157752991
4.218125 0.754909157752991
};
\addplot [semithick, white!18.8235294117647!black, forget plot]
table {%
-0.24875 0.800015032291412
-0.12625 0.800015032291412
};
\addplot [semithick, white!18.8235294117647!black, forget plot]
table {%
-0.12375 0.739399075508118
-0.00125 0.739399075508118
};
\addplot [semithick, white!18.8235294117647!black, forget plot]
table {%
0.00125 0.786243557929993
0.12375 0.786243557929993
};
\addplot [semithick, white!18.8235294117647!black, forget plot]
table {%
0.12625 0.74202561378479
0.24875 0.74202561378479
};
\addplot [semithick, white!18.8235294117647!black, forget plot]
table {%
0.75125 0.768215119838715
0.87375 0.768215119838715
};
\addplot [semithick, white!18.8235294117647!black, forget plot]
table {%
0.87625 0.708528459072113
0.99875 0.708528459072113
};
\addplot [semithick, white!18.8235294117647!black, forget plot]
table {%
1.00125 0.753133416175842
1.12375 0.753133416175842
};
\addplot [semithick, white!18.8235294117647!black, forget plot]
table {%
1.12625 0.70920205116272
1.24875 0.70920205116272
};
\addplot [semithick, white!18.8235294117647!black, forget plot]
table {%
1.75125 0.759525179862976
1.87375 0.759525179862976
};
\addplot [semithick, white!18.8235294117647!black, forget plot]
table {%
1.87625 0.686484694480896
1.99875 0.686484694480896
};
\addplot [semithick, white!18.8235294117647!black, forget plot]
table {%
2.00125 0.742865443229675
2.12375 0.742865443229675
};
\addplot [semithick, white!18.8235294117647!black, forget plot]
table {%
2.12625 0.698873043060303
2.24875 0.698873043060303
};
\addplot [semithick, white!18.8235294117647!black, forget plot]
table {%
2.75125 0.747528731822968
2.87375 0.747528731822968
};
\addplot [semithick, white!18.8235294117647!black, forget plot]
table {%
2.87625 0.679879903793335
2.99875 0.679879903793335
};
\addplot [semithick, white!18.8235294117647!black, forget plot]
table {%
3.00125 0.729218423366547
3.12375 0.729218423366547
};
\addplot [semithick, white!18.8235294117647!black, forget plot]
table {%
3.12625 0.682448983192444
3.24875 0.682448983192444
};
\addplot [semithick, white!18.8235294117647!black, forget plot]
table {%
3.75125 0.73406994342804
3.87375 0.73406994342804
};
\addplot [semithick, white!18.8235294117647!black, forget plot]
table {%
3.87625 0.671289384365082
3.99875 0.671289384365082
};
\addplot [semithick, white!18.8235294117647!black, forget plot]
table {%
4.00125 0.71230274438858
4.12375 0.71230274438858
};
\addplot [semithick, white!18.8235294117647!black, forget plot]
table {%
4.12625 0.661094546318054
4.24875 0.661094546318054
};
\end{axis}

\end{tikzpicture}

%% file: contents/06_conclusion.tex

We have proposed LULA units: hidden units associated with partially zero weights that can be added to any pre-trained MAP network for the purpose of exclusively tuning the uncertainty of a Laplace approximation without affecting its predictive performance. The crux of LULA is the observation that these units induce additional dimensions and thus degrees of freedom in the network's parameter space that do not affect the network output. However, these additional parameters \emph{do} non-trivially affect the curvature of the loss landscape and therefore the covariance matrices of Laplace approximations. Because of this, LULA units are indeed ``uncertainty units''. They can, moreover, be trained via an objective that depends on both inlier and outlier datasets to calibrate the network's predictive uncertainty estimates. We show empirically that LULA provides a cheap yet effective \emph{post-hoc} uncertainty tuning for Laplace approximations.

%% file: contents/91_proofs.tex

\begin{proposition}[Properties]
    Let $f: \R^n \times \R^d \to \R^k$ be a MAP-trained $L$-layer network under dataset $\D$, and let $\theta_\map$ be the MAP estimate. Suppose $\widetilde{f}: \R^n \times \R^{\widetilde{d}} \to \R$ and $\widetilde{\theta}_\map \in \R^{\widetilde{d}}$ are obtained via the previous construction, and $\widetilde{\L}$ is the resulting loss function under $\widetilde{f}$.
    \begin{enumerate}[label=(\alph*)]
        \item For an arbitrary input $x \in \R^n$, we have $\widetilde{f}(x; \widetilde{\theta}_\map) = f(x; \theta_\map)$.
        \item The gradient of $\widetilde{\L}$ w.r.t. the additional weights in $\widetilde{W}^{(L)}$ is non-linear in $\widehat{\theta}$.
    \end{enumerate}
\end{proposition}
\begin{proof}
    As the first order of business, for each layer $l = 1, \dots, L$ we denote the hidden units and pre-activations of $\widetilde{f}$ as $\widetilde{h}^{(l)}$ and $\widetilde{a}^{(l)}$, respectively.

    We begin with (a). Let $x \in \R^n$ be arbitrary. We need to show that the output of $\widetilde{f}$, i.e.~the last pre-activations $\widetilde{a}^{(L)}$, is equal to the last pre-activations $a^{(L)}$ of $f$. For the first layer, we have that
    \begin{equation} \label{eq:forward_pass_first}
        \begin{aligned}
            \widetilde{a}^{(1)} &= \widetilde{W}^{(1)} x + \widetilde{b}^{(1)} \\
                &= \begin{pmatrix}
                    W^{(1)} \\
                    \widehat{W}_1^{(1)}
                \end{pmatrix} x +
                \begin{pmatrix}
                    b^{(1)} \\
                    \widehat{b}^{(1)}
                \end{pmatrix} \\
                &= \begin{pmatrix}
                    W^{(1)} x + b^{(1)} \\
                    \widehat{W}_1^{(1)} x + \widehat{b}^{(1)}
                \end{pmatrix} =:
                \begin{pmatrix}
                    a^{(1)} \\
                    \widehat{a}^{(1)}
                \end{pmatrix} \, .
        \end{aligned}
    \end{equation}
    For every layer $l = 1, \dots, L-1$, we denote the hidden units as the block vector
    \begin{equation*}
        \widetilde{h}^{(l)} =
        \begin{pmatrix}
            \varphi(a^{(l)}) \\
            \varphi(\widehat{a}^{(l)})
        \end{pmatrix}
        =
        \begin{pmatrix}
            h^{(l)} \\
            \widehat{h}^{(l)}
        \end{pmatrix}  \, .
    \end{equation*}
    Now, for the intermediate layer $l = 2, \dots, L-1$, we observe that
    \begin{equation} \label{eq:forward_pass_intermediate}
        \begin{aligned}
            \widetilde{a}^{(l)} &= \widetilde{W}^{(l)} \widetilde{h}^{(l-1)} + \widetilde{b}^{(l)} \\
                &= \begin{pmatrix}
                    W^{(l)} & 0 \\
                    \widehat{W}_1^{(l)} & \widehat{W}_2^{(l)}
                \end{pmatrix}
                \begin{pmatrix}
                    h^{(l-1)} \\
                    \widehat{h}^{(l-1)}
                \end{pmatrix}
                +
                \begin{pmatrix}
                    b^{(l)} \\
                    \widehat{b}^{(l)}
                \end{pmatrix}  \\
                &=
                \begin{pmatrix}
                    W^{(l)} h^{(l-1)} + 0 + b^{(l)} \\
                    \widehat{W}_1^{(l)} h^{(l-1)} + \widehat{W}_2^{(l)} \widehat{h}^{(l-1)} + \widehat{b}^{(l)}
                \end{pmatrix}
                =:
                \begin{pmatrix}
                    a^{(l)} \\
                    \widehat{a}^{(l)}
                \end{pmatrix} \, .
        \end{aligned}
    \end{equation}
    Finally, for the last layer, we get
    \begin{equation} \label{eq:forward_pass_last}
        \begin{aligned}
            \widetilde{a}^{(L)} &= \widetilde{W}^{(L)} x + \widetilde{b}^{(L)} \\
                &= \begin{pmatrix}
                    W^{(L)} & 0
                \end{pmatrix}
                \begin{pmatrix}
                    h^{(L-1)} \\
                    \widehat{h}^{(L-1)}
                \end{pmatrix}
                + b^{(L)} \\
                &= W^{(L)} h^{(L-1)} + 0 + b^{(L)} \\
                &= a^{(L)} \, ,
        \end{aligned}
    \end{equation}
    and thus we have the desired invariance.

    For part (b), we denote the additional (zero) weights in $\widetilde{W}^{(L)}$ by $\widehat{W}^{(L)}$. It is clear from \eqref{eq:forward_pass_last} that the gradient $\nabla_{\widehat{W}^{(L)}}\, a^{(L)}$ is given by $\widehat{h}^{(L-1)}$. Hence, by chain rule we have
    \begin{align*}
        \nabla_{\widehat{W}^{(L)}} \widetilde{\L} &= (\nabla_{a^{(L)}}\, \widetilde{\L}) \, (\nabla_{\widehat{W}^{(L)}}\, a^{(L)}) \\
            &= (\nabla_{a^{(L)}}\, \widetilde{\L}) \, \widehat{h}^{(L-1)} \, .
    \end{align*}
    By observing \eqref{eq:forward_pass_first} and \eqref{eq:forward_pass_intermediate}, along the fact that the non-linearity $\varphi$ is used in the forward pass, it is clear that $\widehat{h}^{(L-1)}$ is non-linear in $\widehat{\theta} = (\widehat{W}_1^{(1)}, \widehat{b}^{(1)}, \dots, \widehat{W}_1^{(L-1)}, \widehat{W}_2^{(L-1)}, \widehat{b}^{(L-1)})$ and therefore $\nabla_{\widehat{W}^{(L)}} \widetilde{\L}$ also is.
\end{proof}

\vspace{1em}

\begin{proposition}[Predictive Uncertainty]
      Suppose $f: \R^n \times \R^d \to \R$ is a real-valued network and $\widetilde{f}$ is as constructed above. Suppose further that diagonal Laplace-approximated posteriors $\N(\theta_\map, \diag{\sigma})$, $\N(\widetilde{\theta}_\map, \diag{\widetilde{\sigma}})$ are employed for $f$ and $\widetilde{f}$, respectively. Under the linearization \eqref{eq:linearization}, for any input $x \in \R^n$, the variance over the output $\widetilde{f}(x; \widetilde{\theta})$ is at least that of $f(x; \theta)$.
\end{proposition}
\begin{proof}
    Let us denote the random variable taking values in the augmented parameter space by $\widetilde{\theta}$. W.l.o.g. we re-arrange $\widetilde{\theta}$ as $( \theta^\top, \widehat{\theta}^\top )^\top$ where $\widehat{\theta} \in \R^{\widetilde{d} - d}$ contains the weights corresponding to the the additional LULA units. If $g(x)$ is the gradient of the output $f(x; \theta)$ w.r.t. $\theta$ at $\theta_\map$, then the gradient of $\widetilde{f}(x; \widetilde{\theta})$ w.r.t. $\widetilde{\theta}$ at $\widetilde{\theta}_\map$, say $\widetilde{g}(x)$, can be written as the concatenation $(g(x)^\top, \widehat{g}(x)^\top)^\top$ where $\widehat{g}(x)$ is the corresponding gradient w.r.t. $\widehat{\theta}$. Furthermore, $\diag{\widetilde{\sigma}}$ has diagonal elements
    \begin{align*}
        \left(\sigma_{11}, \dots, \sigma_{dd}, \widehat{\sigma}_{11}, \dots, \widehat{\sigma}_{\widetilde{d}-d, \widetilde{d}-d} \right)^\top =: ( \sigma^\top, \widehat{\sigma}^\top )^\top \, .
    \end{align*}
    Let $x \in \R^n$ be an arbitrary input. Denoting the output variance of $\widetilde{f}(x; \widetilde{\theta})$ by $\widetilde{v}(x)$, we have
    \begin{align*}
        \widetilde{v}(x) &= \widetilde{g}(x)^\top \diag{\widetilde{\sigma}} \widetilde{g}(x) \\
                &= \underbrace{g(x)^\top \diag{\sigma} g(x)}_{= v(x)} + \widehat{g}(x)^\top \diag{\widehat{\sigma}} \widehat{g}(x) \nonumber \\
                &\geq v(x) \, ,
    \end{align*}
    since $\diag{\widehat{\sigma}}$ is positive-definite by definition.
\end{proof}

%% file: contents/92_further_details.tex

\begin{algorithm}[t!]
    \caption{Adding LULA units.}
    \label{algo:lula_construction}
    \setstretch{1.5}
    \begin{algorithmic}[1]
        \Require
            \Statex $L$-layer net with a MAP estimate $\theta_\map = (W^{(l)}_\map, b^{(l)}_\map)_{l=1}^L$. Sequence of non-negative integers $(m_l)_{l=1}^L$.

        \For{$l=1, \dots, L-1$}
            \State $\mathrm{vec}\,\widehat{W}_1^{(l)} \sim p(\mathrm{vec}\,\widehat{W}_1^{(l)})$   \Comment{Draw from a prior}
            \State $\mathrm{vec}\,\widehat{W}_2^{(l)} \sim p(\mathrm{vec}\,\widehat{W}_2^{(l)})$   \Comment{Draw from a prior}
            \State $\widehat{b}^{(l)} \sim p(\widehat{b}^{(l)})$    \Comment{Draw from a prior}
            \State $\widetilde{W}^{(l)}_\map = \begin{pmatrix}
                W^{(l)}_\map & 0 \\[5pt]
                \widehat{W}_1^{(l)}  & \widehat{W}_2^{(l)} \\
            \end{pmatrix}$      \Comment{$0 \in \R^{n_l \times m_{l-1}}$}
            \State $\widetilde{b}^{(l)}_\map := \begin{pmatrix}
                b^{(l)}_\map \\[5pt]
                \widehat{b}^{(l)}
            \end{pmatrix}$
        \EndFor
        \State $\widetilde{W}^{(L)}_\map = (W^{(L)}_\map, 0)$       \Comment{$0 \in \R^{k \times m_{L-1}}$}
        \State $\widetilde{b}^{(L)}_\map = b^{(L)}_\map$
        \State $\widetilde{\theta}_\map = (\widetilde{W}^{(l)}_\map, \widetilde{b}^{(l)}_\map )_{l=1}^L$
        \State \Return $\widetilde{\theta}_\map$
    \end{algorithmic}
\end{algorithm}

We summarize the augmentation of a network with LULA units in \cref{algo:lula_construction}. Note that the priors of the free parameters $\widehat{W}^{(l)}$, $\widehat{b}^{(l)}$ (lines 2 and 3) can be chosen as independent Gaussians---this reflects the standard procedure for initializing NNs' parameters.

%% file: contents/94_further_exps.tex

\subsection{Toy Dataset}

\begin{figure*}[t]
  \centering

  \subfloat{\includegraphics[width=0.3\textwidth, height=0.08\textheight]{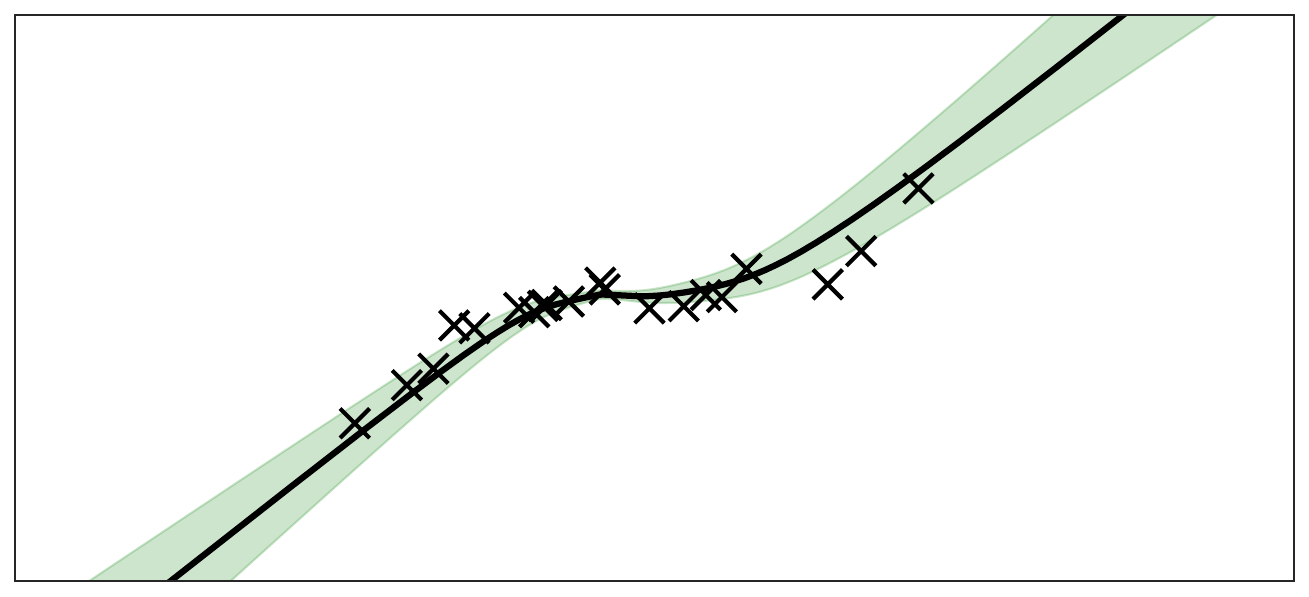}}
  \hfill
  \subfloat{\includegraphics[width=0.3\textwidth, height=0.08\textheight]{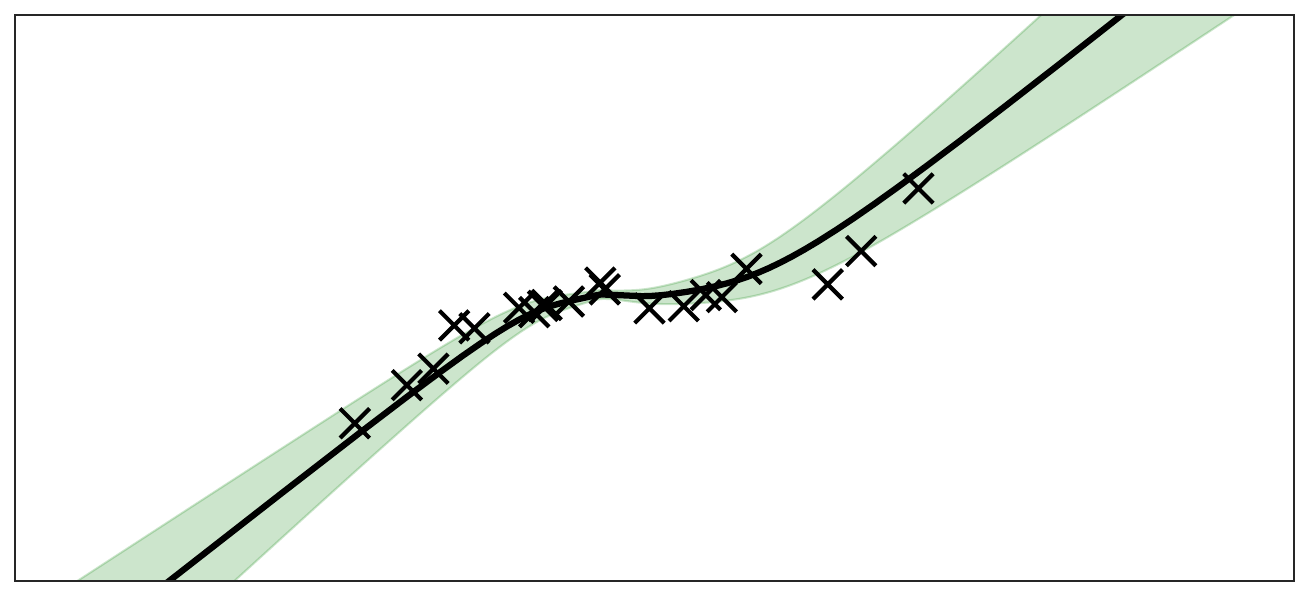}}
  \hfill
  \subfloat{\includegraphics[width=0.3\textwidth, height=0.08\textheight]{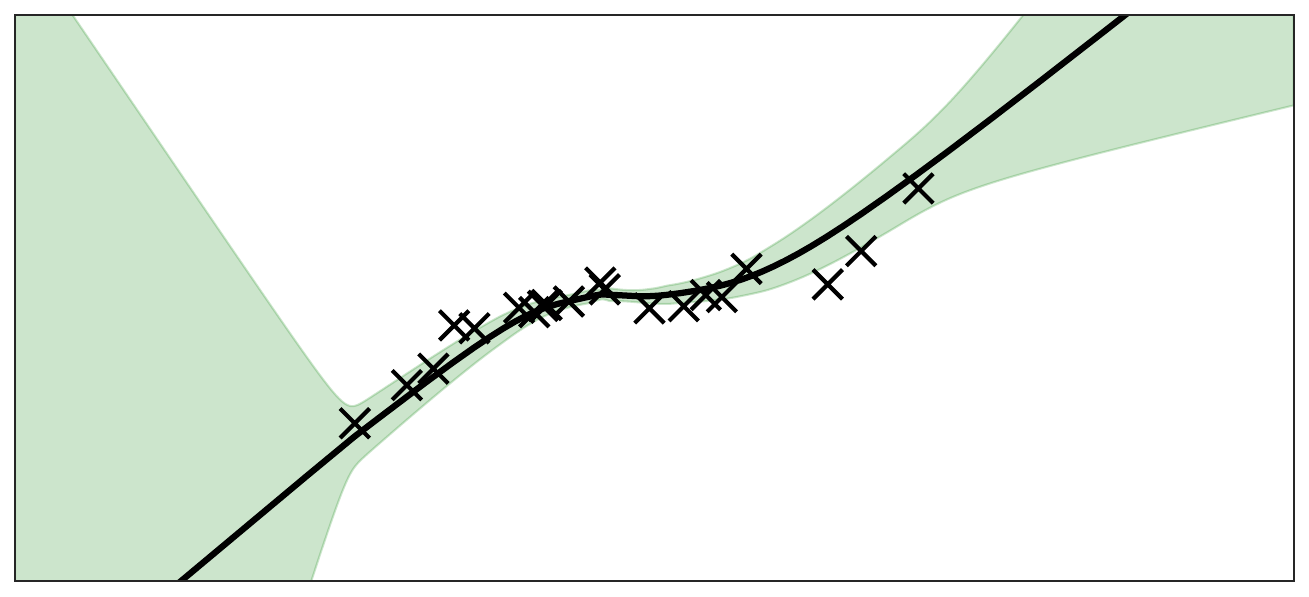}}
  \hfill
  \hphantom{\subfloat{\includegraphics[width=0.032\textwidth, height=0.08\textheight]{toy_moon_colorbar}}}

  \setcounter{subfigure}{0}

  \subfloat[Laplace]{\includegraphics[width=0.3\textwidth, height=0.08\textheight]{toy_moon_laplace}}
  \hfill
  \subfloat[LULA-Untrained]{\includegraphics[width=0.3\textwidth, height=0.08\textheight]{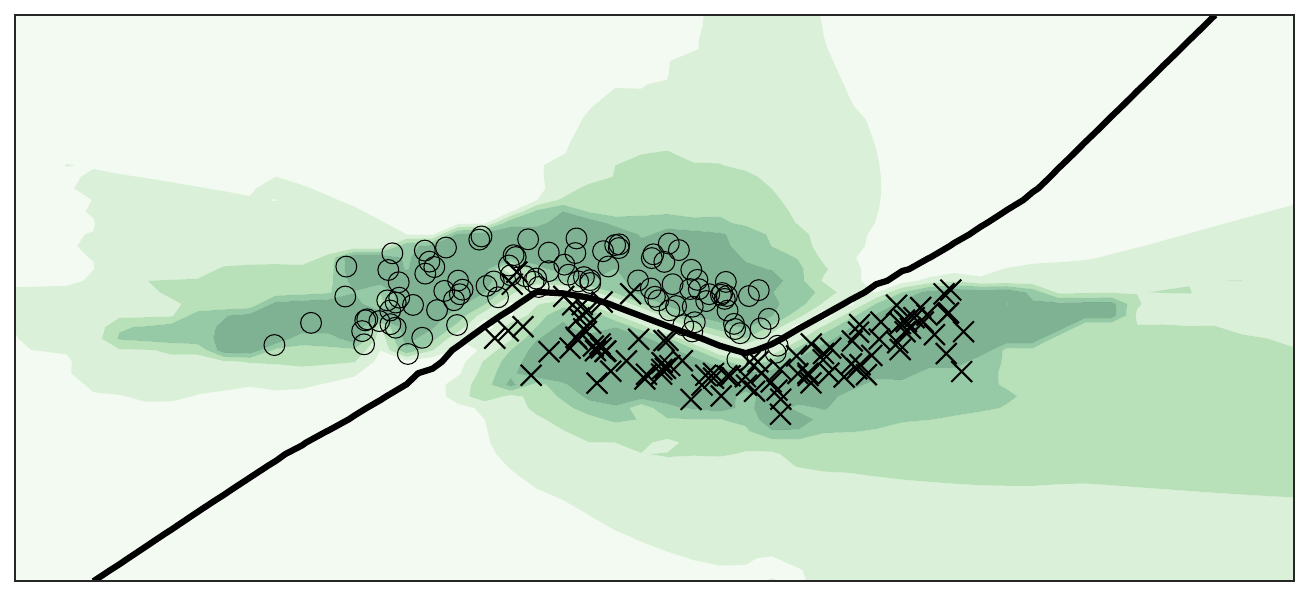}}
  \hfill
  \subfloat[LULA-Trained]{\includegraphics[width=0.3\textwidth, height=0.08\textheight]{toy_moon_lula_full_la}}
  \quad
  \subfloat{\includegraphics[width=0.032\textwidth, height=0.08\textheight]{toy_moon_colorbar}}

  \caption{The effect of LULA training.}

  \label{fig:lula_training}
\end{figure*}

To show the effectiveness of LULA training, we compare the standard Laplace, untrained LULA, and trained LULA in \cref{fig:lula_training}. As predicted by \cref{prop:variance_guarantee}, untrained LULA increases predictive uncertainty estimates. However, this increase of uncertainty is not well-adapted to the data (b). Training $\widetilde{\theta}$ using \eqref{eq:obj_exact} make it more calibrated to both inliers and outliers (c).

\subsection{UCI Regression}
\label{subsec:uci_reg_uq}

\begin{table}[t]
  \caption{UQ performances on UCI datasets. Values are the average (over all data points and ten training-prediction trials) predictive standard deviations, i.e.~the standard deviation of the Gaussian \eqref{eq:linearized_dist}. Lower is better for test data and vice-versa for outliers. By definition, MAP does not have (epistemic) uncertainty.}
  \label{tab:uci_regression_ood}

  \centering
  \footnotesize
  \setlength{\tabcolsep}{7pt}

  \resizebox{\linewidth}{!}{
    \begin{tabular}{lrr >{\columncolor{tablegray}}rrr >{\columncolor{tablegray}}r}
      \toprule

      & \multicolumn{3}{c}{\textbf{Test set} $\downarrow$} & \multicolumn{3}{c}{\textbf{Outliers} $\uparrow$} \\
      \cmidrule(r){2-4} \cmidrule(l){5-7}

      \textbf{Dataset} & \textbf{DE} & \textbf{LA} & \textbf{LA-LULA} & \textbf{DE} & \textbf{LA} & \textbf{LA-LULA} \\

      \midrule

      Housing & 5.82 & {1.26} & 1.37 & 145.33 & 222.76 & {377.92} \\
      Concrete & 8.11 & 10.44 & 16.89 & 964.63 & 30898.92 & {83241.42} \\
      Energy & 4.40 & 1.05 & 1.08 & 126.11 & 1070.09 & {5163.53} \\
      Kin8nm & 0.10 & 0.14 & 0.18 & {2.12} & 0.80 & {2.12} \\
      Power & 19.85 & {2.85} & 3.20 & 12235.87 & 4148.98 & {221287.80} \\
      Wine & 0.64 & 1.15 & 1.22 & 28.57 & 186.76 & {21383.17} \\
      Yacht & 5.17 & 2.08 & 2.78 & 187.41 & 5105.69 & {13119.99} \\

      \bottomrule
    \end{tabular}
  }

\end{table}

\begin{table}[t]
  \caption{Predictive performances on UCI regression datasets in terms of average test log-likelihood. The numbers reported are averages over ten training-prediction runs along with the corresponding standard deviations. The performances of LULA are similar to LA's. The differences between their exact values are likely due to MC-integration.}
  \label{tab:uci_regression}

  \centering
  \footnotesize
  \setlength{\tabcolsep}{7pt}

  \resizebox{\linewidth}{!}{
    \begin{tabular}{lrrr >{\columncolor{tablegray}}r}
      \toprule
      \textbf{Dataset} & \textbf{MAP} & \textbf{DE} & \textbf{LA} & \textbf{LA-LULA} \\

      \midrule

      Housing & -2.794$\pm$0.012 & -3.045$\pm$0.009 & -3.506$\pm$0.055 & -3.495$\pm$0.047 \\
      Concrete & -3.409$\pm$0.036 & -3.951$\pm$0.062 & -4.730$\pm$0.205 & -4.365$\pm$0.094 \\
      Energy & -2.270$\pm$0.128 & -2.673$\pm$0.015 & -2.707$\pm$0.030 & -2.698$\pm$0.014 \\
      Kin8nm & -0.923$\pm$0.000 & 1.086$\pm$0.022 & -0.965$\pm$0.003 & -0.969$\pm$0.003 \\
      Power & -3.154$\pm$0.002 & -54.804$\pm$7.728 & -3.273$\pm$0.015 & -3.277$\pm$0.024 \\
      Wine & -1.190$\pm$0.014 & -1.038$\pm$0.018 & -1.624$\pm$0.075 & -1.630$\pm$0.092 \\
      Yacht & -1.835$\pm$0.053 & -3.272$\pm$0.079 & -2.509$\pm$0.367 & -2.663$\pm$0.276 \\

      \bottomrule
    \end{tabular}
  }
\end{table}

To validate the performance of LULA in regressions, we employ a subset of the UCI regression benchmark datasets. Following previous works, the network architecture used here is a single-hidden-layer ReLU network with $50$ hidden units. The data are standardized to have zero mean and unit variance. We use $50$ LULA units and optimize them for $40$ epochs using OOD data sampled uniformly from $[-10, 10]^n$. For LA and LULA, each prediction is done via MC-integration with $100$ samples. For the evaluation of each dataset, we use a $60$-$20$-$20$ train-validation-test split. We repeat each train-test process $10$ times and take the average.

In \cref{tab:uci_regression_ood} we report the average predictive standard deviation for each dataset. Note that this metric is the direct generalization of the 1D uncertainty estimates in \cref{fig:one} to multi-dimension. The test outliers are sampled uniformly from $[-10,10]^n$. Note that since the inlier data are centered around the origin and have unit variance, they lie approximately in a Euclidean ball with a radius of $2$. Therefore, these outliers are far away from them. Thus, naturally, high uncertainty values over these outliers are desirable. Uncertainties over the test sets are generally low for all methods, although LULA has slightly higher uncertainties compared to the base LA. However, LULA yield much higher uncertainties over outliers across all datasets, significantly more than the baselines. Moreover, in \cref{tab:uci_regression}, we show that LULA maintains the predictive performance of the base LA. Altogether, they imply that LULA can detect outliers better than other methods without costing the predictive performance.

\begin{figure}[t]
  \centering

  \def\figcifarcappendixwidth{0.45\textwidth}
  \def\figcifarcappendixheight{0.2\textheight}

  \hspace{2em}
  \subfloat{\input{figs/cifar10c_ece}}

  \hspace{2em}
  \subfloat{\input{figs/cifar10c_brier}}

  \caption{Summarized ECE and Brier score at each severity level of the CIFAR-10-C dataset.}
  \label{fig:cifar10c_appendix}
\end{figure}
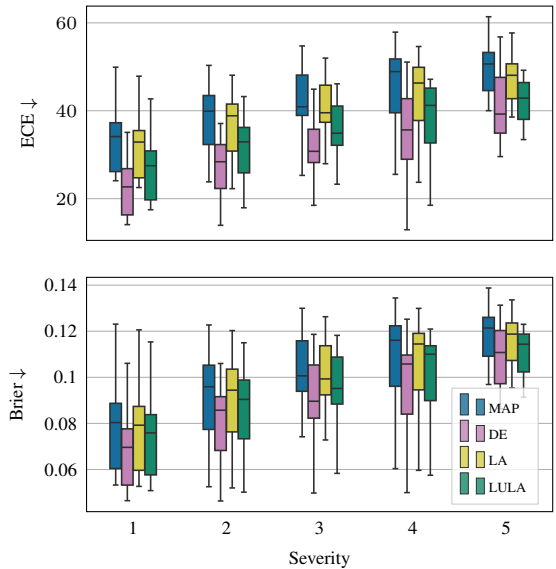

\subsection{Image Classification}
\label{appendix:image_clf}

To complement \cref{fig:cifar10c}, we present the ECE and Brier score results on CIFAR-10-C in \cref{fig:cifar10c_appendix}. As observed in the main text, LULA consistently improves the base LA. Furthermore, LULA is competitive to the state-of-the-art DE, especially in higher severity levels.

We furthermore present the detailed results on OOD detection in terms of MMC, FPR95 (\cref{tab:mmc,tab:fpr95}), and additionally area under ROC (AUROC) and precision-recall (AUPRC) curves (\cref{tab:auroc,tab:auprc}). We use standard datasets: EMNIST, KMNIST, FMNIST, and LSUN. Furthermore, we use the following artificial datasets:
\begin{itemize}
  \item \textsc{GrayCIFAR10}: obtained by converting CIFAR-10 test data into grayscale images.
  \item \textsc{UniformNoise}: obtained by uniformly sampling from the hypercube $[0, 1]^n$.
  \item \textsc{SmoothedNoise}: obtained by permuting, blurring, and contrast re-scaling the original test images \citep{hein2019relu}.
  \item \textsc{FMNIST3D}: obtained by converting the grayscale FMNIST images into 3-channel images.
\end{itemize}
We observe that LULA consistently improves the base LA. Especially, LULA makes the confidence estimates over OOD data lower without introducing underconfidence on in-distribution data.

\begin{table*}[tb]
  \caption{Detailed MMC results. Values are averages over five prediction runs.}
  \label{tab:mmc}

  \centering
  \scriptsize
  \setlength{\tabcolsep}{7pt}

  \resizebox{\textwidth}{!}{
    \begin{tabular}{lccccc >{\columncolor{tablegray}}c c >{\columncolor{tablegray}}c c >{\columncolor{tablegray}}c}

      \toprule

      \textbf{Dataset} & \textbf{MAP} & \textbf{MAP-Temp} & \textbf{DE} & \textbf{DE-Temp} & \textbf{LA} & \textbf{LA-LULA} & \textbf{LLLA} & \textbf{LLLA-LULA} & \textbf{OE} & \textbf{OE-LULA}  \\

      \midrule

      \textbf{MNIST} & 99.8 & 99.8$\pm$0.0 & 99.7 & 99.8$\pm$0.0 & 99.7$\pm$0.0 & 98.3$\pm$0.0 & 99.3$\pm$0.0 & 99.2$\pm$0.0 & 99.4$\pm$0.0 & 71.5$\pm$0.4 \\
      EMNIST & 84.6 & 86.2$\pm$0.0 & 82.8 & 87.4$\pm$0.0 & 84.1$\pm$0.0 & 56.6$\pm$0.5 & 73.7$\pm$0.2 & 67.4$\pm$0.3 & 81.2$\pm$0.0 & 31.7$\pm$0.2 \\
      KMNIST & 71.3 & 73.8$\pm$0.0 & 67.8 & 76.1$\pm$0.0 & 70.5$\pm$0.0 & 34.7$\pm$0.4 & 56.4$\pm$0.3 & 45.6$\pm$0.5 & 66.5$\pm$0.0 & 23.7$\pm$0.1 \\
      FMNIST & 76.7 & 79.0$\pm$0.0 & 69.6 & 80.1$\pm$0.0 & 75.7$\pm$0.0 & 37.6$\pm$0.9 & 57.3$\pm$0.3 & 50.3$\pm$1.2 & 32.6$\pm$0.0 & 22.4$\pm$0.1 \\
      GrayCIFAR10 & 68.2 & 71.0$\pm$0.0 & 55.4 & 66.7$\pm$0.0 & 66.9$\pm$0.0 & 32.4$\pm$0.6 & 46.2$\pm$0.2 & 42.5$\pm$0.7 & 10.2$\pm$0.0 & 22.1$\pm$0.2 \\
      UniformNoise & 82.0 & 83.7$\pm$0.1 & 67.4 & 94.6$\pm$0.1 & 75.7$\pm$0.4 & 29.4$\pm$0.7 & 36.0$\pm$0.9 & 39.6$\pm$1.2 & 10.1$\pm$0.0 & 21.7$\pm$0.7 \\
      Noise & 99.4 & 99.7$\pm$0.0 & 99.5 & 99.9$\pm$0.0 & 99.4$\pm$0.0 & 85.6$\pm$1.5 & 96.4$\pm$0.2 & 95.9$\pm$0.6 & 10.4$\pm$0.0 & 14.2$\pm$0.1 \\

      \midrule

      \textbf{SVHN} & 98.5 & 97.1$\pm$0.0 & 98.1 & 97.5$\pm$0.0 & 98.5$\pm$0.0 & 97.5$\pm$0.0 & 91.8$\pm$0.5 & 95.9$\pm$0.1 & 98.4$\pm$0.0 & 98.4$\pm$0.0 \\
      CIFAR10 & 72.5 & 62.4$\pm$0.0 & 58.7 & 58.1$\pm$0.0 & 71.8$\pm$0.0 & 60.8$\pm$0.1 & 48.5$\pm$0.2 & 52.3$\pm$0.4 & 10.7$\pm$0.0 & 13.3$\pm$0.2 \\
      LSUN & 73.7 & 63.9$\pm$0.0 & 59.0 & 59.6$\pm$0.0 & 73.0$\pm$0.0 & 61.5$\pm$0.2 & 48.2$\pm$0.3 & 52.5$\pm$0.5 & 10.3$\pm$0.0 & 12.8$\pm$0.3 \\
      CIFAR100 & 73.4 & 63.5$\pm$0.0 & 60.0 & 59.6$\pm$0.0 & 72.7$\pm$0.0 & 61.6$\pm$0.1 & 48.9$\pm$0.2 & 52.9$\pm$0.4 & 11.3$\pm$0.0 & 14.0$\pm$0.3 \\
      FMNIST3D & 74.6 & 64.8$\pm$0.0 & 64.1 & 61.4$\pm$0.0 & 74.0$\pm$0.0 & 65.2$\pm$0.2 & 53.3$\pm$0.4 & 57.6$\pm$0.4 & 10.6$\pm$0.0 & 13.7$\pm$0.2 \\
      UniformNoise & 79.1 & 70.8$\pm$0.1 & 54.6 & 63.8$\pm$0.2 & 77.8$\pm$0.2 & 62.5$\pm$0.6 & 43.9$\pm$0.5 & 51.6$\pm$0.3 & 10.0$\pm$0.0 & 12.4$\pm$0.3 \\
      Noise & 64.2 & 55.1$\pm$0.2 & 53.3 & 51.7$\pm$0.2 & 63.5$\pm$0.2 & 53.6$\pm$0.1 & 41.3$\pm$0.2 & 45.8$\pm$0.4 & 55.3$\pm$0.1 & 54.3$\pm$0.1 \\

      \midrule

      \textbf{CIFAR10} & 97.2 & 94.8$\pm$0.0 & 96.1 & 95.7$\pm$0.0 & 96.9$\pm$0.0 & 96.2$\pm$0.0 & 90.6$\pm$0.0 & 83.4$\pm$0.2 & 97.3$\pm$0.0 & 97.0$\pm$0.0 \\
      SVHN & 70.6 & 57.2$\pm$0.0 & 57.2 & 52.6$\pm$0.0 & 67.7$\pm$0.1 & 63.2$\pm$0.3 & 42.1$\pm$0.5 & 35.0$\pm$0.5 & 56.1$\pm$0.0 & 53.3$\pm$0.1 \\
      LSUN & 74.8 & 61.5$\pm$0.0 & 65.6 & 61.8$\pm$0.0 & 73.4$\pm$0.0 & 68.7$\pm$0.2 & 51.3$\pm$0.3 & 40.5$\pm$0.4 & 66.2$\pm$0.0 & 64.4$\pm$0.1 \\
      CIFAR100 & 78.7 & 67.1$\pm$0.0 & 71.2 & 68.3$\pm$0.0 & 77.3$\pm$0.0 & 73.4$\pm$0.1 & 56.6$\pm$0.1 & 46.6$\pm$0.2 & 78.1$\pm$0.0 & 76.6$\pm$0.0 \\
      FMNIST3D & 68.8 & 53.7$\pm$0.0 & 60.7 & 54.7$\pm$0.0 & 66.5$\pm$0.1 & 61.2$\pm$0.1 & 40.4$\pm$0.4 & 32.9$\pm$0.3 & 61.4$\pm$0.0 & 59.2$\pm$0.0 \\
      UniformNoise & 88.0 & 71.5$\pm$0.1 & 89.3 & 82.2$\pm$0.0 & 79.5$\pm$0.6 & 62.6$\pm$1.6 & 30.7$\pm$0.6 & 25.2$\pm$0.2 & 10.1$\pm$0.0 & 12.2$\pm$0.1 \\
      Noise & 64.5 & 52.2$\pm$0.2 & 53.7 & 52.6$\pm$0.1 & 59.6$\pm$0.2 & 53.8$\pm$0.3 & 35.5$\pm$0.4 & 30.3$\pm$0.4 & 48.6$\pm$0.3 & 46.4$\pm$0.2 \\

      \midrule

      \textbf{CIFAR100} & 85.7 & 76.8$\pm$0.0 & 81.5 & 80.6$\pm$0.0 & 80.4$\pm$0.0 & 72.6$\pm$0.1 & 75.7$\pm$0.1 & 63.8$\pm$0.2 & 86.5$\pm$0.0 & 81.2$\pm$0.1 \\
      SVHN & 61.3 & 42.0$\pm$0.0 & 47.5 & 42.2$\pm$0.0 & 52.9$\pm$0.1 & 40.7$\pm$0.5 & 46.8$\pm$0.6 & 33.0$\pm$0.8 & 63.7$\pm$0.0 & 54.6$\pm$0.2 \\
      LSUN & 64.9 & 47.8$\pm$0.0 & 51.7 & 49.3$\pm$0.0 & 56.0$\pm$0.2 & 46.1$\pm$0.1 & 49.1$\pm$0.4 & 37.5$\pm$0.8 & 58.4$\pm$0.0 & 50.8$\pm$0.3 \\
      CIFAR10 & 67.2 & 51.8$\pm$0.0 & 56.1 & 54.4$\pm$0.0 & 58.9$\pm$0.1 & 49.8$\pm$0.1 & 52.6$\pm$0.1 & 41.6$\pm$0.2 & 68.8$\pm$0.0 & 59.7$\pm$0.1 \\
      FMNIST3D & 56.4 & 35.7$\pm$0.0 & 45.8 & 39.2$\pm$0.0 & 49.0$\pm$0.1 & 40.1$\pm$0.3 & 42.7$\pm$0.2 & 32.6$\pm$0.3 & 53.9$\pm$0.0 & 46.2$\pm$0.2 \\
      UniformNoise & 68.3 & 56.5$\pm$0.1 & 29.5 & 43.7$\pm$0.1 & 45.3$\pm$0.5 & 33.0$\pm$0.8 & 36.5$\pm$0.9 & 24.7$\pm$0.8 & 1.7$\pm$0.0 & 1.7$\pm$0.0 \\
      Noise & 68.7 & 55.3$\pm$0.2 & 50.5 & 50.2$\pm$0.3 & 58.1$\pm$0.2 & 36.3$\pm$1.0 & 51.2$\pm$1.0 & 29.0$\pm$1.4 & 64.5$\pm$0.2 & 53.8$\pm$0.4 \\

      \bottomrule
    \end{tabular}
  }
\end{table*}

\begin{table*}[tb]
  \caption{Detailed FPR95 results. Values are averages over five prediction runs.}
  \label{tab:fpr95}

  \centering
  \scriptsize
  \setlength{\tabcolsep}{7pt}

  \resizebox{\textwidth}{!}{
    \begin{tabular}{lccccc >{\columncolor{tablegray}}c c >{\columncolor{tablegray}}c c >{\columncolor{tablegray}}c}

      \toprule

      \textbf{Dataset} & \textbf{MAP} & \textbf{MAP-Temp} & \textbf{DE} & \textbf{DE-Temp} & \textbf{LA} & \textbf{LA-LULA} & \textbf{LLLA} & \textbf{LLLA-LULA} & \textbf{OE} & \textbf{OE-LULA}  \\

      \midrule

      \textbf{MNIST} & - & - & - & - & - & - & - & - & - & - \\
      EMNIST & 23.9 & 24.0$\pm$0.0 & 22.3 & 22.4$\pm$0.0 & 23.9$\pm$0.0 & 23.6$\pm$0.2 & 24.0$\pm$0.2 & 23.5$\pm$0.1 & 27.5$\pm$0.0 & 23.5$\pm$0.6 \\
      KMNIST & 2.4 & 2.4$\pm$0.0 & 1.8 & 2.3$\pm$0.0 & 2.4$\pm$0.0 & 0.8$\pm$0.0 & 1.8$\pm$0.2 & 1.0$\pm$0.1 & 5.1$\pm$0.0 & 3.6$\pm$0.4 \\
      FMNIST & 2.4 & 2.4$\pm$0.0 & 1.1 & 1.8$\pm$0.0 & 2.3$\pm$0.0 & 0.8$\pm$0.0 & 1.5$\pm$0.1 & 0.9$\pm$0.1 & 0.2$\pm$0.0 & 1.8$\pm$0.2 \\
      GrayCIFAR10 & 0.1 & 0.0$\pm$0.0 & 0.0 & 0.0$\pm$0.0 & 0.0$\pm$0.0 & 0.0$\pm$0.0 & 0.0$\pm$0.0 & 0.0$\pm$0.0 & 0.0$\pm$0.0 & 0.8$\pm$0.2 \\
      UniformNoise & 1.1 & 1.0$\pm$0.0 & 0.0 & 0.2$\pm$0.0 & 0.3$\pm$0.0 & 0.0$\pm$0.0 & 0.0$\pm$0.0 & 0.0$\pm$0.0 & 0.0$\pm$0.0 & 0.9$\pm$0.5 \\
      Noise & 0.1 & 0.1$\pm$0.0 & 0.2 & 0.2$\pm$0.1 & 0.1$\pm$0.0 & 9.4$\pm$3.0 & 7.6$\pm$2.2 & 1.4$\pm$0.3 & 0.0$\pm$0.0 & 0.2$\pm$0.2 \\

      \midrule

      \textbf{SVHN} & - & - & - & - & - & - & - & - & - & - \\
      CIFAR10 & 24.0 & 23.2$\pm$0.0 & 11.3 & 14.1$\pm$0.0 & 23.8$\pm$0.1 & 20.7$\pm$0.2 & 23.7$\pm$1.8 & 19.6$\pm$0.4 & 0.0$\pm$0.0 & 0.0$\pm$0.0 \\
      LSUN & 25.7 & 25.3$\pm$0.0 & 11.0 & 16.3$\pm$0.0 & 25.5$\pm$0.2 & 21.3$\pm$0.5 & 22.2$\pm$2.2 & 19.7$\pm$0.8 & 0.0$\pm$0.0 & 0.0$\pm$0.0 \\
      CIFAR100 & 25.5 & 24.8$\pm$0.0 & 13.3 & 16.8$\pm$0.0 & 25.3$\pm$0.1 & 21.9$\pm$0.2 & 24.3$\pm$1.6 & 20.5$\pm$0.4 & 0.2$\pm$0.0 & 0.1$\pm$0.0 \\
      FMNIST3D & 29.7 & 28.9$\pm$0.0 & 22.5 & 22.5$\pm$0.0 & 29.8$\pm$0.1 & 29.4$\pm$0.3 & 33.0$\pm$1.6 & 29.2$\pm$0.3 & 0.0$\pm$0.0 & 0.0$\pm$0.0 \\
      UniformNoise & 33.2 & 34.1$\pm$0.3 & 5.4 & 18.4$\pm$0.3 & 31.7$\pm$0.3 & 19.5$\pm$1.1 & 14.0$\pm$2.1 & 15.6$\pm$0.6 & 0.0$\pm$0.0 & 0.0$\pm$0.0 \\
      Noise & 17.5 & 17.0$\pm$0.5 & 7.8 & 10.0$\pm$0.5 & 17.1$\pm$0.5 & 13.6$\pm$0.3 & 14.7$\pm$1.7 & 12.1$\pm$0.7 & 10.3$\pm$0.1 & 10.2$\pm$0.2 \\

      \midrule

      \textbf{CIFAR10} & - & - & - & - & - & - & - & - & - & - \\
      SVHN & 41.7 & 35.4$\pm$0.0 & 25.0 & 20.1$\pm$0.0 & 38.9$\pm$0.2 & 37.6$\pm$0.5 & 19.6$\pm$0.8 & 20.4$\pm$1.3 & 22.8$\pm$0.0 & 20.9$\pm$0.1 \\
      LSUN & 50.7 & 45.7$\pm$0.0 & 45.3 & 39.3$\pm$0.0 & 50.9$\pm$0.2 & 48.9$\pm$0.5 & 41.9$\pm$0.2 & 37.4$\pm$1.1 & 38.3$\pm$0.0 & 38.4$\pm$0.2 \\
      CIFAR100 & 60.1 & 55.9$\pm$0.0 & 54.6 & 51.7$\pm$0.0 & 59.7$\pm$0.3 & 58.7$\pm$0.2 & 51.4$\pm$0.4 & 50.4$\pm$0.4 & 58.0$\pm$0.0 & 57.5$\pm$0.2 \\
      FMNIST3D & 40.4 & 31.3$\pm$0.0 & 35.2 & 27.0$\pm$0.0 & 39.0$\pm$0.3 & 36.3$\pm$0.3 & 19.1$\pm$0.6 & 17.4$\pm$0.4 & 30.0$\pm$0.0 & 29.2$\pm$0.2 \\
      UniformNoise & 89.0 & 81.6$\pm$0.5 & 99.9 & 99.3$\pm$0.1 & 73.9$\pm$1.7 & 31.3$\pm$5.9 & 0.1$\pm$0.1 & 0.7$\pm$0.3 & 0.0$\pm$0.0 & 0.0$\pm$0.0 \\
      Noise & 36.6 & 31.8$\pm$0.5 & 25.7 & 31.6$\pm$0.2 & 28.9$\pm$0.4 & 24.3$\pm$0.7 & 9.9$\pm$0.7 & 11.0$\pm$1.0 & 15.4$\pm$0.2 & 14.0$\pm$0.3 \\

      \midrule

      \textbf{CIFAR100} & - & - & - & - & - & - & - & - & - & - \\
      SVHN & 73.8 & 67.9$\pm$0.0 & 62.1 & 58.2$\pm$0.0 & 73.3$\pm$0.3 & 68.8$\pm$0.6 & 72.4$\pm$0.9 & 67.5$\pm$0.9 & 75.9$\pm$0.0 & 74.1$\pm$0.3 \\
      LSUN & 81.7 & 81.7$\pm$0.0 & 73.0 & 75.3$\pm$0.0 & 82.4$\pm$0.6 & 82.1$\pm$0.4 & 81.7$\pm$0.6 & 81.0$\pm$0.8 & 69.7$\pm$0.0 & 71.0$\pm$0.8 \\
      CIFAR10 & 83.0 & 81.5$\pm$0.0 & 77.2 & 78.2$\pm$0.0 & 82.9$\pm$0.2 & 82.8$\pm$0.3 & 82.3$\pm$0.2 & 82.7$\pm$0.2 & 82.4$\pm$0.0 & 81.5$\pm$0.2 \\
      FMNIST3D & 70.2 & 59.5$\pm$0.0 & 64.3 & 58.8$\pm$0.0 & 70.6$\pm$0.2 & 69.1$\pm$0.6 & 69.1$\pm$0.3 & 67.8$\pm$0.9 & 63.1$\pm$0.0 & 62.9$\pm$0.4 \\
      UniformNoise & 97.7 & 100.0$\pm$0.0 & 15.7 & 99.5$\pm$0.1 & 89.1$\pm$1.1 & 71.2$\pm$4.3 & 76.9$\pm$3.1 & 57.5$\pm$5.6 & 0.0$\pm$0.0 & 0.0$\pm$0.0 \\
      Noise & 74.2 & 72.0$\pm$0.5 & 63.3 & 63.8$\pm$0.4 & 71.6$\pm$0.5 & 57.1$\pm$1.8 & 69.9$\pm$0.5 & 54.3$\pm$1.6 & 66.6$\pm$0.3 & 61.6$\pm$0.8 \\

      \bottomrule
    \end{tabular}
  }
\end{table*}

\begin{table*}[tb]
  \caption{Detailed AUROC results. Values are averages over five prediction runs.}
  \label{tab:auroc}

  \centering
  \scriptsize
  \setlength{\tabcolsep}{7pt}

  \resizebox{\textwidth}{!}{
    \begin{tabular}{lccccc >{\columncolor{tablegray}}c c >{\columncolor{tablegray}}c c >{\columncolor{tablegray}}c}

      \toprule

      \textbf{Dataset} & \textbf{MAP} & \textbf{MAP-Temp} & \textbf{DE} & \textbf{DE-Temp} & \textbf{LA} & \textbf{LA-LULA} & \textbf{LLLA} & \textbf{LLLA-LULA} & \textbf{OE} & \textbf{OE-LULA}  \\

      \midrule

      \textbf{MNIST} & - & - & - & - & - & - & - & - & - & - \\
      EMNIST & 89.5 & 89.5$\pm$0.0 & 89.8 & 89.6$\pm$0.0 & 89.5$\pm$0.0 & 90.6$\pm$0.2 & 89.6$\pm$0.1 & 90.3$\pm$0.0 & 92.9$\pm$0.0 & 93.2$\pm$0.1 \\
      KMNIST & 98.9 & 98.9$\pm$0.0 & 99.1 & 98.9$\pm$0.0 & 98.9$\pm$0.0 & 99.5$\pm$0.0 & 99.3$\pm$0.0 & 99.5$\pm$0.0 & 98.6$\pm$0.0 & 98.9$\pm$0.1 \\
      FMNIST & 98.8 & 98.8$\pm$0.0 & 99.2 & 99.0$\pm$0.0 & 98.9$\pm$0.0 & 99.3$\pm$0.0 & 99.3$\pm$0.0 & 99.3$\pm$0.1 & 99.7$\pm$0.0 & 99.2$\pm$0.0 \\
      GrayCIFAR10 & 99.7 & 99.6$\pm$0.0 & 99.8 & 99.8$\pm$0.0 & 99.7$\pm$0.0 & 99.6$\pm$0.0 & 99.8$\pm$0.0 & 99.7$\pm$0.0 & 100.0$\pm$0.0 & 99.3$\pm$0.0 \\
      UniformNoise & 99.1 & 99.2$\pm$0.0 & 99.8 & 99.1$\pm$0.0 & 99.5$\pm$0.0 & 99.8$\pm$0.0 & 100.0$\pm$0.0 & 99.8$\pm$0.0 & 100.0$\pm$0.0 & 99.4$\pm$0.1 \\
      Noise & 97.4 & 97.3$\pm$0.0 & 96.9 & 96.8$\pm$0.0 & 97.4$\pm$0.0 & 96.3$\pm$0.3 & 96.7$\pm$0.1 & 96.7$\pm$0.1 & 100.0$\pm$0.0 & 99.9$\pm$0.0 \\

      \midrule

      \textbf{SVHN} & - & - & - & - & - & - & - & - & - & - \\
      CIFAR10 & 95.2 & 95.3$\pm$0.0 & 97.7 & 97.2$\pm$0.0 & 95.3$\pm$0.0 & 96.2$\pm$0.1 & 95.5$\pm$0.3 & 96.5$\pm$0.0 & 100.0$\pm$0.0 & 100.0$\pm$0.0 \\
      LSUN & 94.9 & 94.9$\pm$0.0 & 97.9 & 96.9$\pm$0.0 & 94.9$\pm$0.0 & 96.0$\pm$0.1 & 95.8$\pm$0.3 & 96.5$\pm$0.1 & 100.0$\pm$0.0 & 100.0$\pm$0.0 \\
      CIFAR100 & 94.6 & 94.6$\pm$0.0 & 97.2 & 96.5$\pm$0.0 & 94.7$\pm$0.0 & 95.8$\pm$0.0 & 95.3$\pm$0.3 & 96.2$\pm$0.0 & 100.0$\pm$0.0 & 100.0$\pm$0.0 \\
      FMNIST3D & 94.2 & 94.3$\pm$0.0 & 96.2 & 96.0$\pm$0.0 & 94.2$\pm$0.0 & 94.4$\pm$0.1 & 93.0$\pm$0.5 & 94.3$\pm$0.1 & 100.0$\pm$0.0 & 100.0$\pm$0.0 \\
      UniformNoise & 93.8 & 93.4$\pm$0.1 & 98.5 & 96.5$\pm$0.0 & 94.1$\pm$0.1 & 96.6$\pm$0.2 & 97.4$\pm$0.2 & 97.3$\pm$0.1 & 100.0$\pm$0.0 & 100.0$\pm$0.0 \\
      Noise & 96.6 & 96.6$\pm$0.1 & 98.3 & 97.9$\pm$0.1 & 96.6$\pm$0.1 & 97.4$\pm$0.0 & 97.2$\pm$0.3 & 97.7$\pm$0.1 & 97.9$\pm$0.1 & 97.9$\pm$0.1 \\

      \midrule

      \textbf{CIFAR10} & - & - & - & - & - & - & - & - & - & - \\
      SVHN & 94.6 & 95.3$\pm$0.0 & 96.6 & 97.1$\pm$0.0 & 94.9$\pm$0.0 & 95.0$\pm$0.1 & 96.9$\pm$0.1 & 96.6$\pm$0.2 & 97.0$\pm$0.0 & 97.2$\pm$0.0 \\
      LSUN & 92.5 & 93.5$\pm$0.0 & 93.7 & 94.3$\pm$0.0 & 92.5$\pm$0.0 & 92.8$\pm$0.1 & 93.2$\pm$0.1 & 93.9$\pm$0.2 & 94.9$\pm$0.0 & 94.9$\pm$0.0 \\
      CIFAR100 & 90.0 & 90.6$\pm$0.0 & 91.1 & 91.6$\pm$0.0 & 90.1$\pm$0.0 & 90.1$\pm$0.0 & 90.2$\pm$0.1 & 90.0$\pm$0.1 & 90.1$\pm$0.0 & 90.2$\pm$0.0 \\
      FMNIST3D & 94.7 & 95.8$\pm$0.0 & 95.3 & 96.3$\pm$0.0 & 94.9$\pm$0.0 & 95.3$\pm$0.0 & 97.0$\pm$0.1 & 97.2$\pm$0.1 & 95.9$\pm$0.0 & 96.0$\pm$0.0 \\
      UniformNoise & 91.5 & 92.6$\pm$0.0 & 88.6 & 91.0$\pm$0.0 & 93.6$\pm$0.1 & 96.2$\pm$0.3 & 99.4$\pm$0.1 & 99.3$\pm$0.0 & 100.0$\pm$0.0 & 100.0$\pm$0.0 \\
      Noise & 95.2 & 95.7$\pm$0.1 & 96.6 & 95.9$\pm$0.1 & 96.0$\pm$0.1 & 96.7$\pm$0.1 & 98.1$\pm$0.1 & 98.0$\pm$0.1 & 97.1$\pm$0.1 & 97.4$\pm$0.1 \\

      \midrule

      \textbf{CIFAR100} & - & - & - & - & - & - & - & - & - & - \\
      SVHN & 80.2 & 83.9$\pm$0.0 & 85.0 & 86.7$\pm$0.0 & 80.5$\pm$0.1 & 83.5$\pm$0.4 & 80.7$\pm$0.4 & 84.1$\pm$0.7 & 80.1$\pm$0.0 & 80.2$\pm$0.2 \\
      LSUN & 78.1 & 80.1$\pm$0.0 & 82.5 & 82.7$\pm$0.0 & 78.5$\pm$0.2 & 79.1$\pm$0.1 & 79.4$\pm$0.4 & 79.8$\pm$0.9 & 83.7$\pm$0.0 & 83.2$\pm$0.2 \\
      CIFAR10 & 75.4 & 76.4$\pm$0.0 & 78.7 & 78.6$\pm$0.0 & 75.5$\pm$0.1 & 75.4$\pm$0.2 & 75.8$\pm$0.1 & 75.3$\pm$0.2 & 75.4$\pm$0.0 & 75.8$\pm$0.0 \\
      FMNIST3D & 84.1 & 88.2$\pm$0.0 & 86.6 & 89.1$\pm$0.0 & 83.7$\pm$0.1 & 84.1$\pm$0.2 & 84.3$\pm$0.2 & 84.5$\pm$0.3 & 86.4$\pm$0.0 & 86.2$\pm$0.1 \\
      UniformNoise & 78.7 & 75.5$\pm$0.1 & 96.8 & 88.5$\pm$0.0 & 88.1$\pm$0.4 & 91.2$\pm$0.6 & 90.4$\pm$0.6 & 93.0$\pm$0.6 & 100.0$\pm$0.0 & 100.0$\pm$0.0 \\
      Noise & 69.3 & 71.5$\pm$0.2 & 80.9 & 78.1$\pm$0.2 & 74.3$\pm$0.3 & 86.2$\pm$0.9 & 75.5$\pm$0.9 & 87.0$\pm$1.4 & 75.1$\pm$0.2 & 78.5$\pm$0.4 \\

      \bottomrule
    \end{tabular}
  }
\end{table*}

\begin{table*}[tb]
  \caption{Detailed AUPRC results. Values are averages over five prediction runs.}
  \label{tab:auprc}

  \centering
  \scriptsize
  \setlength{\tabcolsep}{7pt}

  \resizebox{\textwidth}{!}{
    \begin{tabular}{lccccc >{\columncolor{tablegray}}c c >{\columncolor{tablegray}}c c >{\columncolor{tablegray}}c}

      \toprule

      \textbf{Dataset} & \textbf{MAP} & \textbf{MAP-Temp} & \textbf{DE} & \textbf{DE-Temp} & \textbf{LA} & \textbf{LA-LULA} & \textbf{LLLA} & \textbf{LLLA-LULA} & \textbf{OE} & \textbf{OE-LULA}  \\

      \midrule

      \textbf{MNIST} & - & - & - & - & - & - & - & - & - & - \\
      EMNIST & 67.3 & 67.2$\pm$0.0 & 67.0 & 66.5$\pm$0.0 & 67.3$\pm$0.2 & 69.7$\pm$0.7 & 67.9$\pm$0.4 & 69.1$\pm$0.3 & 84.5$\pm$0.0 & 81.0$\pm$0.3 \\
      KMNIST & 97.9 & 97.9$\pm$0.0 & 98.4 & 98.0$\pm$0.0 & 98.0$\pm$0.0 & 99.5$\pm$0.0 & 99.2$\pm$0.1 & 99.4$\pm$0.0 & 98.6$\pm$0.0 & 99.0$\pm$0.0 \\
      FMNIST & 98.3 & 98.4$\pm$0.0 & 98.9 & 98.5$\pm$0.0 & 98.4$\pm$0.0 & 99.3$\pm$0.0 & 99.1$\pm$0.0 & 99.3$\pm$0.1 & 99.7$\pm$0.0 & 99.3$\pm$0.0 \\
      GrayCIFAR10 & 99.7 & 99.7$\pm$0.0 & 99.9 & 99.8$\pm$0.0 & 99.7$\pm$0.0 & 99.7$\pm$0.0 & 99.8$\pm$0.0 & 99.7$\pm$0.0 & 100.0$\pm$0.0 & 99.4$\pm$0.0 \\
      UniformNoise & 99.8 & 99.8$\pm$0.0 & 100.0 & 99.8$\pm$0.0 & 99.9$\pm$0.0 & 100.0$\pm$0.0 & 100.0$\pm$0.0 & 100.0$\pm$0.0 & 100.0$\pm$0.0 & 99.9$\pm$0.0 \\
      Noise & 99.5 & 99.4$\pm$0.0 & 99.4 & 99.3$\pm$0.0 & 99.5$\pm$0.0 & 99.2$\pm$0.1 & 99.3$\pm$0.0 & 99.3$\pm$0.0 & 100.0$\pm$0.0 & 100.0$\pm$0.0 \\

      \midrule

      \textbf{SVHN} & - & - & - & - & - & - & - & - & - & - \\
      CIFAR10 & 97.8 & 97.8$\pm$0.0 & 99.1 & 98.8$\pm$0.0 & 97.8$\pm$0.0 & 98.3$\pm$0.0 & 98.1$\pm$0.2 & 98.5$\pm$0.0 & 100.0$\pm$0.0 & 100.0$\pm$0.0 \\
      LSUN & 99.9 & 99.9$\pm$0.0 & 100.0 & 100.0$\pm$0.0 & 99.9$\pm$0.0 & 99.9$\pm$0.0 & 99.9$\pm$0.0 & 100.0$\pm$0.0 & 100.0$\pm$0.0 & 100.0$\pm$0.0 \\
      CIFAR100 & 97.4 & 97.3$\pm$0.0 & 98.8 & 98.4$\pm$0.0 & 97.4$\pm$0.0 & 98.0$\pm$0.0 & 97.9$\pm$0.2 & 98.3$\pm$0.0 & 100.0$\pm$0.0 & 100.0$\pm$0.0 \\
      FMNIST3D & 97.4 & 97.4$\pm$0.0 & 98.5 & 98.3$\pm$0.0 & 97.3$\pm$0.0 & 97.5$\pm$0.0 & 96.8$\pm$0.3 & 97.4$\pm$0.0 & 100.0$\pm$0.0 & 100.0$\pm$0.0 \\
      UniformNoise & 99.4 & 99.3$\pm$0.0 & 99.9 & 99.7$\pm$0.0 & 99.4$\pm$0.0 & 99.7$\pm$0.0 & 99.8$\pm$0.0 & 99.8$\pm$0.0 & 100.0$\pm$0.0 & 100.0$\pm$0.0 \\
      Noise & 99.7 & 99.7$\pm$0.0 & 99.8 & 99.8$\pm$0.0 & 99.7$\pm$0.0 & 99.8$\pm$0.0 & 99.8$\pm$0.0 & 99.8$\pm$0.0 & 99.8$\pm$0.0 & 99.8$\pm$0.0 \\

      \midrule

      \textbf{CIFAR10} & - & - & - & - & - & - & - & - & - & - \\
      SVHN & 91.5 & 92.2$\pm$0.0 & 94.5 & 95.0$\pm$0.0 & 91.9$\pm$0.0 & 92.0$\pm$0.1 & 94.3$\pm$0.2 & 93.9$\pm$0.3 & 94.3$\pm$0.0 & 94.6$\pm$0.0 \\
      LSUN & 99.7 & 99.7$\pm$0.0 & 99.7 & 99.8$\pm$0.0 & 99.7$\pm$0.0 & 99.7$\pm$0.0 & 99.7$\pm$0.0 & 99.7$\pm$0.0 & 99.8$\pm$0.0 & 99.8$\pm$0.0 \\
      CIFAR100 & 90.3 & 90.6$\pm$0.0 & 91.2 & 91.6$\pm$0.0 & 90.3$\pm$0.0 & 90.3$\pm$0.0 & 89.6$\pm$0.1 & 89.1$\pm$0.1 & 90.3$\pm$0.0 & 90.3$\pm$0.0 \\
      FMNIST3D & 95.3 & 96.1$\pm$0.0 & 95.7 & 96.6$\pm$0.0 & 95.5$\pm$0.0 & 95.7$\pm$0.0 & 97.0$\pm$0.1 & 97.2$\pm$0.1 & 96.1$\pm$0.0 & 96.2$\pm$0.0 \\
      UniformNoise & 98.1 & 98.4$\pm$0.0 & 97.5 & 98.1$\pm$0.0 & 98.6$\pm$0.0 & 99.2$\pm$0.1 & 99.9$\pm$0.0 & 99.8$\pm$0.0 & 100.0$\pm$0.0 & 100.0$\pm$0.0 \\
      Noise & 98.8 & 98.9$\pm$0.0 & 99.1 & 99.0$\pm$0.0 & 99.0$\pm$0.0 & 99.2$\pm$0.0 & 99.5$\pm$0.0 & 99.5$\pm$0.0 & 99.2$\pm$0.0 & 99.3$\pm$0.0 \\

      \midrule

      \textbf{CIFAR100} & - & - & - & - & - & - & - & - & - & - \\
      SVHN & 67.8 & 72.3$\pm$0.0 & 73.1 & 75.4$\pm$0.0 & 67.4$\pm$0.2 & 71.7$\pm$1.0 & 66.5$\pm$0.9 & 71.9$\pm$1.6 & 69.4$\pm$0.0 & 68.3$\pm$0.3 \\
      LSUN & 99.0 & 99.1$\pm$0.0 & 99.2 & 99.2$\pm$0.0 & 99.0$\pm$0.0 & 99.0$\pm$0.0 & 99.0$\pm$0.0 & 99.0$\pm$0.1 & 99.3$\pm$0.0 & 99.2$\pm$0.0 \\
      CIFAR10 & 74.7 & 75.2$\pm$0.0 & 77.8 & 77.6$\pm$0.0 & 74.4$\pm$0.1 & 73.8$\pm$0.2 & 74.6$\pm$0.3 & 73.1$\pm$0.4 & 75.3$\pm$0.0 & 75.3$\pm$0.1 \\
      FMNIST3D & 85.0 & 88.5$\pm$0.0 & 87.5 & 89.6$\pm$0.0 & 84.3$\pm$0.1 & 84.2$\pm$0.2 & 84.5$\pm$0.2 & 84.3$\pm$0.3 & 87.1$\pm$0.0 & 86.6$\pm$0.1 \\
      UniformNoise & 94.7 & 94.0$\pm$0.0 & 99.3 & 97.4$\pm$0.0 & 97.2$\pm$0.1 & 98.0$\pm$0.2 & 97.8$\pm$0.2 & 98.4$\pm$0.1 & 100.0$\pm$0.0 & 100.0$\pm$0.0 \\
      Noise & 90.2 & 91.0$\pm$0.1 & 94.2 & 92.8$\pm$0.1 & 92.2$\pm$0.1 & 96.2$\pm$0.3 & 92.5$\pm$0.4 & 96.4$\pm$0.5 & 92.1$\pm$0.1 & 93.3$\pm$0.2 \\

      \bottomrule
    \end{tabular}
  }
\end{table*}

%% file: figs/cifar10c_ece.tex

\begin{tikzpicture}[baseline, trim axis left]

\definecolor{color0}{rgb}{0.0906862745098039,0.425980392156863,0.611274509803922}
\definecolor{color1}{rgb}{0.758823529411765,0.511764705882353,0.711764705882353}
\definecolor{color2}{rgb}{0.834803921568628,0.802450980392157,0.290686274509804}
\definecolor{color3}{rgb}{0.084313725490196,0.543137254901961,0.416666666666667}

\tikzstyle{every node}=[font=\scriptsize]

\begin{axis}[
width=\figcifarcappendixwidth,
height=\figcifarcappendixheight,
legend cell align={left},
legend style={fill opacity=0.8, draw opacity=1, text opacity=1, at={(0.97,0.03)}, anchor=south east, draw=white!80!black},
tick align=inside,
tick pos=left,
xmajorticks=false,
x grid style={white!69.0196078431373!black},
xmin=-0.5, xmax=4.5,
xtick style={draw=none},
xtick={0,1,2,3,4},
xticklabels={1,2,3,4,5},
y grid style={white!69.0196078431373!black},
ylabel={ECE $\downarrow$},
ymin=10.4959216087877, ymax=63.8145184430819,
ytick style={draw=none},
ymajorgrids
]
\path [draw=white!18.8235294117647!black, fill=color0, semithick]
(axis cs:-0.24875,26.154682731973)
--(axis cs:-0.12625,26.154682731973)
--(axis cs:-0.12625,37.2735804015945)
--(axis cs:-0.24875,37.2735804015945)
--(axis cs:-0.24875,26.154682731973)
--cycle;
\path [draw=white!18.8235294117647!black, fill=color1, semithick]
(axis cs:-0.12375,16.2966509090997)
--(axis cs:-0.00125,16.2966509090997)
--(axis cs:-0.00125,26.8210790337983)
--(axis cs:-0.12375,26.8210790337983)
--(axis cs:-0.12375,16.2966509090997)
--cycle;
\path [draw=white!18.8235294117647!black, fill=color2, semithick]
(axis cs:0.00125,24.7413226977366)
--(axis cs:0.12375,24.7413226977366)
--(axis cs:0.12375,35.490949961668)
--(axis cs:0.00125,35.490949961668)
--(axis cs:0.00125,24.7413226977366)
--cycle;
\path [draw=white!18.8235294117647!black, fill=color3, semithick]
(axis cs:0.12625,19.7176999657374)
--(axis cs:0.24875,19.7176999657374)
--(axis cs:0.24875,30.8586725278069)
--(axis cs:0.12625,30.8586725278069)
--(axis cs:0.12625,19.7176999657374)
--cycle;
\path [draw=white!18.8235294117647!black, fill=color0, semithick]
(axis cs:0.75125,32.3081777556577)
--(axis cs:0.87375,32.3081777556577)
--(axis cs:0.87375,43.4691378619113)
--(axis cs:0.75125,43.4691378619113)
--(axis cs:0.75125,32.3081777556577)
--cycle;
\path [draw=white!18.8235294117647!black, fill=color1, semithick]
(axis cs:0.87625,22.302484947421)
--(axis cs:0.99875,22.302484947421)
--(axis cs:0.99875,32.2989396022749)
--(axis cs:0.87625,32.2989396022749)
--(axis cs:0.87625,22.302484947421)
--cycle;
\path [draw=white!18.8235294117647!black, fill=color2, semithick]
(axis cs:1.00125,30.8184647841)
--(axis cs:1.12375,30.8184647841)
--(axis cs:1.12375,41.5185859580157)
--(axis cs:1.00125,41.5185859580157)
--(axis cs:1.00125,30.8184647841)
--cycle;
\path [draw=white!18.8235294117647!black, fill=color3, semithick]
(axis cs:1.12625,25.8663019548549)
--(axis cs:1.24875,25.8663019548549)
--(axis cs:1.24875,36.2028864426037)
--(axis cs:1.12625,36.2028864426037)
--(axis cs:1.12625,25.8663019548549)
--cycle;
\path [draw=white!18.8235294117647!black, fill=color0, semithick]
(axis cs:1.75125,38.929828149668)
--(axis cs:1.87375,38.929828149668)
--(axis cs:1.87375,48.1170145298489)
--(axis cs:1.75125,48.1170145298489)
--(axis cs:1.75125,38.929828149668)
--cycle;
\path [draw=white!18.8235294117647!black, fill=color1, semithick]
(axis cs:1.87625,28.2130998686219)
--(axis cs:1.99875,28.2130998686219)
--(axis cs:1.99875,35.7832921793715)
--(axis cs:1.87625,35.7832921793715)
--(axis cs:1.87625,28.2130998686219)
--cycle;
\path [draw=white!18.8235294117647!black, fill=color2, semithick]
(axis cs:2.00125,37.3707809383596)
--(axis cs:2.12375,37.3707809383596)
--(axis cs:2.12375,45.8125068992304)
--(axis cs:2.00125,45.8125068992304)
--(axis cs:2.00125,37.3707809383596)
--cycle;
\path [draw=white!18.8235294117647!black, fill=color3, semithick]
(axis cs:2.12625,32.1379221160441)
--(axis cs:2.24875,32.1379221160441)
--(axis cs:2.24875,41.0672270065889)
--(axis cs:2.12625,41.0672270065889)
--(axis cs:2.12625,32.1379221160441)
--cycle;
\path [draw=white!18.8235294117647!black, fill=color0, semithick]
(axis cs:2.75125,39.5376212375875)
--(axis cs:2.87375,39.5376212375875)
--(axis cs:2.87375,51.7839310678161)
--(axis cs:2.75125,51.7839310678161)
--(axis cs:2.75125,39.5376212375875)
--cycle;
\path [draw=white!18.8235294117647!black, fill=color1, semithick]
(axis cs:2.87625,28.953346062835)
--(axis cs:2.99875,28.953346062835)
--(axis cs:2.99875,42.7501236324068)
--(axis cs:2.87625,42.7501236324068)
--(axis cs:2.87625,28.953346062835)
--cycle;
\path [draw=white!18.8235294117647!black, fill=color2, semithick]
(axis cs:3.00125,37.7807056792584)
--(axis cs:3.12375,37.7807056792584)
--(axis cs:3.12375,49.8851324910236)
--(axis cs:3.00125,49.8851324910236)
--(axis cs:3.00125,37.7807056792584)
--cycle;
\path [draw=white!18.8235294117647!black, fill=color3, semithick]
(axis cs:3.12625,32.6723593596974)
--(axis cs:3.24875,32.6723593596974)
--(axis cs:3.24875,45.1479984260704)
--(axis cs:3.12625,45.1479984260704)
--(axis cs:3.12625,32.6723593596974)
--cycle;
\path [draw=white!18.8235294117647!black, fill=color0, semithick]
(axis cs:3.75125,44.581411603319)
--(axis cs:3.87375,44.581411603319)
--(axis cs:3.87375,53.267654652285)
--(axis cs:3.75125,53.267654652285)
--(axis cs:3.75125,44.581411603319)
--cycle;
\path [draw=white!18.8235294117647!black, fill=color1, semithick]
(axis cs:3.87625,34.9107031328404)
--(axis cs:3.99875,34.9107031328404)
--(axis cs:3.99875,47.5955636110461)
--(axis cs:3.87625,47.5955636110461)
--(axis cs:3.87625,34.9107031328404)
--cycle;
\path [draw=white!18.8235294117647!black, fill=color2, semithick]
(axis cs:4.00125,42.7438856721301)
--(axis cs:4.12375,42.7438856721301)
--(axis cs:4.12375,50.6716842649104)
--(axis cs:4.00125,50.6716842649104)
--(axis cs:4.00125,42.7438856721301)
--cycle;
\path [draw=white!18.8235294117647!black, fill=color3, semithick]
(axis cs:4.12625,38.0315118152524)
--(axis cs:4.24875,38.0315118152524)
--(axis cs:4.24875,46.4160735922224)
--(axis cs:4.12625,46.4160735922224)
--(axis cs:4.12625,38.0315118152524)
--cycle;
\draw[draw=white!18.8235294117647!black,fill=color0,line width=0.3pt] (axis cs:0,0) rectangle (axis cs:0,0);
\addlegendimage{ybar,ybar legend,draw=white!18.8235294117647!black,fill=color0,line width=0.3pt};

\draw[draw=white!18.8235294117647!black,fill=color1,line width=0.3pt] (axis cs:0,0) rectangle (axis cs:0,0);
\addlegendimage{ybar,ybar legend,draw=white!18.8235294117647!black,fill=color1,line width=0.3pt};

\draw[draw=white!18.8235294117647!black,fill=color2,line width=0.3pt] (axis cs:0,0) rectangle (axis cs:0,0);
\addlegendimage{ybar,ybar legend,draw=white!18.8235294117647!black,fill=color2,line width=0.3pt};

\draw[draw=white!18.8235294117647!black,fill=color3,line width=0.3pt] (axis cs:0,0) rectangle (axis cs:0,0);
\addlegendimage{ybar,ybar legend,draw=white!18.8235294117647!black,fill=color3,line width=0.3pt};

\addplot [semithick, white!18.8235294117647!black, forget plot]
table {%
-0.1875 26.154682731973
-0.1875 24.0732997766151
};
\addplot [semithick, white!18.8235294117647!black, forget plot]
table {%
-0.1875 37.2735804015945
-0.1875 49.9083038382701
};
\addplot [semithick, white!18.8235294117647!black, forget plot]
table {%
-0.218125 24.0732997766151
-0.156875 24.0732997766151
};
\addplot [semithick, white!18.8235294117647!black, forget plot]
table {%
-0.218125 49.9083038382701
-0.156875 49.9083038382701
};
\addplot [semithick, white!18.8235294117647!black, forget plot]
table {%
-0.0625 16.2966509090997
-0.0625 14.0766728995591
};
\addplot [semithick, white!18.8235294117647!black, forget plot]
table {%
-0.0625 26.8210790337983
-0.0625 35.0773181427334
};
\addplot [semithick, white!18.8235294117647!black, forget plot]
table {%
-0.093125 14.0766728995591
-0.031875 14.0766728995591
};
\addplot [semithick, white!18.8235294117647!black, forget plot]
table {%
-0.093125 35.0773181427334
-0.031875 35.0773181427334
};
\addplot [semithick, white!18.8235294117647!black, forget plot]
table {%
0.0625 24.7413226977366
0.0625 22.521326598645
};
\addplot [semithick, white!18.8235294117647!black, forget plot]
table {%
0.0625 35.490949961668
0.0625 47.8626276080546
};
\addplot [semithick, white!18.8235294117647!black, forget plot]
table {%
0.031875 22.521326598645
0.093125 22.521326598645
};
\addplot [semithick, white!18.8235294117647!black, forget plot]
table {%
0.031875 47.8626276080546
0.093125 47.8626276080546
};
\addplot [semithick, white!18.8235294117647!black, forget plot]
table {%
0.1875 19.7176999657374
0.1875 17.4711272751659
};
\addplot [semithick, white!18.8235294117647!black, forget plot]
table {%
0.1875 30.8586725278069
0.1875 42.7006289385669
};
\addplot [semithick, white!18.8235294117647!black, forget plot]
table {%
0.156875 17.4711272751659
0.218125 17.4711272751659
};
\addplot [semithick, white!18.8235294117647!black, forget plot]
table {%
0.156875 42.7006289385669
0.218125 42.7006289385669
};
\addplot [semithick, white!18.8235294117647!black, forget plot]
table {%
0.8125 32.3081777556577
0.8125 23.8134671636575
};
\addplot [semithick, white!18.8235294117647!black, forget plot]
table {%
0.8125 43.4691378619113
0.8125 50.3199309737319
};
\addplot [semithick, white!18.8235294117647!black, forget plot]
table {%
0.781875 23.8134671636575
0.843125 23.8134671636575
};
\addplot [semithick, white!18.8235294117647!black, forget plot]
table {%
0.781875 50.3199309737319
0.843125 50.3199309737319
};
\addplot [semithick, white!18.8235294117647!black, forget plot]
table {%
0.9375 22.302484947421
0.9375 13.9281921066055
};
\addplot [semithick, white!18.8235294117647!black, forget plot]
table {%
0.9375 32.2989396022749
0.9375 37.0983175252101
};
\addplot [semithick, white!18.8235294117647!black, forget plot]
table {%
0.906875 13.9281921066055
0.968125 13.9281921066055
};
\addplot [semithick, white!18.8235294117647!black, forget plot]
table {%
0.906875 37.0983175252101
0.968125 37.0983175252101
};
\addplot [semithick, white!18.8235294117647!black, forget plot]
table {%
1.0625 30.8184647841
1.0625 22.2659918547533
};
\addplot [semithick, white!18.8235294117647!black, forget plot]
table {%
1.0625 41.5185859580157
1.0625 48.0846116323367
};
\addplot [semithick, white!18.8235294117647!black, forget plot]
table {%
1.031875 22.2659918547533
1.093125 22.2659918547533
};
\addplot [semithick, white!18.8235294117647!black, forget plot]
table {%
1.031875 48.0846116323367
1.093125 48.0846116323367
};
\addplot [semithick, white!18.8235294117647!black, forget plot]
table {%
1.1875 25.8663019548549
1.1875 17.9182545705275
};
\addplot [semithick, white!18.8235294117647!black, forget plot]
table {%
1.1875 36.2028864426037
1.1875 43.2271971953681
};
\addplot [semithick, white!18.8235294117647!black, forget plot]
table {%
1.156875 17.9182545705275
1.218125 17.9182545705275
};
\addplot [semithick, white!18.8235294117647!black, forget plot]
table {%
1.156875 43.2271971953681
1.218125 43.2271971953681
};
\addplot [semithick, white!18.8235294117647!black, forget plot]
table {%
1.8125 38.929828149668
1.8125 25.2887359958223
};
\addplot [semithick, white!18.8235294117647!black, forget plot]
table {%
1.8125 48.1170145298489
1.8125 54.7504323344921
};
\addplot [semithick, white!18.8235294117647!black, forget plot]
table {%
1.781875 25.2887359958223
1.843125 25.2887359958223
};
\addplot [semithick, white!18.8235294117647!black, forget plot]
table {%
1.781875 54.7504323344921
1.843125 54.7504323344921
};
\addplot [semithick, white!18.8235294117647!black, forget plot]
table {%
1.9375 28.2130998686219
1.9375 18.4725068639878
};
\addplot [semithick, white!18.8235294117647!black, forget plot]
table {%
1.9375 35.7832921793715
1.9375 44.9090849726613
};
\addplot [semithick, white!18.8235294117647!black, forget plot]
table {%
1.906875 18.4725068639878
1.968125 18.4725068639878
};
\addplot [semithick, white!18.8235294117647!black, forget plot]
table {%
1.906875 44.9090849726613
1.968125 44.9090849726613
};
\addplot [semithick, white!18.8235294117647!black, forget plot]
table {%
2.0625 37.3707809383596
2.0625 27.9496406724907
};
\addplot [semithick, white!18.8235294117647!black, forget plot]
table {%
2.0625 45.8125068992304
2.0625 51.9824955787384
};
\addplot [semithick, white!18.8235294117647!black, forget plot]
table {%
2.031875 27.9496406724907
2.093125 27.9496406724907
};
\addplot [semithick, white!18.8235294117647!black, forget plot]
table {%
2.031875 51.9824955787384
2.093125 51.9824955787384
};
\addplot [semithick, white!18.8235294117647!black, forget plot]
table {%
2.1875 32.1379221160441
2.1875 23.282490037323
};
\addplot [semithick, white!18.8235294117647!black, forget plot]
table {%
2.1875 41.0672270065889
2.1875 46.1236487835709
};
\addplot [semithick, white!18.8235294117647!black, forget plot]
table {%
2.156875 23.282490037323
2.218125 23.282490037323
};
\addplot [semithick, white!18.8235294117647!black, forget plot]
table {%
2.156875 46.1236487835709
2.218125 46.1236487835709
};
\addplot [semithick, white!18.8235294117647!black, forget plot]
table {%
2.8125 39.5376212375875
2.8125 25.518774719136
};
\addplot [semithick, white!18.8235294117647!black, forget plot]
table {%
2.8125 51.7839310678161
2.8125 57.8852631442099
};
\addplot [semithick, white!18.8235294117647!black, forget plot]
table {%
2.781875 25.518774719136
2.843125 25.518774719136
};
\addplot [semithick, white!18.8235294117647!black, forget plot]
table {%
2.781875 57.8852631442099
2.843125 57.8852631442099
};
\addplot [semithick, white!18.8235294117647!black, forget plot]
table {%
2.9375 28.953346062835
2.9375 12.9194941921647
};
\addplot [semithick, white!18.8235294117647!black, forget plot]
table {%
2.9375 42.7501236324068
2.9375 51.0733596565318
};
\addplot [semithick, white!18.8235294117647!black, forget plot]
table {%
2.906875 12.9194941921647
2.968125 12.9194941921647
};
\addplot [semithick, white!18.8235294117647!black, forget plot]
table {%
2.906875 51.0733596565318
2.968125 51.0733596565318
};
\addplot [semithick, white!18.8235294117647!black, forget plot]
table {%
3.0625 37.7807056792584
3.0625 23.7181165979857
};
\addplot [semithick, white!18.8235294117647!black, forget plot]
table {%
3.0625 49.8851324910236
3.0625 54.6193554881931
};
\addplot [semithick, white!18.8235294117647!black, forget plot]
table {%
3.031875 23.7181165979857
3.093125 23.7181165979857
};
\addplot [semithick, white!18.8235294117647!black, forget plot]
table {%
3.031875 54.6193554881931
3.093125 54.6193554881931
};
\addplot [semithick, white!18.8235294117647!black, forget plot]
table {%
3.1875 32.6723593596974
3.1875 18.4886444255691
};
\addplot [semithick, white!18.8235294117647!black, forget plot]
table {%
3.1875 45.1479984260704
3.1875 47.1577819220036
};
\addplot [semithick, white!18.8235294117647!black, forget plot]
table {%
3.156875 18.4886444255691
3.218125 18.4886444255691
};
\addplot [semithick, white!18.8235294117647!black, forget plot]
table {%
3.156875 47.1577819220036
3.218125 47.1577819220036
};
\addplot [semithick, white!18.8235294117647!black, forget plot]
table {%
3.8125 44.581411603319
3.8125 40.0191973210327
};
\addplot [semithick, white!18.8235294117647!black, forget plot]
table {%
3.8125 53.267654652285
3.8125 61.3909458597049
};
\addplot [semithick, white!18.8235294117647!black, forget plot]
table {%
3.781875 40.0191973210327
3.843125 40.0191973210327
};
\addplot [semithick, white!18.8235294117647!black, forget plot]
table {%
3.781875 61.3909458597049
3.843125 61.3909458597049
};
\addplot [semithick, white!18.8235294117647!black, forget plot]
table {%
3.9375 34.9107031328404
3.9375 29.5660902765674
};
\addplot [semithick, white!18.8235294117647!black, forget plot]
table {%
3.9375 47.5955636110461
3.9375 56.7930903828187
};
\addplot [semithick, white!18.8235294117647!black, forget plot]
table {%
3.906875 29.5660902765674
3.968125 29.5660902765674
};
\addplot [semithick, white!18.8235294117647!black, forget plot]
table {%
3.906875 56.7930903828187
3.968125 56.7930903828187
};
\addplot [semithick, white!18.8235294117647!black, forget plot]
table {%
4.0625 42.7438856721301
4.0625 38.5638361046666
};
\addplot [semithick, white!18.8235294117647!black, forget plot]
table {%
4.0625 50.6716842649104
4.0625 57.698986243221
};
\addplot [semithick, white!18.8235294117647!black, forget plot]
table {%
4.031875 38.5638361046666
4.093125 38.5638361046666
};
\addplot [semithick, white!18.8235294117647!black, forget plot]
table {%
4.031875 57.698986243221
4.093125 57.698986243221
};
\addplot [semithick, white!18.8235294117647!black, forget plot]
table {%
4.1875 38.0315118152524
4.1875 33.4356014763258
};
\addplot [semithick, white!18.8235294117647!black, forget plot]
table {%
4.1875 46.4160735922224
4.1875 49.2087743102403
};
\addplot [semithick, white!18.8235294117647!black, forget plot]
table {%
4.156875 33.4356014763258
4.218125 33.4356014763258
};
\addplot [semithick, white!18.8235294117647!black, forget plot]
table {%
4.156875 49.2087743102403
4.218125 49.2087743102403
};
\addplot [semithick, white!18.8235294117647!black, forget plot]
table {%
-0.24875 34.1156731676724
-0.12625 34.1156731676724
};
\addplot [semithick, white!18.8235294117647!black, forget plot]
table {%
-0.12375 22.6691379500151
-0.00125 22.6691379500151
};
\addplot [semithick, white!18.8235294117647!black, forget plot]
table {%
0.00125 32.8706692252297
0.12375 32.8706692252297
};
\addplot [semithick, white!18.8235294117647!black, forget plot]
table {%
0.12625 27.4915629448059
0.24875 27.4915629448059
};
\addplot [semithick, white!18.8235294117647!black, forget plot]
table {%
0.75125 39.908248535626
0.87375 39.908248535626
};
\addplot [semithick, white!18.8235294117647!black, forget plot]
table {%
0.87625 28.3939525727347
0.99875 28.3939525727347
};
\addplot [semithick, white!18.8235294117647!black, forget plot]
table {%
1.00125 38.8561948312576
1.12375 38.8561948312576
};
\addplot [semithick, white!18.8235294117647!black, forget plot]
table {%
1.12625 32.9140197225445
1.24875 32.9140197225445
};
\addplot [semithick, white!18.8235294117647!black, forget plot]
table {%
1.75125 40.8958033642603
1.87375 40.8958033642603
};
\addplot [semithick, white!18.8235294117647!black, forget plot]
table {%
1.87625 30.7662963832676
1.99875 30.7662963832676
};
\addplot [semithick, white!18.8235294117647!black, forget plot]
table {%
2.00125 39.5419257899851
2.12375 39.5419257899851
};
\addplot [semithick, white!18.8235294117647!black, forget plot]
table {%
2.12625 34.8936209649844
2.24875 34.8936209649844
};
\addplot [semithick, white!18.8235294117647!black, forget plot]
table {%
2.75125 48.9033439811294
2.87375 48.9033439811294
};
\addplot [semithick, white!18.8235294117647!black, forget plot]
table {%
2.87625 35.6374823226114
2.99875 35.6374823226114
};
\addplot [semithick, white!18.8235294117647!black, forget plot]
table {%
3.00125 46.3072894053953
3.12375 46.3072894053953
};
\addplot [semithick, white!18.8235294117647!black, forget plot]
table {%
3.12625 41.2175849700592
3.24875 41.2175849700592
};
\addplot [semithick, white!18.8235294117647!black, forget plot]
table {%
3.75125 50.6464971675529
3.87375 50.6464971675529
};
\addplot [semithick, white!18.8235294117647!black, forget plot]
table {%
3.87625 39.2462511011068
3.99875 39.2462511011068
};
\addplot [semithick, white!18.8235294117647!black, forget plot]
table {%
4.00125 48.0728829074591
4.12375 48.0728829074591
};
\addplot [semithick, white!18.8235294117647!black, forget plot]
table {%
4.12625 42.8748634565636
4.24875 42.8748634565636
};
\end{axis}

\end{tikzpicture}

%% file: figs/cifar10c_brier.tex
\begin{tikzpicture}[baseline, trim axis left]

\definecolor{color0}{rgb}{0.0906862745098039,0.425980392156863,0.611274509803922}
\definecolor{color1}{rgb}{0.758823529411765,0.511764705882353,0.711764705882353}
\definecolor{color2}{rgb}{0.834803921568628,0.802450980392157,0.290686274509804}
\definecolor{color3}{rgb}{0.084313725490196,0.543137254901961,0.416666666666667}

\tikzstyle{every node}=[font=\scriptsize]

\begin{axis}[
width=\figcifarcappendixwidth,
height=\figcifarcappendixheight,
legend cell align={left},
legend style={nodes={scale=0.75, transform shape}, fill opacity=0.8, draw opacity=1, text opacity=1, at={(0.97,0.03)}, anchor=south east, draw=white!80!black},
tick align=inside,
tick pos=left,
x grid style={white!69.0196078431373!black},
xlabel={Severity},
xmin=-0.5, xmax=4.5,
xtick style={draw=none},
xtick={0,1,2,3,4},
xticklabels={1,2,3,4,5},
y grid style={white!69.0196078431373!black},
ylabel={Brier $\downarrow$},
ymin=0.0417033419013023, ymax=0.143381090462208,
ytick style={draw=none},
ytick={0.06, 0.08, 0.1, 0.12, 0.14},
yticklabels={0.06, 0.08, 0.1, 0.12, 0.14},
ymajorgrids
]
\path [draw=white!18.8235294117647!black, fill=color0, semithick]
(axis cs:-0.24875,0.0604143254458904)
--(axis cs:-0.12625,0.0604143254458904)
--(axis cs:-0.12625,0.0887218117713928)
--(axis cs:-0.24875,0.0887218117713928)
--(axis cs:-0.24875,0.0604143254458904)
--cycle;
\path [draw=white!18.8235294117647!black, fill=color1, semithick]
(axis cs:-0.12375,0.053229933604598)
--(axis cs:-0.00125,0.053229933604598)
--(axis cs:-0.00125,0.0776454173028469)
--(axis cs:-0.12375,0.0776454173028469)
--(axis cs:-0.12375,0.053229933604598)
--cycle;
\path [draw=white!18.8235294117647!black, fill=color2, semithick]
(axis cs:0.00125,0.0596404708921909)
--(axis cs:0.12375,0.0596404708921909)
--(axis cs:0.12375,0.0873748622834682)
--(axis cs:0.00125,0.0873748622834682)
--(axis cs:0.00125,0.0596404708921909)
--cycle;
\path [draw=white!18.8235294117647!black, fill=color3, semithick]
(axis cs:0.12625,0.0576611720025539)
--(axis cs:0.24875,0.0576611720025539)
--(axis cs:0.24875,0.0838083103299141)
--(axis cs:0.12625,0.0838083103299141)
--(axis cs:0.12625,0.0576611720025539)
--cycle;
\path [draw=white!18.8235294117647!black, fill=color0, semithick]
(axis cs:0.75125,0.0773797854781151)
--(axis cs:0.87375,0.0773797854781151)
--(axis cs:0.87375,0.105267431586981)
--(axis cs:0.75125,0.105267431586981)
--(axis cs:0.75125,0.0773797854781151)
--cycle;
\path [draw=white!18.8235294117647!black, fill=color1, semithick]
(axis cs:0.87625,0.0682545825839043)
--(axis cs:0.99875,0.0682545825839043)
--(axis cs:0.99875,0.0916072577238083)
--(axis cs:0.87625,0.0916072577238083)
--(axis cs:0.87625,0.0682545825839043)
--cycle;
\path [draw=white!18.8235294117647!black, fill=color2, semithick]
(axis cs:1.00125,0.076315850019455)
--(axis cs:1.12375,0.076315850019455)
--(axis cs:1.12375,0.10354145616293)
--(axis cs:1.00125,0.10354145616293)
--(axis cs:1.00125,0.076315850019455)
--cycle;
\path [draw=white!18.8235294117647!black, fill=color3, semithick]
(axis cs:1.12625,0.0733269304037094)
--(axis cs:1.24875,0.0733269304037094)
--(axis cs:1.24875,0.0988099426031113)
--(axis cs:1.12625,0.0988099426031113)
--(axis cs:1.12625,0.0733269304037094)
--cycle;
\path [draw=white!18.8235294117647!black, fill=color0, semithick]
(axis cs:1.75125,0.0939298607409)
--(axis cs:1.87375,0.0939298607409)
--(axis cs:1.87375,0.11585333943367)
--(axis cs:1.75125,0.11585333943367)
--(axis cs:1.75125,0.0939298607409)
--cycle;
\path [draw=white!18.8235294117647!black, fill=color1, semithick]
(axis cs:1.87625,0.0822720304131508)
--(axis cs:1.99875,0.0822720304131508)
--(axis cs:1.99875,0.105352386832237)
--(axis cs:1.87625,0.105352386832237)
--(axis cs:1.87625,0.0822720304131508)
--cycle;
\path [draw=white!18.8235294117647!black, fill=color2, semithick]
(axis cs:2.00125,0.0923782028257847)
--(axis cs:2.12375,0.0923782028257847)
--(axis cs:2.12375,0.113708455115557)
--(axis cs:2.00125,0.113708455115557)
--(axis cs:2.00125,0.0923782028257847)
--cycle;
\path [draw=white!18.8235294117647!black, fill=color3, semithick]
(axis cs:2.12625,0.0883896537125111)
--(axis cs:2.24875,0.0883896537125111)
--(axis cs:2.24875,0.108790814876556)
--(axis cs:2.12625,0.108790814876556)
--(axis cs:2.12625,0.0883896537125111)
--cycle;
\path [draw=white!18.8235294117647!black, fill=color0, semithick]
(axis cs:2.75125,0.096099991351366)
--(axis cs:2.87375,0.096099991351366)
--(axis cs:2.87375,0.122362051159143)
--(axis cs:2.75125,0.122362051159143)
--(axis cs:2.75125,0.096099991351366)
--cycle;
\path [draw=white!18.8235294117647!black, fill=color1, semithick]
(axis cs:2.87625,0.0840246006846428)
--(axis cs:2.99875,0.0840246006846428)
--(axis cs:2.99875,0.109671700745821)
--(axis cs:2.87625,0.109671700745821)
--(axis cs:2.87625,0.0840246006846428)
--cycle;
\path [draw=white!18.8235294117647!black, fill=color2, semithick]
(axis cs:3.00125,0.0945499539375305)
--(axis cs:3.12375,0.0945499539375305)
--(axis cs:3.12375,0.119078602641821)
--(axis cs:3.00125,0.119078602641821)
--(axis cs:3.00125,0.0945499539375305)
--cycle;
\path [draw=white!18.8235294117647!black, fill=color3, semithick]
(axis cs:3.12625,0.0898945517838001)
--(axis cs:3.24875,0.0898945517838001)
--(axis cs:3.24875,0.11369539052248)
--(axis cs:3.12625,0.11369539052248)
--(axis cs:3.12625,0.0898945517838001)
--cycle;
\path [draw=white!18.8235294117647!black, fill=color0, semithick]
(axis cs:3.75125,0.109168417751789)
--(axis cs:3.87375,0.109168417751789)
--(axis cs:3.87375,0.126027338206768)
--(axis cs:3.75125,0.126027338206768)
--(axis cs:3.75125,0.109168417751789)
--cycle;
\path [draw=white!18.8235294117647!black, fill=color1, semithick]
(axis cs:3.87625,0.0972301252186298)
--(axis cs:3.99875,0.0972301252186298)
--(axis cs:3.99875,0.120301518589258)
--(axis cs:3.87625,0.120301518589258)
--(axis cs:3.87625,0.0972301252186298)
--cycle;
\path [draw=white!18.8235294117647!black, fill=color2, semithick]
(axis cs:4.00125,0.107304651290178)
--(axis cs:4.12375,0.107304651290178)
--(axis cs:4.12375,0.12358695268631)
--(axis cs:4.00125,0.12358695268631)
--(axis cs:4.00125,0.107304651290178)
--cycle;
\path [draw=white!18.8235294117647!black, fill=color3, semithick]
(axis cs:4.12625,0.102336078882217)
--(axis cs:4.24875,0.102336078882217)
--(axis cs:4.24875,0.118761137127876)
--(axis cs:4.12625,0.118761137127876)
--(axis cs:4.12625,0.102336078882217)
--cycle;
\draw[draw=white!18.8235294117647!black,fill=color0,line width=0.3pt] (axis cs:0,0) rectangle (axis cs:0,0);
\addlegendimage{ybar,ybar legend,draw=white!18.8235294117647!black,fill=color0,line width=0.3pt};
\addlegendentry{MAP}

\draw[draw=white!18.8235294117647!black,fill=color1,line width=0.3pt] (axis cs:0,0) rectangle (axis cs:0,0);
\addlegendimage{ybar,ybar legend,draw=white!18.8235294117647!black,fill=color1,line width=0.3pt};
\addlegendentry{DE}

\draw[draw=white!18.8235294117647!black,fill=color2,line width=0.3pt] (axis cs:0,0) rectangle (axis cs:0,0);
\addlegendimage{ybar,ybar legend,draw=white!18.8235294117647!black,fill=color2,line width=0.3pt};
\addlegendentry{LA}

\draw[draw=white!18.8235294117647!black,fill=color3,line width=0.3pt] (axis cs:0,0) rectangle (axis cs:0,0);
\addlegendimage{ybar,ybar legend,draw=white!18.8235294117647!black,fill=color3,line width=0.3pt};
\addlegendentry{LULA}

\addplot [semithick, white!18.8235294117647!black, forget plot]
table {%
-0.1875 0.0604143254458904
-0.1875 0.0532573647797108
};
\addplot [semithick, white!18.8235294117647!black, forget plot]
table {%
-0.1875 0.0887218117713928
-0.1875 0.123079895973206
};
\addplot [semithick, white!18.8235294117647!black, forget plot]
table {%
-0.218125 0.0532573647797108
-0.156875 0.0532573647797108
};
\addplot [semithick, white!18.8235294117647!black, forget plot]
table {%
-0.218125 0.123079895973206
-0.156875 0.123079895973206
};
\addplot [semithick, white!18.8235294117647!black, forget plot]
table {%
-0.0625 0.053229933604598
-0.0625 0.0464726015925407
};
\addplot [semithick, white!18.8235294117647!black, forget plot]
table {%
-0.0625 0.0776454173028469
-0.0625 0.106073886156082
};
\addplot [semithick, white!18.8235294117647!black, forget plot]
table {%
-0.093125 0.0464726015925407
-0.031875 0.0464726015925407
};
\addplot [semithick, white!18.8235294117647!black, forget plot]
table {%
-0.093125 0.106073886156082
-0.031875 0.106073886156082
};
\addplot [semithick, white!18.8235294117647!black, forget plot]
table {%
0.0625 0.0596404708921909
0.0625 0.0526655726134777
};
\addplot [semithick, white!18.8235294117647!black, forget plot]
table {%
0.0625 0.0873748622834682
0.0625 0.120597258210182
};
\addplot [semithick, white!18.8235294117647!black, forget plot]
table {%
0.031875 0.0526655726134777
0.093125 0.0526655726134777
};
\addplot [semithick, white!18.8235294117647!black, forget plot]
table {%
0.031875 0.120597258210182
0.093125 0.120597258210182
};
\addplot [semithick, white!18.8235294117647!black, forget plot]
table {%
0.1875 0.0576611720025539
0.1875 0.0508342981338501
};
\addplot [semithick, white!18.8235294117647!black, forget plot]
table {%
0.1875 0.0838083103299141
0.1875 0.115359835326672
};
\addplot [semithick, white!18.8235294117647!black, forget plot]
table {%
0.156875 0.0508342981338501
0.218125 0.0508342981338501
};
\addplot [semithick, white!18.8235294117647!black, forget plot]
table {%
0.156875 0.115359835326672
0.218125 0.115359835326672
};
\addplot [semithick, white!18.8235294117647!black, forget plot]
table {%
0.8125 0.0773797854781151
0.8125 0.0525201708078384
};
\addplot [semithick, white!18.8235294117647!black, forget plot]
table {%
0.8125 0.105267431586981
0.8125 0.122719511389732
};
\addplot [semithick, white!18.8235294117647!black, forget plot]
table {%
0.781875 0.0525201708078384
0.843125 0.0525201708078384
};
\addplot [semithick, white!18.8235294117647!black, forget plot]
table {%
0.781875 0.122719511389732
0.843125 0.122719511389732
};
\addplot [semithick, white!18.8235294117647!black, forget plot]
table {%
0.9375 0.0682545825839043
0.9375 0.0463250577449799
};
\addplot [semithick, white!18.8235294117647!black, forget plot]
table {%
0.9375 0.0916072577238083
0.9375 0.105999797582626
};
\addplot [semithick, white!18.8235294117647!black, forget plot]
table {%
0.906875 0.0463250577449799
0.968125 0.0463250577449799
};
\addplot [semithick, white!18.8235294117647!black, forget plot]
table {%
0.906875 0.105999797582626
0.968125 0.105999797582626
};
\addplot [semithick, white!18.8235294117647!black, forget plot]
table {%
1.0625 0.076315850019455
1.0625 0.0519917346537113
};
\addplot [semithick, white!18.8235294117647!black, forget plot]
table {%
1.0625 0.10354145616293
1.0625 0.120278224349022
};
\addplot [semithick, white!18.8235294117647!black, forget plot]
table {%
1.031875 0.0519917346537113
1.093125 0.0519917346537113
};
\addplot [semithick, white!18.8235294117647!black, forget plot]
table {%
1.031875 0.120278224349022
1.093125 0.120278224349022
};
\addplot [semithick, white!18.8235294117647!black, forget plot]
table {%
1.1875 0.0733269304037094
1.1875 0.0501674152910709
};
\addplot [semithick, white!18.8235294117647!black, forget plot]
table {%
1.1875 0.0988099426031113
1.1875 0.114995509386063
};
\addplot [semithick, white!18.8235294117647!black, forget plot]
table {%
1.156875 0.0501674152910709
1.218125 0.0501674152910709
};
\addplot [semithick, white!18.8235294117647!black, forget plot]
table {%
1.156875 0.114995509386063
1.218125 0.114995509386063
};
\addplot [semithick, white!18.8235294117647!black, forget plot]
table {%
1.8125 0.0939298607409
1.8125 0.0741906762123108
};
\addplot [semithick, white!18.8235294117647!black, forget plot]
table {%
1.8125 0.11585333943367
1.8125 0.129963532090187
};
\addplot [semithick, white!18.8235294117647!black, forget plot]
table {%
1.781875 0.0741906762123108
1.843125 0.0741906762123108
};
\addplot [semithick, white!18.8235294117647!black, forget plot]
table {%
1.781875 0.129963532090187
1.843125 0.129963532090187
};
\addplot [semithick, white!18.8235294117647!black, forget plot]
table {%
1.9375 0.0822720304131508
1.9375 0.0497765690088272
};
\addplot [semithick, white!18.8235294117647!black, forget plot]
table {%
1.9375 0.105352386832237
1.9375 0.118642263114452
};
\addplot [semithick, white!18.8235294117647!black, forget plot]
table {%
1.906875 0.0497765690088272
1.968125 0.0497765690088272
};
\addplot [semithick, white!18.8235294117647!black, forget plot]
table {%
1.906875 0.118642263114452
1.968125 0.118642263114452
};
\addplot [semithick, white!18.8235294117647!black, forget plot]
table {%
2.0625 0.0923782028257847
2.0625 0.0728114023804665
};
\addplot [semithick, white!18.8235294117647!black, forget plot]
table {%
2.0625 0.113708455115557
2.0625 0.126295611262321
};
\addplot [semithick, white!18.8235294117647!black, forget plot]
table {%
2.031875 0.0728114023804665
2.093125 0.0728114023804665
};
\addplot [semithick, white!18.8235294117647!black, forget plot]
table {%
2.031875 0.126295611262321
2.093125 0.126295611262321
};
\addplot [semithick, white!18.8235294117647!black, forget plot]
table {%
2.1875 0.0883896537125111
2.1875 0.05832115188241
};
\addplot [semithick, white!18.8235294117647!black, forget plot]
table {%
2.1875 0.108790814876556
2.1875 0.118219941854477
};
\addplot [semithick, white!18.8235294117647!black, forget plot]
table {%
2.156875 0.05832115188241
2.218125 0.05832115188241
};
\addplot [semithick, white!18.8235294117647!black, forget plot]
table {%
2.156875 0.118219941854477
2.218125 0.118219941854477
};
\addplot [semithick, white!18.8235294117647!black, forget plot]
table {%
2.8125 0.096099991351366
2.8125 0.0603846572339535
};
\addplot [semithick, white!18.8235294117647!black, forget plot]
table {%
2.8125 0.122362051159143
2.8125 0.134432405233383
};
\addplot [semithick, white!18.8235294117647!black, forget plot]
table {%
2.781875 0.0603846572339535
2.843125 0.0603846572339535
};
\addplot [semithick, white!18.8235294117647!black, forget plot]
table {%
2.781875 0.134432405233383
2.843125 0.134432405233383
};
\addplot [semithick, white!18.8235294117647!black, forget plot]
table {%
2.9375 0.0840246006846428
2.9375 0.0499765276908875
};
\addplot [semithick, white!18.8235294117647!black, forget plot]
table {%
2.9375 0.109671700745821
2.9375 0.125214219093323
};
\addplot [semithick, white!18.8235294117647!black, forget plot]
table {%
2.906875 0.0499765276908875
2.968125 0.0499765276908875
};
\addplot [semithick, white!18.8235294117647!black, forget plot]
table {%
2.906875 0.125214219093323
2.968125 0.125214219093323
};
\addplot [semithick, white!18.8235294117647!black, forget plot]
table {%
3.0625 0.0945499539375305
3.0625 0.0596353374421597
};
\addplot [semithick, white!18.8235294117647!black, forget plot]
table {%
3.0625 0.119078602641821
3.0625 0.129914954304695
};
\addplot [semithick, white!18.8235294117647!black, forget plot]
table {%
3.031875 0.0596353374421597
3.093125 0.0596353374421597
};
\addplot [semithick, white!18.8235294117647!black, forget plot]
table {%
3.031875 0.129914954304695
3.093125 0.129914954304695
};
\addplot [semithick, white!18.8235294117647!black, forget plot]
table {%
3.1875 0.0898945517838001
3.1875 0.0575108006596565
};
\addplot [semithick, white!18.8235294117647!black, forget plot]
table {%
3.1875 0.11369539052248
3.1875 0.120935447514057
};
\addplot [semithick, white!18.8235294117647!black, forget plot]
table {%
3.156875 0.0575108006596565
3.218125 0.0575108006596565
};
\addplot [semithick, white!18.8235294117647!black, forget plot]
table {%
3.156875 0.120935447514057
3.218125 0.120935447514057
};
\addplot [semithick, white!18.8235294117647!black, forget plot]
table {%
3.8125 0.109168417751789
3.8125 0.0968920886516571
};
\addplot [semithick, white!18.8235294117647!black, forget plot]
table {%
3.8125 0.126027338206768
3.8125 0.13875937461853
};
\addplot [semithick, white!18.8235294117647!black, forget plot]
table {%
3.781875 0.0968920886516571
3.843125 0.0968920886516571
};
\addplot [semithick, white!18.8235294117647!black, forget plot]
table {%
3.781875 0.13875937461853
3.843125 0.13875937461853
};
\addplot [semithick, white!18.8235294117647!black, forget plot]
table {%
3.9375 0.0972301252186298
3.9375 0.0863288640975952
};
\addplot [semithick, white!18.8235294117647!black, forget plot]
table {%
3.9375 0.120301518589258
3.9375 0.131301909685135
};
\addplot [semithick, white!18.8235294117647!black, forget plot]
table {%
3.906875 0.0863288640975952
3.968125 0.0863288640975952
};
\addplot [semithick, white!18.8235294117647!black, forget plot]
table {%
3.906875 0.131301909685135
3.968125 0.131301909685135
};
\addplot [semithick, white!18.8235294117647!black, forget plot]
table {%
4.0625 0.107304651290178
4.0625 0.0954312086105347
};
\addplot [semithick, white!18.8235294117647!black, forget plot]
table {%
4.0625 0.12358695268631
4.0625 0.133596956729889
};
\addplot [semithick, white!18.8235294117647!black, forget plot]
table {%
4.031875 0.0954312086105347
4.093125 0.0954312086105347
};
\addplot [semithick, white!18.8235294117647!black, forget plot]
table {%
4.031875 0.133596956729889
4.093125 0.133596956729889
};
\addplot [semithick, white!18.8235294117647!black, forget plot]
table {%
4.1875 0.102336078882217
4.1875 0.0913707390427589
};
\addplot [semithick, white!18.8235294117647!black, forget plot]
table {%
4.1875 0.118761137127876
4.1875 0.123005226254463
};
\addplot [semithick, white!18.8235294117647!black, forget plot]
table {%
4.156875 0.0913707390427589
4.218125 0.0913707390427589
};
\addplot [semithick, white!18.8235294117647!black, forget plot]
table {%
4.156875 0.123005226254463
4.218125 0.123005226254463
};
\addplot [semithick, white!18.8235294117647!black, forget plot]
table {%
-0.24875 0.0803732797503471
-0.12625 0.0803732797503471
};
\addplot [semithick, white!18.8235294117647!black, forget plot]
table {%
-0.12375 0.0696142390370369
-0.00125 0.0696142390370369
};
\addplot [semithick, white!18.8235294117647!black, forget plot]
table {%
0.00125 0.0792200341820717
0.12375 0.0792200341820717
};
\addplot [semithick, white!18.8235294117647!black, forget plot]
table {%
0.12625 0.0758809298276901
0.24875 0.0758809298276901
};
\addplot [semithick, white!18.8235294117647!black, forget plot]
table {%
0.75125 0.0959075391292572
0.87375 0.0959075391292572
};
\addplot [semithick, white!18.8235294117647!black, forget plot]
table {%
0.87625 0.085718922317028
0.99875 0.085718922317028
};
\addplot [semithick, white!18.8235294117647!black, forget plot]
table {%
1.00125 0.0944178774952888
1.12375 0.0944178774952888
};
\addplot [semithick, white!18.8235294117647!black, forget plot]
table {%
1.12625 0.0904277712106705
1.24875 0.0904277712106705
};
\addplot [semithick, white!18.8235294117647!black, forget plot]
table {%
1.75125 0.100647307932377
1.87375 0.100647307932377
};
\addplot [semithick, white!18.8235294117647!black, forget plot]
table {%
1.87625 0.0896196216344833
1.99875 0.0896196216344833
};
\addplot [semithick, white!18.8235294117647!black, forget plot]
table {%
2.00125 0.0992972180247307
2.12375 0.0992972180247307
};
\addplot [semithick, white!18.8235294117647!black, forget plot]
table {%
2.12625 0.0951529890298843
2.24875 0.0951529890298843
};
\addplot [semithick, white!18.8235294117647!black, forget plot]
table {%
2.75125 0.116086676716805
2.87375 0.116086676716805
};
\addplot [semithick, white!18.8235294117647!black, forget plot]
table {%
2.87625 0.105777718126774
2.99875 0.105777718126774
};
\addplot [semithick, white!18.8235294117647!black, forget plot]
table {%
3.00125 0.114461623132229
3.12375 0.114461623132229
};
\addplot [semithick, white!18.8235294117647!black, forget plot]
table {%
3.12625 0.110007733106613
3.24875 0.110007733106613
};
\addplot [semithick, white!18.8235294117647!black, forget plot]
table {%
3.75125 0.121379412710667
3.87375 0.121379412710667
};
\addplot [semithick, white!18.8235294117647!black, forget plot]
table {%
3.87625 0.110767565667629
3.99875 0.110767565667629
};
\addplot [semithick, white!18.8235294117647!black, forget plot]
table {%
4.00125 0.118747562170029
4.12375 0.118747562170029
};
\addplot [semithick, white!18.8235294117647!black, forget plot]
table {%
4.12625 0.114368513226509
4.24875 0.114368513226509
};
\end{axis}

\end{tikzpicture}